%% file: paper-neurips-revised.tex
\documentclass{article}

\usepackage[main,final]{lib/neurips_2026}

\usepackage[utf8]{inputenc} 
\usepackage[T1]{fontenc}    
\usepackage{hyperref}       
\usepackage{url}            
\usepackage{booktabs}       
\usepackage{amsfonts}       
\usepackage{nicefrac}       
\usepackage{microtype}      
\usepackage{xcolor}         

\usepackage{subcaption}
\usepackage{mathtools}
\usepackage{amsmath, amssymb, amsthm}
\usepackage[capitalise]{cleveref}
\usepackage{lib/cmd}
\usepackage{graphicx}

\graphicspath{{media/}{media/theory/}}

\theoremstyle{plain}
\newtheorem{theorem}{Theorem}[section]
\newtheorem{proposition}[theorem]{Proposition}

\newtheorem{corollary}[theorem]{Corollary}
\theoremstyle{definition}
\newtheorem{definition}[theorem]{Definition}

\theoremstyle{remark}

\title{How Large Must a Model Be to Grok? \\ \large A competition between memorisation and generalisation timescales}

\author{%
  Yiding `Vincent' Song \\
  Harvard College\\
  Cambridge, MA 02138 \\
  \texttt{yidingsong@college.harvard.edu} \\
  \And
  Hanming Ye \\
  Harvard College\\
  Cambridge, MA 02138 \\
  \texttt{hanmingye@college.harvard.edu}
}

\begin{document}

\raggedbottom

\maketitle

\begin{abstract}
Despite a known dependence between grokking and model parameter count, how model capacity shapes grokking remains an open question. Focusing on small Transformers and modular arithmetic, we give a quantitative account of the relationship between model capacity and grokking as a competition between two timescales: generalisation ($T_{\text{gen}}$) and memorisation ($T_{\text{mem}}$).
Building on prior formalisms for the empirical information capacity of Transformers, we define and measure $T_{\text{gen}}$ on modular arithmetic datasets and $T_{\text{mem}}$ on random-label datasets of equivalent complexity, and show that their intersection correlates with model sizes where consistent grokking begins.
We also observe empirical laws for $T_{\text{gen}}$ and $T_{\text{mem}}$ as models and datasets scale, motivating their forms theoretically and finding that our model of $T_{\text{mem}}$ is consistent with prior observations that larger models tend to memorise faster.
Moreover, extrapolating our empirical timescale laws predicted the existence of datasets for which no model of any size would consistently grok, a pattern supported on the modular arithmetic datasets of larger primes, demonstrating that our framework can make qualitative predictions out-of-distribution.
\end{abstract}

\section{Introduction}

The grokking phenomenon \citep{power2022grokking} on modular arithmetic, where training accuracy saturates much earlier than test accuracy, has become a popular topic for studying the dynamics of memorisation and generalisation in neural networks.
Existing accounts investigate the mechanisms and training dynamics underlying grokking, for example: representation-learning dynamics \citep{liu2022towards}, implicit-bias and lazy-to-rich dynamical transitions \citep{lyu2024dichotomy,kumar2024grokking}, circuit-efficiency arguments under weight decay \citep{varma2023circuit,huang2024unified}, selective norm growth and the emergence of sparse generalising subnetworks in sparse-parity models \citep{merrill2023tale}, sample-complexity bounds for kernel and feature-learning regimes in modular addition \citep{mohamadi2024why}, and mechanistic analyses of Fourier circuits and their formation during Transformer training \citep{nanda2023progress}.

In addition, several papers have identified different learning regimes as model size and dataset size vary \citep{davies2023unifying,huang2024unified}.
However, the existing body of work does not explain fully how grokking depends on model size.
Here we investigate whether the parameter size at which grokking begins can be predicted.
A natural hypothesis is that grokking should begin only once a model's information capacity exceeds the dataset size it is trying to learn, giving it the possibility of memorisation. However, we show that empirically, models with estimated capacities well above the dataset complexity continue to generalise immediately.

We investigate whether independently measured memorisation and generalisation timescales can predict the model size at which grokking begins.
Building on the framework of \citet{morris2025memorize} to estimate the empirical information capacity of a model, we define two measurable timescales for a Transformer of parameter count \(P\) on a modular arithmetic task with prime \(p\): a {generalisation timescale} \(T_{\text{gen}}(P,p)\), the number of epochs to reach high validation accuracy on the modular arithmetic task; and a {memorisation timescale} \(T_{\text{mem}}(P,p)\), the number of epochs to fit a random-label dataset of equivalent information content. We focus on modular division over prime fields and a family of small decoder-only Transformers.
We find that the parameter count at which these two timescales intersect correlates with the empirical onset of grokking across primes, albeit with a systematic offset.
Even if the model has enough capacity to represent a memorising solution, memorisation may not be the preferred pattern discovered by gradient descent. Our hypothesis is that the competition between the timescales of memorisation and generalisation helps explain the observed onset of grokking.

\begin{figure*}[!t]
    \centering
    \begin{subfigure}{0.3043\linewidth}
        \includegraphics[width=\linewidth]{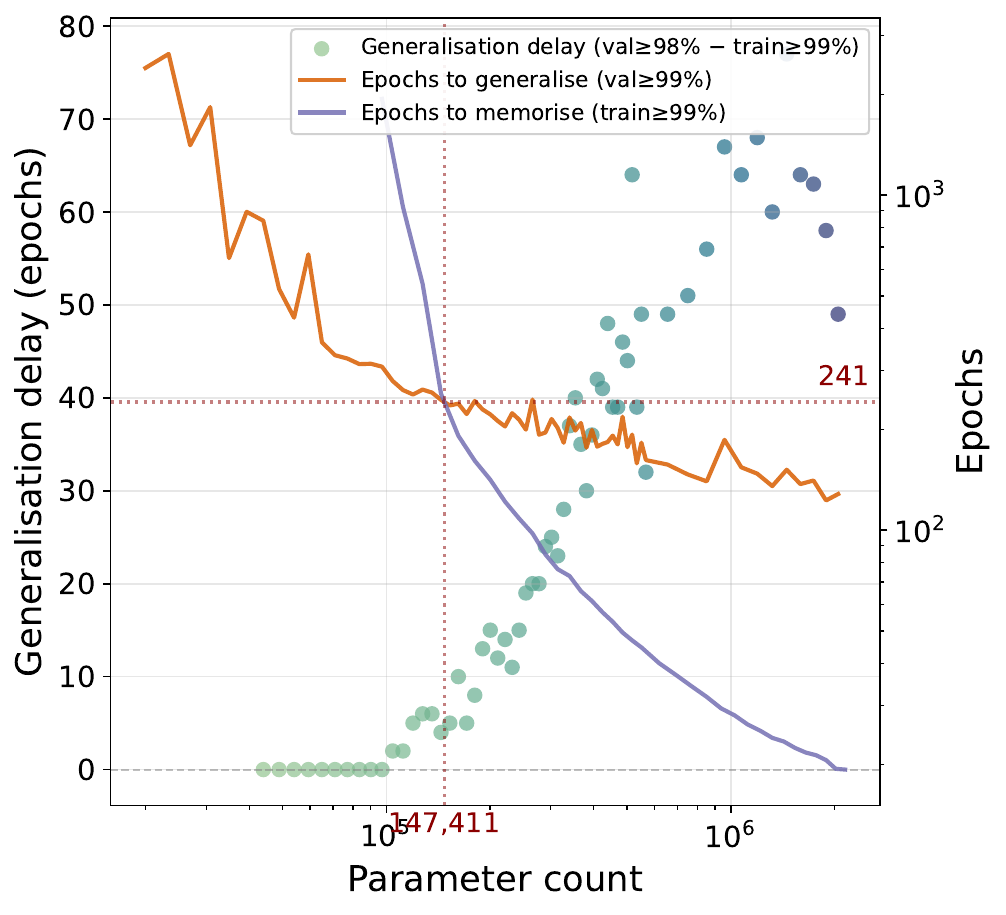}
        \caption{$p=97$}
    \end{subfigure}\hfill
    \begin{subfigure}{0.3043\linewidth}
        \includegraphics[width=\linewidth]{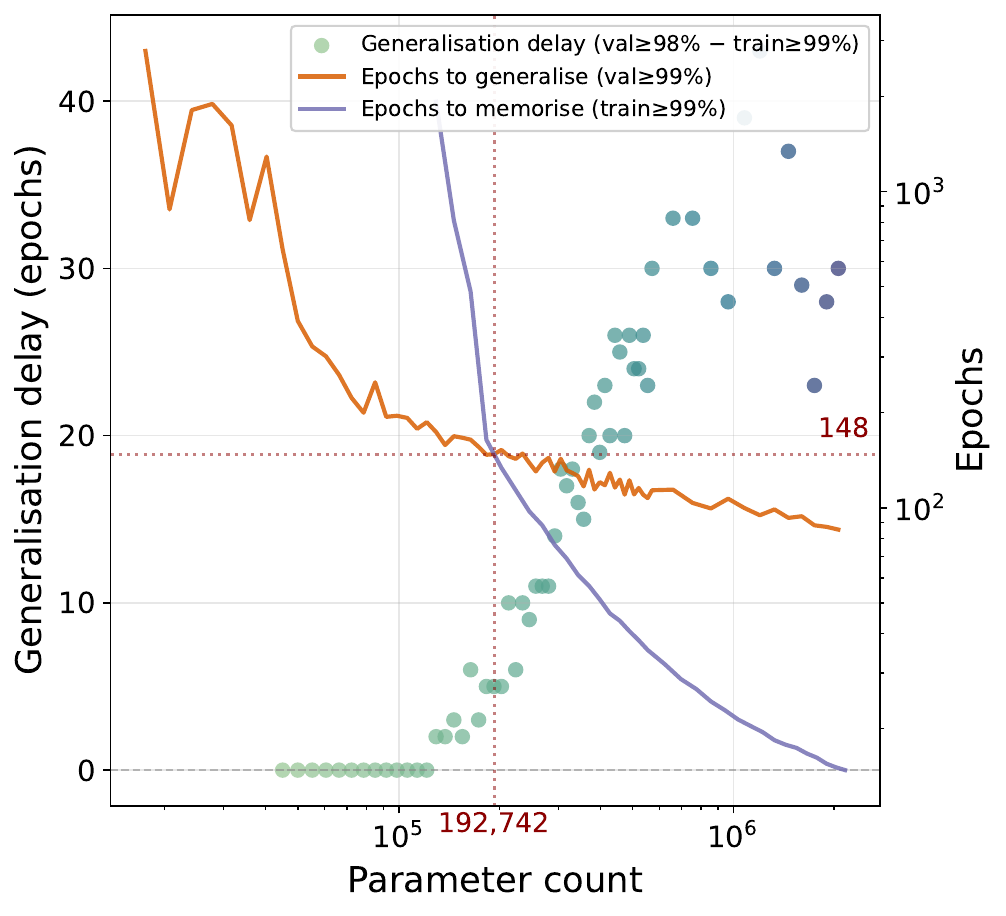}
        \caption{$p=113$}
    \end{subfigure}\hfill
    \begin{subfigure}{0.3514\linewidth}
        \includegraphics[width=\linewidth]{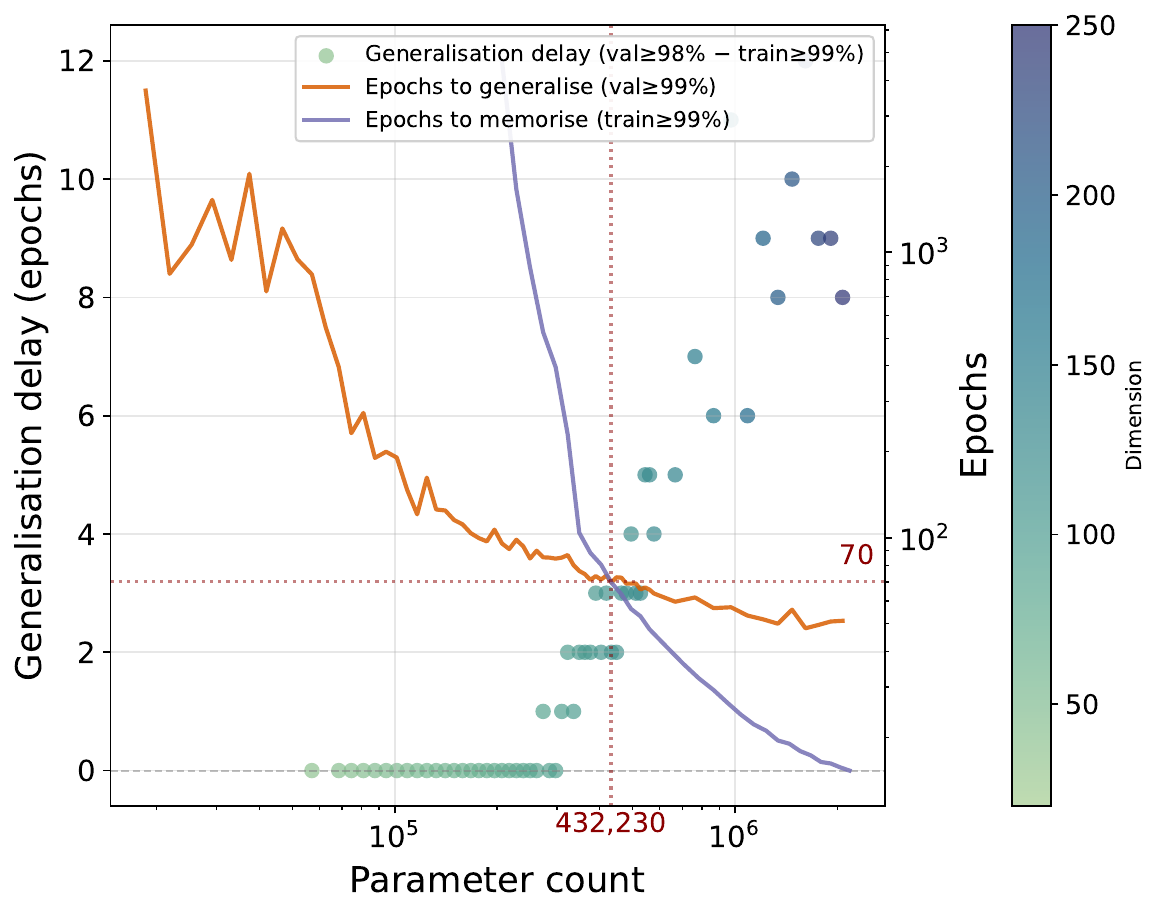}
        \caption{$p=139$}
    \end{subfigure}
    \caption{Generalisation delay and learning timescales across models of different sizes for three primes. Orange curve: epochs to generalise; purple curve: epochs to memorise a random dataset of the same complexity.
    The colour of each dot encodes its model dimension $d$, which is proportional to parameter count.
    Generalisation delay is taken as the minimum delay over ten seeds; timescale curves are arithmetic means of measured times.
    The dotted vertical line marks out $\widehat{P}_{\text{cross}}(p)$, the parameter size at which the timescale of memorisation intersects that of generalisation.
    Empirically, $\widehat{P}_{\text{cross}}(p)$ correlates with parameter count where generalisation delay first becomes consistently non-zero.}

    \label{fig:delay-with-speed}
\end{figure*}

We observe three regimes as we scale model parameter count $P$ (\cref{fig:delay-with-speed}) for a fixed dataset: an {underfitting} regime where both train and test accuracy remain low within the training budget; an {immediate-generalisation} regime where $T_{\text{gen}}(P,p) < T_{\text{mem}}(P,p)$ and there is no generalisation delay; and a {grokking} regime where $T_{\text{mem}}(P,p) \lesssim T_{\text{gen}}(P,p)$ and the model memorises before generalising.

Motivated by these observations, we develop our framework and make three main contributions:\footnote{The code for this paper is available at \url{https://github.com/PerceptronV/grokking-capacity}.}
\begin{enumerate}
    \item \textbf{Quantitative framework of how grokking depends on model capacity, via competing memorisation and generalisation timescales.}
    Adapting the framework of \citet{morris2025memorize} to estimate the information capacity of Transformers, we formalise measurements of $T_{\text{mem}}(P,p)$ and $T_{\text{gen}}(P,p)$ for Transformer models on modular arithmetic and show that the parameter count at which they intersect correlates with the empirical onset of grokking across different primes and hyperparameters.
    \item \textbf{Unified account of immediate generalisation in smaller models and faster memorisation in larger models.}
    In our modular-arithmetic experiments, smaller models can generalise without a delay even when their estimated model capacity exceeds the dataset complexity. Our framework shows that this corresponds to the regime in which generalisation is faster than memorisation. Empirically we also find that across different model and dataset sizes, memorisation time can be accurately predicted from the the ratio of dataset complexity to model capacity alone. This suggests faster memorisation as model size increases, consistent with earlier observations in language models \citep{tirumala2022memorization}.
    \item \textbf{Empirical memorisation and generalisation scaling laws, with qualitative predictions of model grokking behaviour on larger datasets.}
    We fit empirical laws to estimate $T_{\text{mem}}$ and $T_{\text{gen}}$, and provide theoretical motivations for their forms (\cref{sec:principled}). Extrapolating beyond the primes used to fit the laws predicts a critical prime $p^*\approx167.89$ beyond which no model size would consistently grok (\cref{prop:fold}). Although this estimate is not an exact boundary, the empirical pattern holds at larger primes: at $p=197$ and $199$, no sampled model size exhibits positive generalisation delay in all ten seeds (\cref{fig:cone,sec:theory-region}).
\end{enumerate}

\begin{figure}[!t]
    \centering
    \input{theory/cone-panels}
    \caption{(\subref{fig:cone-observed}) The region where grokking occurs, as a function of model parameter count and prime $p$; note that dataset size increases with $p$.
    The generalisation delay of each model during training was evaluated across 10 seeds for each combination of parameter count and dataset size, and the minimum observed generalisation delay reported.
    The colour of each dot shows the positive minimum generalisation delay; hollow markers denote a zero minimum. The red line is the predicted parameter size for which generalisation first becomes consistently delayed, calculated from the intersection of our empirical $T_{\mathrm{gen}}$ and $T_{\mathrm{mem}}$ scaling laws.
    The laws were fitted on primes up to and including $p=139$ (intersection show as solid red line) and extrapolated for larger primes (intersection shown as dotted red line).
    The vertical dashed line marks the critical prime $p^*=167.89$, beyond which our empirical laws predict no model would consistently grok.
    (\subref{fig:cone-mem},\subref{fig:cone-gen}) Memorisation and generalisation timescales, in epochs, as a function of parameter count and prime $p$.
    For points $p\le139$, we show the arithmetic means of observed times across 10 seeds; these were used to fit our empirical laws.
    For $p>139$, we show the timescales predicted by our empirical laws. Contours join points with equal timescales ($10$, $100$, and $1000$ epochs).
    The figures share a logarithmic colour scale, capped at yellow for any point above $10^4$.}
    \label{fig:cone}
\end{figure}

\section{Problem Setting and Notation}
\label{sec:setup}

\subsection{Modular arithmetic datasets}

Fix a prime $p$ and an operation $\circ\in\{+, -, \times, /\}$ on $\mathbb{Z}_p$.
For modular division we use $a/b \equiv a\cdot b^{-1} \pmod p$ with $b\in\{1,\dots,p-1\}$.
Following standard grokking benchmarks, we enumerate all valid pairs $(a,b)$ and build a dataset
\(
\mathcal{D}_p = \{(X_i, y_i)\}_{i=1}^{N_{\text{full}}}
\)
where each input $X_i$ is a length-$4$ token sequence encoding the equation
\(
X_i = [a,\ \texttt{op},\ b,\ \texttt{=}],
\)
and the label $y_i = a \circ b \in\{0,\dots,p-1\}$.
We reserve two special tokens for the operator and equals sign, so the vocabulary size is
\(
V = p + 2.
\)

For modular division the number of valid pairs is $N_{\text{full}} = p(p-1)$.
We randomly select a training subset $\mathcal{D}^{\text{train}}_p$ containing a fraction $\alpha\in(0,1)$ of these pairs and use the remainder as a held-out test set $\mathcal{D}^{\text{test}}_p$.
Throughout, we focus on $\alpha=1/2$, so that
\(
N_{\text{train}} = \alpha p(p-1)
\)
and $N_{\text{test}} = N_{\text{train}}$.

\subsection{Dataset complexity}

We are interested in the worst-case number of bits needed to memorise the labels
$y_i$ given inputs $X_i$.
Assuming random labels, each $y_i$ is uniformly distributed over $V$ possibilities and therefore requires $\log_2 V$ bits to specify.
For $n$ random labels, we call this information content {\it dataset complexity} and write it as $K=n\log_2 V$. For the modular-division training set, its dataset complexity is
\begin{equation}
    K_{\text{mem}}(p,\alpha)
    = N_{\text{train}} \log_2 V
    = \alpha p(p-1)\log_2(p+2).
    \label{eq:table-complexity}
\end{equation}
Note that modular arithmetic is a structured task which may admit a much more compact algorithmic description; that algorithmic complexity is beyond the current scope of this paper.

\subsection{Model architecture}

All experiments in our main sweep use the same decoder-only Transformer architecture with embedding width $d$, depth 2, and a single attention head. The model receives a sequence of integer token IDs $x\in\{0,\dots,V-1\}^L$ with length $L=4$ and vocabulary size $V=p+2$, and outputs logits over the vocabulary for the final position. Each of the two residual blocks contains causal self-attention followed by a gated feed-forward network (FFN) with hidden width $4d$, with root-mean-square (RMS) normalisation before each sub-block. A final RMSNorm and linear projection from $\mathbb{R}^d$ to $\mathbb{R}^V$ produce the output logits; we vary $P$ by sweeping the embedding width $d \in [10, 1000]$. Implementation details are given in \cref{sec:exp-arch}.

\subsection{Total memorisation}
\label{sec:memorisation-defs}

We briefly summarise the information-theoretic notions of memorisation we use, adapting \citet{morris2025memorize} to our discrete classification setting.
Consider a dataset
\(
D = \{(x_i, y_i)\}_{i=1}^n
\)
of $n$ examples with labels $y_i\in \{0,\dots,V-1\}$ and a model $p_\theta(y\mid x)$ parameterised by weights $\theta$.
We measure log-probabilities in bits, i.e.\ using $\log_2$.

\begin{definition}[$M_T$]
\label{def:mt}
The total memorisation of $p_\theta$ on $D$ is
\begin{equation}
    M_T(\theta; D)
    = \sum_{i=1}^n
      \left(
        \log_2 V
        + \log_2 p_\theta(y_i \mid x_i)
      \right).
    \label{eq:total-mem}
\end{equation}
\end{definition}
The measure compares the model's predictions against a uniform predictor that assigns probability $1/V$ to each label, requiring $\log_2 V$ bits to encode. Using the model's prediction instead requires $-\log_2 p_\theta(y_i\mid x_i)$ bits. Each term in $M_T$ therefore measures the reduction in bits over a uniform predictor for one data point. Under this definition, uniform guesses have $M_T=0$ while perfect predictions attain the maximum $n\log_2 V$.

\section{Model Capacity, Learning Timescales, and Grokking}
\label{sec:theory}

We now formalise our two main quantitative measures: the information capacity of a model, measured in bits per parameter, and memorisation and generalisation timescales, measured in the number of training epochs required to reach a saturation threshold.

\subsection{Empirical information capacity of a model}
\label{sec:theory-capacity}

Model capacity measures how much information a model can store. Following \citet{morris2025memorize}, we characterise it through total memorisation on random-label data. Let $\mathcal D$ be the collection of random-label datasets of all sizes, and let $\theta_P(D)$ denote the weights obtained by training a randomly initialised model with $P$ trainable parameters on $D$, within a fixed architecture family and under a fixed training protocol. Using the total memorisation formalised in \cref{def:mt}, we define
\begin{equation}
    \mathrm{Cap}(P):=\sup_{D\in\mathcal D}M_T\bigl(\theta_P(D);D\bigr).
\end{equation}
In practice, we estimate this supremum by taking maxima over a finite range of dataset sizes and averaging them across seeds. Meanwhile, we also introduce the capacity constant $C_{\mathrm{model}}$ to describe how model capacity scales with parameter count. We model this dependence as
\begin{equation}
    \mathrm{Cap}(P)\approx C_{\mathrm{model}}P+b_{\mathrm{cap}},
\end{equation}
where $C_{\mathrm{model}}$ is measured in bits per parameter and $b_{\mathrm{cap}}$ is the intercept.
Empirically, we find that $C_{\mathrm{model}}\approx0.785$ bits per parameter and $b_{\mathrm{cap}}\approx625$ bits. We also write $C:=C_{\mathrm{model}}$ for brevity. The measurements and fit diagnostics are reported in \cref{sec:exp-capacity}.

Combining $C_{\text{model}}$ with the dataset complexity $K_{\text{mem}}(p,\alpha)$ of \cref{eq:table-complexity} defines a parameter count
\begin{equation}
    P_{\text{mem}}(p,\alpha)
    = \frac{K_{\text{mem}}(p,\alpha)}{C_{\text{model}}},
    \label{eq:param-threshold}
\end{equation}
at which the predicted capacity of the model matches the dataset complexity; models smaller than this are not expected to be able to memorise the dataset.

\subsection{Defining learning timescales}
\label{sec:theory-speeds}

To capture learning dynamics, we treat grokking as a competition between the timescales of two processes: (1) a memorisation process that moves towards a solution overfitting the training set; (2) a generalisation process that moves towards a solution implementing modular division. We formalise these timescales using first-passage times under actual training. We write the timescales as functions of model parameter count $P$ and prime $p$, which determines dataset complexity at a fixed training fraction; note that each timescale also implicitly depends on training settings.

\paragraph{Memorisation timescale.}
To isolate purely the process of memorisation, we define memorisation timescale over a random-label dataset of equivalent complexity to the corresponding modular arithmetic dataset. Specifically, for a model with $P$ parameters trained on $n$ random-label examples with vocabulary size $V=p+2$, we record the first epoch at which training accuracy reaches $99\%$ and denote the resulting timescale by $\tau_{\text{mem}}(P,p,n)$. To compare this with generalisation on modular division, we use a random-label dataset with the same dataset complexity as the task's training set:
\begin{equation}
    T_{\text{mem}}(P,p)
    := \tau_{\text{mem}}(P,p,n_{\text{equiv}}(p,\alpha)),
    \quad
    n_{\text{equiv}}(p,\alpha)
    = \frac{K_{\text{mem}}(p,\alpha)}{\log_2 V},
\end{equation}
Thus $T_{\text{mem}}(P,p)$ measures the time needed to fit an equivalent amount of random-label information.

\paragraph{Generalisation timescale.}
The generalisation timescale $T_{\text{gen}}(P,p)$ measures how long a model takes to generalise to unseen modular-division examples. We train on $\mathcal{D}^{\text{train}}_p$ and record the first epoch at which validation accuracy on $\mathcal{D}^{\text{test}}_p$ reaches $99\%$. Note that some memorisation may also happen during this period; hence it is a limitation of our framework that our measurement of $T_{\text{gen}}$ may not be fully faithful. Questions of interference between memorisation and generalisation processes are beyond the current scope of this paper and left for future work.

For both timescales, our main experiment sweep uses ten seeds per combination of parameter count and prime $(P, p)$. We report arithmetic means of observed timescales. For each timescale, we discard runs in which the relevant training or validation accuracy does not saturate within the training budget. The averaging and fitting procedures are detailed in \cref{sec:stats-methods}.

To compare learning times across model and dataset sizes, we define the \emph{capacity fraction} as the ratio of dataset complexity to estimated model capacity:
\begin{equation}
    f(P,n) = \frac{K}{C_{\text{model}}P}
           = \frac{n\log_2 V}{C_{\text{model}}P},
    \label{eq:capacity-fraction}
\end{equation}
where $K=n\log_2 V$ is the dataset complexity in bits and $C_{\text{model}}P$ is the estimated model capacity in bits. For the memorisation timescale $T_{\mathrm{mem}}(P,p)$, we write $f$ for $f(P,n_{\mathrm{equiv}})$. Given a fixed dataset, this fraction decreases as model size increases. We empirically find that $T_{\text{mem}}(P,p) \propto e^{\gamma f}$ for a fitted memorisation exponent $\gamma$ and $f \in [0.018, 0.513]$ (\cref{sec:result-capacity}).

\subsection{Quantifying grokking}
\label{sec:theory-grokking}

\paragraph{Generalisation delay.}
We quantify grokking by the delay between memorising the training data and generalising. For seed $s$, this delay is
\begin{equation}
    \Delta E_s(P)
    =
    \max\big\{0,\,
        E_{\text{val},s}(P) - E_{\text{train},s}(P)
    \big\},
\end{equation}
where $E_{\text{train},s}(P)$ is the first epoch when training accuracy reaches $99\%$ and $E_{\text{val},s}(P)$ is the first epoch when validation accuracy reaches $98\%$ for that seed.\footnote{We set the validation threshold at $98\%$ because some observed runs sit above $98\%$ for many epochs before matching the training accuracy, and we did not want to count those as grokking. See \cref{tab:threshold-sensitivity} for results under matched thresholds.}
A zero $\Delta E_s(P)$ corresponds to immediate generalisation (no grokking).
We compute $\Delta E_s(P)$ across ten seeds for each model size and define $\Delta E(P):=\min_s\Delta E_s(P)$, because we are interested in finding the smallest model size for which every seed has positive delay.
Positive $\Delta E(P)$ corresponds to consistent grokking across every seed.

\paragraph{Grokking point.}
We define the grokking point $P_{\text{onset}}(p)$ as the smallest measured model size $P$ for which $\Delta E(P)>0$. This marks the beginning of consistent grokking across seeds. Conventions on unresolved outcomes and the range of model widths used are given in \cref{sec:stats-methods}.

\section{Experiments and Results}
\label{sec:results}

\subsection{Model capacity and memorisation speed}
\label{sec:result-capacity}

Following the established random-label capacity framework of \citet{morris2025memorize}, we estimate $C_{\mathrm{model}}\approx0.785$ bits per parameter for our Transformer family. This empirical scale lets us compare memorisation times across model and dataset sizes through the capacity fraction $f=K/(C_{\mathrm{model}}P)$. The capacity measurements and fit are reported in \cref{sec:exp-capacity,fig:capacity-subfigs}.

\begin{figure*}[t]
    \centering
    \begin{subfigure}{0.48\linewidth}
        \centering
        \includegraphics[width=\linewidth]{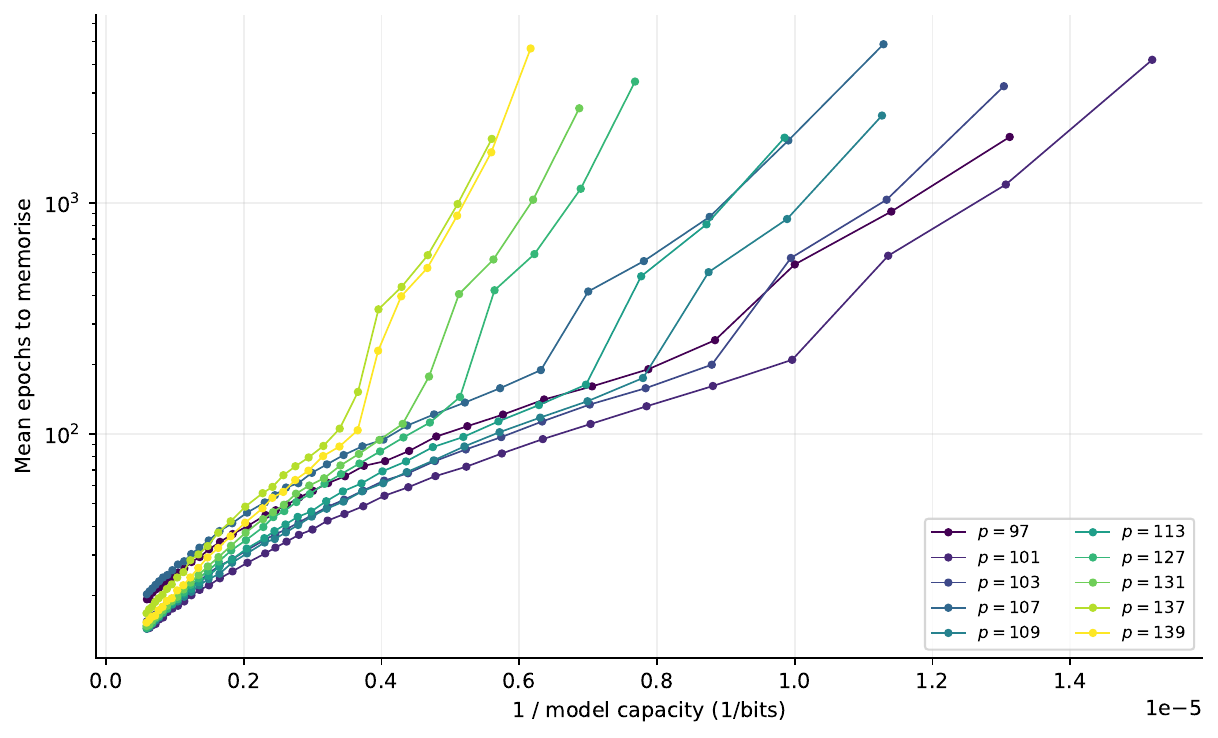}
        \caption{Saturation time against $1/(C_{\text{model}} P)$.}
        \label{fig:saturation-inverse-capacity}
    \end{subfigure}\hfill
    \begin{subfigure}{0.48\linewidth}
        \centering
        \includegraphics[width=\linewidth]{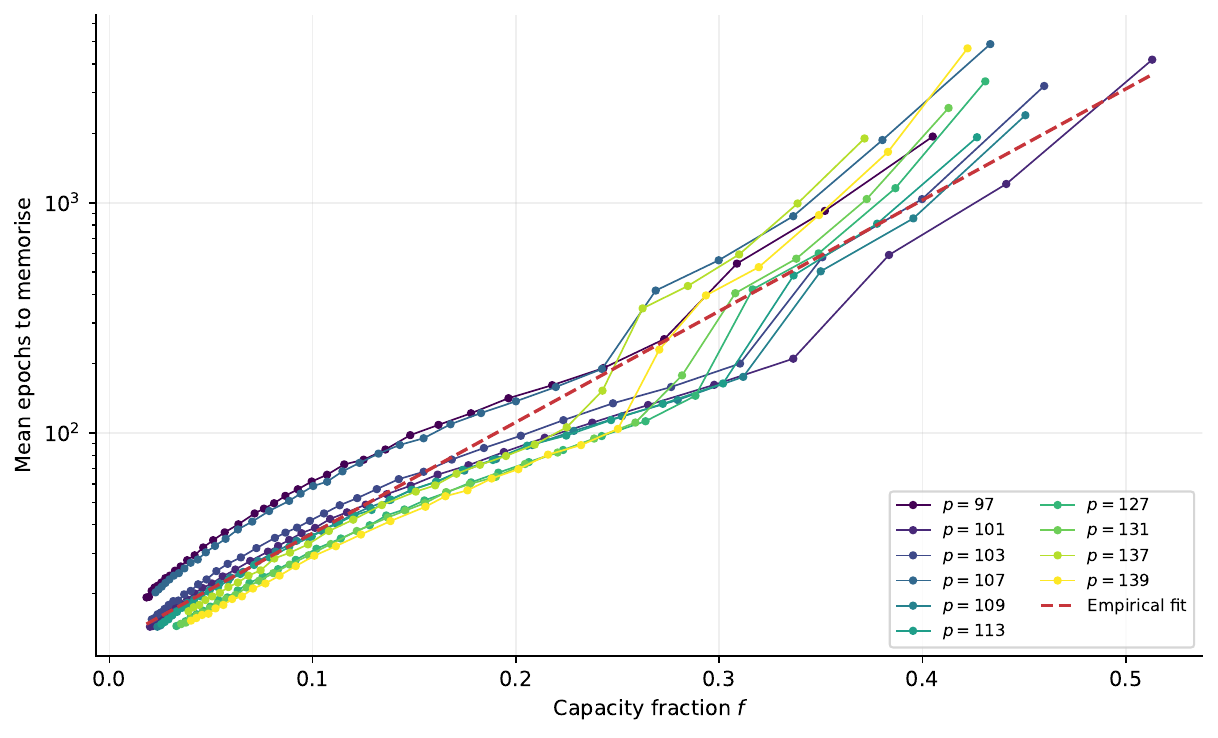}
        \caption{Saturation time against $f(P, n_{\text{equiv}})$.}
        \label{fig:saturation-capacity-fraction}
    \end{subfigure}
    \caption{Memorisation timescale experiments. \cref{fig:saturation-inverse-capacity} shows that memorisation time increases with inverse model capacity $1/(C_{\text{model}} P)$ across datasets; i.e. smaller models take longer to memorise. The shape of the curve varies across datasets. \cref{fig:saturation-capacity-fraction} shows that when epochs to saturation is plotted against capacity fraction $f(P, n_{\text{equiv}})$, all the datasets collapse roughly onto a single curve, suggesting that memorisation timescale may be estimated from the capacity fraction alone.}
    \label{fig:speed-summary}
\end{figure*}

We measured the memorisation time $\tau_{\text{mem}}(P,p,n)$ for several model sizes and dataset sizes, and for each we computed the corresponding capacity fraction $f(P, n)$.
\cref{fig:saturation-inverse-capacity} shows that memorisation time increases with inverse model capacity $1/(C_{\text{model}}P)$ across datasets; i.e. smaller models take longer to memorise. Meanwhile, \cref{fig:saturation-capacity-fraction} shows that when epochs to saturation is plotted against capacity fraction, points across all the datasets collapse roughly onto a single curve, suggesting that memorisation time can be estimated from capacity fraction alone, rather than a combination of model size or dataset complexity. In particular, epochs until memorisation increases with capacity fraction, and we find that empirically, $T_{\text{mem}}(P,p) \approx 12.068\exp(11.109f)$ over the observed range $f \in [0.0183,0.5130]$. We note that this exponential is an empirical functional fit rather than a mechanistically-derived law; nonetheless, it provides a principled account of the previously reported observation that larger models memorise faster \citep{tirumala2022memorization} through the dependence on capacity fraction: a fixed dataset complexity occupies a smaller fraction of a larger model's capacity, hence larger models memorise the same datasets faster. Fit details are reported in \cref{sec:exp-speed}.

\subsection{Generalisation delay}
\label{sec:result-grokking}
We trained a range of models with different internal dimensions $d$ and primes $p$ and recorded training and validation accuracies and losses at each epoch, allowing us to compute $E_{\text{train},s}(P), E_{\text{val},s}(P),$ and $\Delta E(P)$.
Training protocols and experimental details are given in \cref{sec:exp-arch,sec:exp-grokking}.
We also plotted the generalisation delay $\Delta E(P)$ against parameter count in \cref{fig:delay-with-speed} for several primes, allowing us to visualise the various regimes of memorisation and generalisation as a function of model size.
Using \cref{eq:param-threshold}, we compute the estimated parameter count $P_{\text{mem}}$ at which model capacity matches the complexity of each modular-division dataset.
We find that for several primes (including $p=113$ and $p=139$ shown in \cref{fig:delay-with-speed}), the first models capable of achieving near-perfect training and test accuracy have parameter counts $P$ slightly below but on the same order as $P_{\text{mem}}$, indicating that gradient descent is able to find algorithmic solutions that are more compact than the raw dataset.

Surprisingly, $\Delta E(P)$ remains at zero as model size increases for small models, even when their estimated model capacity exceeds the dataset complexity. The exact point at which $\Delta E(P)$ becomes non-zero is dependent on the prime $p$ and the training fraction $\alpha$, and is strongly correlated with the intersection of the memorisation and generalisation timescales, as we show next.

\subsection{The intersection of $T_{\text{mem}}$ and $T_{\text{gen}}$ correlates with the onset of grokking}
\label{sec:result-intersection}

For each prime $p$, we inspect the memorisation timescale experiments for matching model sizes, where we computed the memorisation time $T_{\text{mem}}(P,p)$ for a random dataset of the same complexity $K_{\text{mem}}(p,\alpha)$.
We plot the generalisation delay $\Delta E(P)$ and the generalisation time $T_{\text{gen}}(P,p)$ against parameter count on a common $x$-axis.
\cref{fig:delay-with-speed} shows the resulting plots for three primes.

Several patterns emerge across the plots:
\begin{enumerate}
    \item For small parameter counts, we see $T_{\text{gen}}(P,p) < T_{\text{mem}}(P,p)$: generalisation is faster than memorisation, and generalisation delays are zero.
    \item As $P$ increases, $T_{\text{mem}}(P,p)$ decreases more steeply than $T_{\text{gen}}(P,p)$.
    At the measured lower intersection $\widehat{P}_{\text{cross}}(p)$, the two curves intersect.
    \item Empirically, $\widehat{P}_{\text{cross}}(p)$ correlates with the onset of grokking with a systematic offset.
\end{enumerate}

To assess this quantitatively, we test three aspects of the hypothesis that the intersection of $T_{\text{mem}}$ and $T_{\text{gen}}$ predicts the empirical grokking onset: whether predicted and observed onsets are correlated, whether their parameter counts agree, and whether the intersection improves on simpler predictors. We report a test for each (\cref{tab:hypothesis-tests}).
\textit{(i) Correlation with observed onset.} The parameter counts at the measured intersections are strongly correlated with observed onset parameter counts (Spearman $\rho=0.988$, permutation $p=10^{-4}$).
\textit{(ii) Agreement in parameter count.} The slope of log observed onset against log predicted onset is $0.948$, but the predictions systematically overestimate the onset parameter count ($F=1007.40$, $p=2.45\times10^{-10}$). The median $\log_{10}$ ratio of observed to predicted onset parameter count is $-0.170$: onset is about $32\%$ below the intersection prediction, equivalently the prediction is about $48\%$ above onset.
\textit{(iii) Comparison with simpler predictors.} Adding the parameter count at the intersection to a linear model using prime alone improves its fit to the observed onsets ($F=155.79$, $p=4.88\times10^{-6}$). Adding prime to the model based on the intersection gives no statistically significant improvement ($p=0.396$). Adding the intersection also improves a model based on $P_{\mathrm{mem}}$ alone ($F=170.35$, $p=3.61\times10^{-6}$), suggesting that the measured learning timescales provide information beyond the estimate from model capacity.

Taken together, these results provide evidence that grokking onset is associated with the intersection of memorisation and generalisation timescales, with a systematic offset. Although the association is strong, ten prime-level comparisons provide limited precision for estimating its accuracy and distinguishing the contributions of correlated predictors. Methods and sensitivity analyses are given in \cref{sec:stats-methods,tab:threshold-sensitivity}.

\begin{table}[t]
\centering\small
\caption{Tests of whether the parameter count at the intersection of memorisation and generalisation timescales predicts where grokking begins, across all 10 primes in the main sweep. Spearman and Kendall correlations assess whether predictions rank the observed onsets correctly; Lin's concordance correlation coefficient (CCC) and the following rows assess agreement in their values. The lower rows compare four models for log onset parameter count: $M_0$ predicts the same onset for every prime, $M_1$ uses log parameter count at the timescale intersection, $M_2$ uses only prime $p$, and $M_3$ uses both predictors. Each comparison tests whether adding a predictor improves the fit. The final row tests whether the intersection improves on $P_{\mathrm{mem}}$, the estimated parameter count at which model capacity matches dataset complexity. These comparisons use the same observations for fitting and evaluation. Statistical procedures are described in \cref{sec:stats-methods}.}
\label{tab:hypothesis-tests}
\begin{tabular}{lll}
\toprule
Test & Statistic & $p$-value \\
\midrule
Spearman $\rho$ & $0.988$ & $10.00\times10^{-5}$ \\
Kendall $\tau$ & $0.956$ & $5.51\times10^{-6}$ \\
Lin's CCC (95\% interval) & $0.699\;[0.468,0.763]$ & --- \\
Log--log slope & $0.948$ & --- \\
Intercept & $0.108$ & --- \\
Calibration $F$ test & $1007.40$ & $2.45\times10^{-10}$ \\
Median log-residual; Wilcoxon & $-0.170$ & $0.002$ \\
$M_1$ vs. $M_0$ & $2231.39$ & $4.46\times10^{-11}$ \\
$M_3$ vs. $M_2$ & $155.79$ & $4.88\times10^{-6}$ \\
$M_3$ vs. $M_1$ & $0.82$ & $0.396$ \\
Intersection adds over $P_{\mathrm{mem}}$ & $170.35$ & $3.61\times10^{-6}$ \\
\bottomrule
\end{tabular}
\end{table}

\paragraph{Robustness to other hyperparameters, architectures, and tasks.}
The main experimental sweep holds every hyperparameter constant except for the prime $p$ and the embedding width $d$. We test the framework's robustness to other settings by varying weight decay (\cref{sec:hparam-weight-decay}), learning rate (\cref{sec:hparam-lr}), initialisation scale (\cref{sec:hparam-init-scale}), training fraction (\cref{sec:hparam-train-fraction}), and model depth (\cref{sec:arch-depth}), as well as the different task types of modular addition and multiplication (\cref{sec:task-modular-add,sec:task-modular-mul}).
The weight-decay, learning-rate, training-fraction, and task sweeps support the framework: there is an association between the intersection of timescales and empirical grokking onset, given both could be identified successfully. However, the offset between the intersection prediction and the observed onset varies across settings, so quantitative predictions require calibration for each training setting and task. Meanwhile, the results for initialisation scale and model depth remain inconclusive: given our training protocol and budgets, both the timescale intersection and empirical grokking onset could only be determined for five comparisons in the former and two in the latter, each corresponding to a particular prime and hyperparameter value. Further experiments are hence needed to assess the framework's applicability in these cases. Refer to \cref{sec:stats-methods} for details.

\input{theory/theory-section}

\section{Related Work}
\label{sec:related}

\paragraph{Grokking (delayed generalisation).}
Grokking was introduced by \citet{power2022grokking} in small Transformers trained on algorithmic tasks, including modular division.
Subsequent work has examined the algorithms learned on these tasks: \citet{nanda2023progress} identify Fourier circuits in modular addition and trace their formation before test accuracy improves, while \citet{furuta2024interpreting} study a broader family of modular functions.
Related studies investigate in-context learning and skill composition on modular tasks \citep{he2024learning}, and implicit composition and comparison over synthetic factual knowledge \citep{wang2024grokked}.
Grokking has also been induced on non-algorithmic data \citep{liu2022omnigrok}, while \citet{humayun2024deep} study delayed improvements in adversarial robustness on standard datasets.
Generalisation delay also depends on how a model represents the task: \citet{pearce2023memorize} illustrate immediate generalisation with a fixed circular embedding, and \citet{minegishi2025bridging} show that sparse subnetworks selected from grokked models can learn with much less delay when retrained from their initial weights. These works mainly focus on representations and datasets.

\paragraph{Mechanisms and learning dynamics.}
Several accounts explain why generalising solutions are favoured but take time to dominate.
\citet{varma2023circuit} describe competition between a rapidly learned memorising circuit and a slower, more norm-efficient generalising circuit favoured by weight decay.
\citet{huang2024unified} extend this picture across model and dataset sizes, including partial generalisation before complete memorisation.
In sparse-parity experiments, \citet{merrill2023tale} associate grokking with selective norm growth in a small subnetwork.
\citet{liu2022towards} derive an effective generalisation timescale from competing representation-learning and decoder-fitting dynamics in a toy addition model.
\citet{liu2022omnigrok} provide a complementary account based on training and test losses as functions of weight norm.

Other accounts emphasise feature learning.
\citet{kumar2024grokking} show how output scaling, initial kernel-task alignment, and dataset size can produce delayed feature learning.
\citet{lyu2024dichotomy} derive distinct early- and late-phase implicit biases under large initialisation and small weight decay, yielding grokking in specific models.
For modular addition, \citet{mohamadi2024why} establish squared-loss limitations for permutation-equivariant kernels and conditional generalisation guarantees for feature-learning networks.
\citet{rubin2023grokking} analyse equilibrium feature-learning transitions in two-layer models, including those driven by width.
Other work tracks grokking through information synergy and redundancy \citep{clauw2024information} or network complexity estimated through lossy compression \citep{demoss2025complexity}.
These approaches relate learning progress to internal network changes. We study whether measured learning timescales predict the model size at which grokking begins.

\paragraph{Pattern-learning speeds and training interventions.}
Our framework builds on the idea that memorising and generalising patterns are learned at different rates.
\citet{davies2023unifying} use this idea to connect grokking with double descent and demonstrate model-wise grokking by varying Transformer size on modular division.
Epoch-wise double descent also shows how generalisation can change non-monotonically with training time \citep{nakkiran2020deep}.
Training interventions further demonstrate the importance of optimisation: \citet{lee2024grokfast} accelerate grokking by amplifying slowly varying gradient components, and \citet{thilak2022slingshot} identify late-stage slingshot cycles in adaptive optimisers that often accompany grokking without explicit regularisation.
\citet{manir2026systematic} study interactions between architecture, optimisation, and regularisation on modular arithmetic, reporting substantial dependence on the training protocol.

\paragraph{Memorisation and empirical capacity.}
Our random-label measurements build on established methods of studying memorisation.
\citet{zhang2017rethinking} show that deep networks can fit random labels.
\citet{arpit2017closer} compare fitting dynamics on real and noisy data, finding that increasing model capacity accelerates fitting more strongly on noisy data.
Random-label controls also appear in \citet{varma2023circuit}, who use them to isolate memorising circuits and measure their efficiency.
In language models, \citet{carlini2022quantifying} quantify verbatim training-data extraction, while \citet{tirumala2022memorization} measure training-token prediction over time and find that larger models memorise faster.
Our random-label experiments and empirical scaling laws for memorisation recover a similar size dependence.
Most directly, \citet{morris2025memorize} estimate capacity from training on random tokens and relate improved generalisation to dataset size exceeding that capacity.
We adapt their capacity measurement to study learning timescales on modular arithmetic.

\section{Discussion}
\label{sec:discussion}

We formalise memorisation $T_{\text{mem}}$ and generalisation $T_{\text{gen}}$ timescales for small Transformers learning modular arithmetic. Across a range of primes and training protocols, we show that the onset of consistent grokking correlates with a model size at which $T_{\text{mem}}$ and $T_{\text{gen}}$ intersect. Additionally, we observe empirical laws for how $T_{\text{mem}}$ and $T_{\text{gen}}$ scale with model and dataset sizes, and provide some theoretical motivation for the forms of the laws. We show that these empirical laws, in conjunction with our framework, can make qualitative predictions about grokking behaviour on out-of-distribution model and dataset sizes, albeit their quantitative accuracy remains imperfect.

\paragraph{Limitations.}
Our main statistical comparison uses only observations from ten primes. More data and extrapolation tests on out-of-distribution prime datasets would better assess predictive accuracy (\cref{sec:stats-methods}). Experiments on initialisation scale and model depth remain inconclusive, and improvements over simpler predictors remain unclear across tasks (\cref{sec:hparam-invariance,sec:arch-invariance,sec:task-invariance}). Broader model-size coverage, longer training budgets, and more thorough hyperparameter sweeps would help resolve these gaps.
Moreover, our definition of $T_{\mathrm{gen}}$ does not account for interference from the likely concurrent process of memorisation; we leave the question of isolating measurements of $T_{\mathrm{gen}}$ without memorisation to future work. Additionally, our framework remains largely empirical despite its theoretical underpinnings.
Our simplified theoretical models do not recover the observed scaling of $T_{\mathrm{gen}}$ with prime and model size (\cref{sec:principled});
meanwhile, our hyperparameter sweeps show that making quantitative predictions across different training protocols requires new calibrations for each setting. Deriving these empirical constants and scaling laws theoretically is a worthwhile goal.

\paragraph{Broader outlook.}
We hope to extend our framework to tasks beyond modular arithmetic and different model architectures, such as MLPs or larger Transformers.
In light of prior work on grokking beyond algorithmic datasets \citep{liu2022omnigrok}, we are particularly interested in naturalistic tasks. A major project here would be to rigorously define, measure, and hopefully derive a model's generalisation timescales on naturalistic data.

\section*{Acknowledgements}

The authors are grateful to Professor Cengiz Pehlevan for suggestions and feedback on the paper, and for teaching the class at Harvard that inspired it (AM 226, Theory of Neural Computation). This paper would not have come about without his thoughts and enthusiasm.
The authors also thank Ziming Liu, Chongxi Lai, Kazuki Irie, Christopher Bates, and Professor Sam Gershman for discussions on ideas and questions related to the scope of this paper.
The authors would like to further thank their family, friends, and mentors, Yanjun Ding, Xingkai Song, W. S. Lam, T. Lee, M. Alifierakis, C. T. Pollitt, J. A. P. Bedford, S. C. Lim, R. C. H. Allison, S. Mishra-Sharma, J. Thaler, U. U. Zampa, H. Leung, W. Tang, A. G. Robbins, J. M. Cox-Donovan, M. Jihad, L. Blackappl, S. Karim, and C. Gensbigler for inspiration in all walks of life.
Finally, the authors extend their gratitude to E. Boddie, B. Tyler-Funes, B. Nasir, S. Calon, Dr. U. Kelechi, Dr. J. Dion, and the mighty Pacific Ocean.

Parts of the paper were completed while the authors were interns at Waddle Laboratories Inc. (`Waddle'), for which Waddle provided compute resources (2x H100s) and LLM agent subscriptions (access to the latest Claude Code and Codex plans) for general machine learning research and scholarship. Waddle did not influence the design, implementation, or analysis of the studies presented in this paper in any way.

\label{end-main}
\bibliographystyle{plainnat}
\bibliography{references}


\appendix

\section{Experiment details}
\label{sec:experiments}

\subsection{Modular division dataset}
The main sweep uses the ten primes between 90 and 140: 97, 101, 103, 107, 109, 113, 127, 131, 137, and 139. The secondary sweeps and extrapolation experiments use the prime sets specified below.

We enumerate the $p(p-1)$ ordered pairs $(a,b)$ with $a\in\{0,\ldots,p-1\}$ and $b\in\{1,\ldots,p-1\}$, and encode each as the four-token sequence $[a,\mathrm{op},b,\mathrm{equals}]$. The two special tokens have IDs $p$ and $p+1$. A seeded NumPy permutation selects $\lfloor\alpha p(p-1)\rfloor$ training pairs without replacement; the remaining pairs form the validation set. Main-sweep runs use $\alpha=1/2$ and seeds $42$--$51$. Reusing a seed across widths preserves the data split; observations sharing a seed should therefore not automatically be regarded as independent replicates.

The random-data experiments sample all four input tokens and the target independently and uniformly from the $V=p+2$ vocabulary entries. They do not preserve arithmetic input syntax. Thus $K_{\mathrm{mem}}=n\log_2 V$ is the chosen information-matching convention; the arithmetic labels themselves occupy only $p$ numeric classes. Equal nominal information content does not imply equal input geometry or optimisation difficulty.

\subsection{Architecture and optimisation}
\label{sec:exp-arch}
All experiments in the main sweep train decoder-only Transformers in PyTorch using AdamW \citep{loshchilov2017decoupled} with learning rate $\eta=10^{-3}$, moment coefficients $(\beta_1,\beta_2)=(0.9,0.98)$, weight decay $\lambda=1.0$, and batch size $B_{\mathrm{batch}}=512$, for a maximum of $5{,}000$ epochs. We fix depth $L_{\text{depth}}=2$ and a single attention head $H=1$, varying model size by changing the embedding width $d$. Dropout is $0.2$ in all three experiment types.

\paragraph{Implemented architecture.}
Each residual block applies RMS normalisation before causal self-attention and before a gated SiLU feed-forward network of hidden width $4d$. RMSNorm uses $\epsilon=10^{-5}$ and a trainable gain initialised to one; a fifth RMSNorm precedes the final-token readout. Input embeddings and the vocabulary projection are untied, and all linear maps in this one-head configuration are bias-free. Dropout acts on the elementwise product of the SiLU-transformed and linear feed-forward branches. The attention module's dropout member is unused, and its optional additional output projection/dropout path is absent at $H=1$; the attention output projection remains. The total parameter count is
\begin{equation}
 P=32d^2+(2p+9)d.
 \label{eq:implemented-param-count}
\end{equation}

\paragraph{Effective position handling.}
The implementation contains a rotary positional embedding (RoPE) module, but passes queries and keys in the order (batch, heads, tokens, features), whereas the module expects (batch, tokens, heads, features). It therefore indexes the head axis as position. With $H=1$, every token receives position zero, and the rotation is exactly the identity. The reported one-head experiments consequently have no effective rotary positional encoding. Their causal attention mask still distinguishes preceding from subsequent tokens. A correctly ordered RoPE implementation would define a different experimental protocol; the results here describe the implementation used to generate the data.

\paragraph{Initialisation and optimisation.}
We use the default PyTorch initialisation of the embedding and linear modules and unit RMSNorm gains. The initialisation-scale setting multiplies every parameter, including the gains; it is one in the main sweep. AdamW uses $\epsilon=10^{-8}$ and applies weight decay to all trainable parameters, including embeddings and normalisation gains. The learning rate is constant, with no scheduler, gradient clipping, or mixed-precision autocast in the training loop.

\paragraph{Epochs, accuracy, and stopping.}
Each epoch reshuffles the training examples using \texttt{torch.randperm} and includes the final partial minibatch. Training accuracy accumulates predictions made before each minibatch update, with dropout active. Validation accuracy is evaluated after the epoch with dropout disabled. Primary first-passage times count completed epochs from one; no pre-training evaluation is included. Random-label speed runs stop at the first training-$99\%$ event. Main-sweep task runs stop when both training and validation accuracy reach $99\%$, or at the epoch cap. Delay uses training $99\%$ and validation $98\%$, whereas both timescale laws use $99\%$. Unobserved events remain censored and are excluded before forming timescale means; a cap is not substituted for an event time. Campaign-specific budgets and unresolved-delay rules are given in \cref{sec:exp-cone}; auxiliary character traces use the separate evaluation convention in \cref{sec:character-protocol}.

\subsection{Compute resources}
\label{sec:exp-compute}
Experiments use NVIDIA A100 and H100 GPUs, with each run assigned to one GPU. Hardware and protocol are retained in the run manifests; both primary timescale laws use A100 runs, the matched capacity estimate uses H100 runs, and each cone cell uses one hardware cohort. Per-run elapsed time is recorded where available. A complete project-wide GPU-hour accounting, including preliminary and failed runs, has not been verified, so no aggregate cost is claimed here.

The matched capacity manifest records float32 parameters, PyTorch 2.7.0, CUDA 12.8, and an H100 80\,GB HBM3 GPU, with matrix-multiplication TF32 disabled and cuDNN TF32 enabled. These recorded environment settings apply to that campaign; they are not asserted for A100 runs without equivalent runtime records. Matching architecture and training settings across devices does not imply bit-identical trajectories. The fitted laws, capacity calibration, and cone cohorts are selected separately as specified in \cref{sec:exp-capacity,sec:exp-cone,sec:extrapolation-details}.

\subsection{Estimating capacity on random data}
\label{sec:exp-capacity}

For each model dimension $d \in \{10, 12, 14, 16, 18, 22\}$ and dataset size $n \in \{1000,\, 1374,\, 1891,\, 2601,\, 3576,\, 4917,\, 6761,\, 9300\}$ we generate a random dataset $(X_{\text{train}},T_{\text{train}})$ of size $n$ (inputs and targets, respectively), sampling each token and target uniformly from the vocabulary of size $V=p+2$ at $p=113$. We train as in \cref{sec:exp-arch}, with two seeds (20261103, 20261104) per cell, giving $96$ H100 runs. Early stopping monitors online training loss: training ends after $100$ epochs without an improvement exceeding $10^{-4}$, or at $5{,}000$ epochs. We evaluate the final checkpoint with dropout disabled to compute total memorisation $M_T$ via \cref{eq:total-mem} and bits per example $m_T=M_T/n$. We do not restore a best-loss checkpoint or maximise storage over epochs.

The score in \cref{eq:total-mem} has no finite lower bound: if the probability assigned to any observed label approaches zero, its log-probability, and hence $M_T$, tends to $-\infty$.

For each width and seed, we take the maximum measured storage over the eight dataset sizes, then average the two seed maxima:
\begin{equation}
    \widehat{\mathrm{Cap}}_k=\frac{1}{2}\sum_{s=1}^{2}\max_{D\in\mathcal D_{k,s}}M_T\bigl(\theta_{P_k,s}(D);D\bigr).
\end{equation}
Here $k$ indexes the model widths, $P_k$ is the corresponding parameter count, and $s$ indexes the two seeds. The collection $\mathcal D_{k,s}$ contains the eight random-label datasets used at width $k$ and seed $s$, and $\theta_{P_k,s}(D)$ denotes the weights obtained after training the corresponding model on $D$. An affine least-squares fit of these six width-level values against parameter count estimates the capacity slope and intercept. The $95\%$ interval uses $20{,}000$ bootstrap draws, resampling widths with replacement and the two seed labels jointly across sampled widths; degenerate draws with only one distinct width are discarded. This uncertainty is conditional on the tested grid and the two seeds; it does not quantify sensitivity to other optimisers or larger widths.

The affine fit gives $C_{\mathrm{model}}=0.784933$ bits per parameter and $b_{\mathrm{cap}}=625$ bits; the intercept is about $12.8\%$ of the smallest measured maximum. Constraining the fit through the origin gives $0.830358$ bits per parameter. The fit has $R^2=0.999$ and a $95\%$ bootstrap interval $[0.766,0.825]$ for the slope. We use the affine slope as a normalisation throughout. Each observed maximum is an achieved storage value on a finite dataset-size grid, not an architecture-wide capacity upper bound.

\begin{figure}[!htbp]
    \centering
    \begin{subfigure}{0.48\textwidth}
        \centering
        \includegraphics[width=\linewidth]{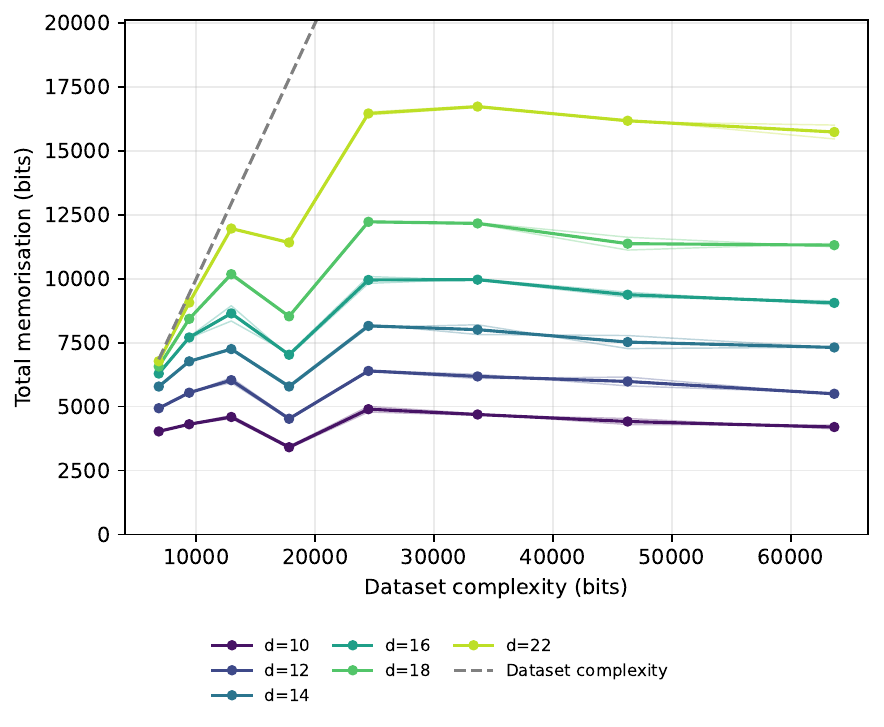}
        \caption{Total memorisation against dataset complexity.}
        \label{fig:capacity-curves}
    \end{subfigure}\hfill
    \begin{subfigure}{0.48\textwidth}
        \centering
        \includegraphics[width=\linewidth]{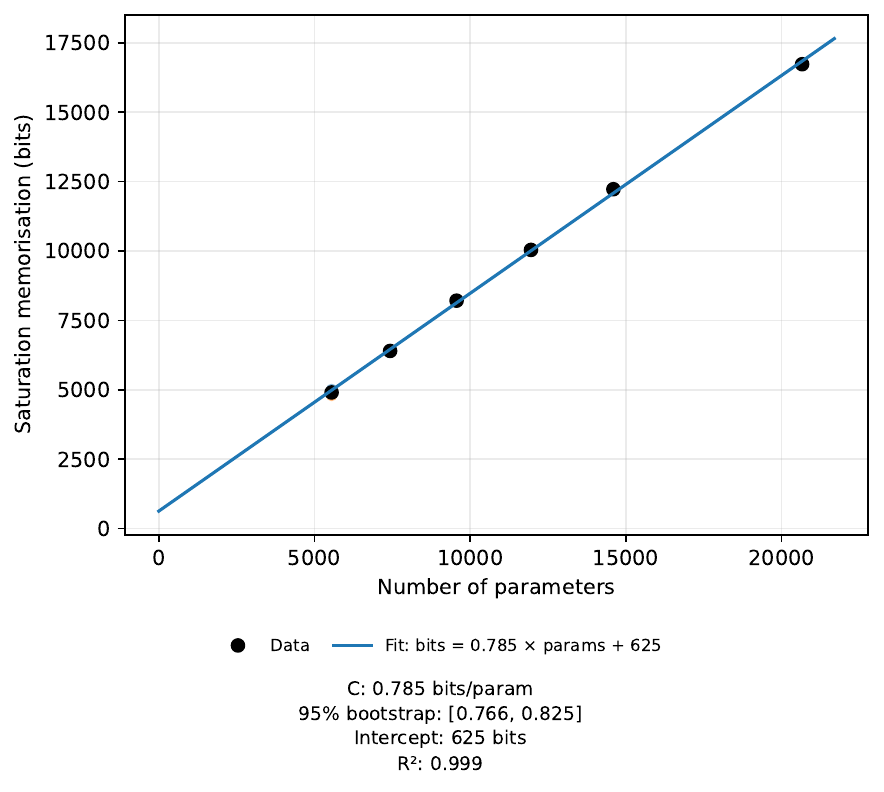}
        \caption{Saturation memorisation against parameter count.}
        \label{fig:capacity-estimation}
    \end{subfigure}
    \caption{Random-label capacity experiments following \citet{morris2025memorize}. \cref{fig:capacity-curves} shows total memorisation $M_T$ against dataset complexity on random-label datasets for several model sizes. The dashed grey line shows the dataset complexity in bits. \cref{fig:capacity-estimation} plots the mean of the two seed-specific maximum storage values over the tested dataset sizes against parameter count. The slope of the linear fit yields the bits-per-parameter constant $C_{\text{model}}$ for our Transformer family.}
    \label{fig:capacity-subfigs}
\end{figure}

\Cref{fig:capacity-subfigs} shows total memorisation as dataset size increases and the resulting fit against parameter count. For small datasets, storage grows approximately as $n\log_2V$, consistent with memorising nearly every label. The maxima then scale approximately linearly with parameter count, as in earlier capacity measurements \citep{roberts2020much,lu2024scaling,allen2024physics,morris2025memorize}.

\subsection{Measuring memorisation timescales on random data}
\label{sec:exp-speed}
For each prime $p$ in the 10-prime sweep (\cref{sec:experiments}) and each of the $44$ model widths $d\in\{20,24,\ldots,128\}\cup\{136,144,\ldots,256\}$, we generate a random-label dataset of size $n_{\text{equiv}}(p,\alpha) = K_{\text{mem}}(p,\alpha)/\log_2 V$. We instantiate a Transformer of width $d$ and train it as in \cref{sec:exp-arch} (dropout $0.2$, weight decay $1.0$), recording training accuracy at each epoch. We declare saturation once the running training accuracy, accumulated over the epoch's minibatches with dropout active, first reaches $99\%$, and report $T_{\text{mem}}(P,p)$ as the arithmetic mean of observed saturation epochs from ten planned seeds (42--51), excluding censored times. The convention matters: scoring training accuracy in evaluation mode after each epoch instead gives a timescale two to fourteen times faster on identical cells, fastest at narrow width; the reported random-label timescales use the former convention. Ancillary Tobit and intervention analyses are labelled separately from the primary cell-mean fits.

The full-range descriptive fit in \cref{fig:speed-summary} uses $335$ cells and $3{,}324$ observed A100 runs, excluding $1{,}076$ censored runs, and gives $T_{\text{mem}}(P,p)\approx12.068\exp(11.109f)$ with log-space $R^2=0.934$. Both timescales use arithmetic means of observed event times within each cell. A separate law is fitted on the selected capacity-fraction window (\cref{sec:theory-crossing}); neither fit establishes that the true timescale ends at the largest uncensored fraction.

\subsection{Grokking experiments on modular division}
\label{sec:exp-grokking}

For each prime $p$ in the 10-prime sweep and each model dimension $d \in \{20, 22, 24, \ldots, 128\} \cup \{130, 140, 150, \ldots, 1000\}$ (restricted to $d\le256$ for the main onset/intersection comparison), we construct a modular-division dataset $(X_{\text{train}}, T_{\text{train}}, X_{\text{test}}, T_{\text{test}})$ with training fraction $\alpha=0.5$ and a random train--test split. We instantiate a Transformer of width $d$ and train it as in \cref{sec:exp-arch} (dropout $0.2$, weight decay $1.0$), recording training and validation accuracies and losses at each epoch. We use 10 seeds per cell (seeds 42--51); $T_{\text{gen}}(P,p)$ is reported as the mean and $\Delta E(P)$ as the per-seed minimum, as defined in \cref{sec:theory-grokking}.

For the main-text examples in \cref{fig:delay-with-speed}, the delay scatter omits cells containing unresolved seed outcomes. The full cohort remains in the numerical analyses, and the timescale curves retain their observed-event means.

\paragraph{Measured and extrapolated timescale panels.}
The timescale panels in \cref{fig:cone} use the primary ten-seed A100 cohort for $p\le139$: memorisation is measured at training accuracy $99\%$ on random labels, and generalisation at validation accuracy $99\%$ on the modular task. Arithmetic means of observed events give $335$ memorisation cells and $679$ generalisation cells. All-censored cells and widths without measurements in this cohort are omitted. For $p>139$, the existing fitted laws are evaluated on the cone's sampled parameter grid.

\paragraph{Cone cohorts and incomplete outcomes.}\label{sec:exp-cone}
For \cref{fig:cone}, we retain the A100 central cohort and H100 extension
campaigns under the model and optimiser settings above, including
initialisation scale $1$, with a random split and the same learning rate for every parameter. Each $(p,d)$ cell uses one hardware cohort. Repeated observations
of a seed within that cohort count once: we select the largest planned epoch budget, then the
earliest run; all $45$ repeated observations have identical first-threshold
times. All $1{,}736$ cells have ten seeds ($42$--$51$), with
planned budgets of $1{,}500$, $2{,}500$, or $5{,}000$ epochs.
We score delay from the first time evaluation-mode validation accuracy reaches $98\%$,
relative to the first time online training accuracy reaches $99\%$. Reaching the validation
threshold while the training threshold remains unobserved establishes zero delay. If validation
has not reached its threshold, we retain a right-censored outcome: it is never counted as
successful delayed generalisation or as an observed zero. A cell is filled only when every retained
seed has an observed positive delay; any observed zero fixes its minimum at
zero; otherwise the cell is unresolved. Cells containing an unresolved seed
are omitted from the cone panel, including cells whose minimum is fixed at zero
by another seed. Minimum-delay colours use only complete positive cells. Restricting the
current cohort to the same five seeds ($42$--$46$) in every cell changes $76$
of $1{,}736$ classifications; the all-seed region
therefore depends on the number of seeds and is not a seed-independent phase
boundary.

The retained cohort includes $369$ runs with neither threshold observed and $1$ with training observed but validation unresolved; $89$ runs establish zero delay by reaching validation before an unobserved training threshold. In total, $13{,}407$ seed outcomes have positive delay, $3{,}583$ have zero delay, and $370$ remain unresolved. There are $1{,}102$ positive-minimum cells, $633$ zero-minimum cells, and $1$ unresolved-minimum cell. The $370$ unresolved seeds occur in $102$ cells: $101$ have a known zero minimum, while all ten seeds at $p=149,d=20$ remain unresolved.

For $24$ cells selected before the replication campaign (eight affected cells
and their nearest measured widths on either side), the initial replication collected $235$
fresh H100 runs with the same seed sets and a $5{,}000$-epoch cap. The archived
A100 observations are retained for the paired replication comparison,
separately from the plotted cell cohorts. Training stops after both
online training and evaluation-mode validation accuracies reach $99\%$,
or at the planned epoch cap. This targeted replication covers
$p=97$--$139$; it does not test the proposed cutoff at $p=197$ or refit the
primary central onset/intersection statistics.

The H100 replication agrees with the archived cell classification in $24$ of $24$ cells and with the individual-seed classification in $234$ of $235$ pairs. The same-five-seed restriction changes $76$ cell classifications in the current plotted cohort. All preselected comparisons are reported in \cref{tab:cone-replication}. A later independent sample agrees in $192/200$ paired seed states and $19/20$ cell classifications; at $p=127,d=74$ the minimum changes from one epoch on A100 to zero on H100. These replications cover central cells and do not establish hardware identity or validate a high-prime cutoff.

\paragraph{Ten-seed completion.} Before collecting the additional runs, we
fixed all $269$ incomplete cells. We added seeds $47$--$51$ to $239$ H100
cells and replaced $30$ five-seed A100 cells with H100 seeds $42$--$51$,
retaining all $150$ replaced observations for paired comparison. These
$1{,}495$ runs used both H100s, with the same model, optimiser, precision, and
per-cell epoch budgets; every planned run is included and every new
checkpoint was independently evaluated on CPU. The $1{,}467$ previously
complete cells are unchanged. The final figure contains $1{,}407$ A100 cells
and $329$ H100 cells, with no hardware pooling within a cell.
Completion changes the minimum delay in $53$ cells; $12$ change from positive
to zero minimum. The $p=149,d=96$ cell has six positive and four zero delays (\cref{sec:theory-prereg}). The fitted laws use arithmetic cell means; the primary measured-curve test uses a selected ten-seed A100 cohort, separately from the expanded cone. The original paired replication table retains its
matched seed sets; the plotted $p=139,d=224$ cell now has ten H100 seeds.

\paragraph{Extension to $p=199$.} Before collecting this column, we fixed
nineteen widths: the original eighteen-width $p=197$ grid and $d=316$,
with ten seeds ($42$--$51$) at every width. The $190$ runs use the same
fixed-rate protocol on one H100, with a $5{,}000$-epoch cap and the scoring
and censoring rules above. This extension covers $d=52$--$556$ (approximately $10^5$--$10^7$
parameters). All planned runs are included; this extension itself left the
earlier columns and fitted laws unchanged. Every new checkpoint was independently evaluated on CPU.
Of the $19$ cells, $0$ have positive delay in all ten seeds, $19$ have at least one observed zero, and $0$ remain unresolved. Across all $190$ seeds, $30$ have observed positive delay, $157$ have zero delay, and $3$ have unresolved delay. These finite-grid outcomes do not establish disappearance at every width or in every seed.
Restricting the new column to the common five seeds ($42$--$46$) changes $1$ of its $19$ cell classifications.
At $d=364$, the minimum is $1$ epoch with the common five seeds and $0$ with all ten; $7$ of the ten seeds have positive delay.

\begin{table}[!htbp]
\centering
\small
\caption{Initial matched cone replication on one H100. Each row lists the affected
width in the middle and its two preselected neighbouring widths. Entries
are minimum observed delays (epochs) across the same seed IDs on each
machine; the three entries follow the listed width order. The archival
A100 measurements and fresh H100 measurements are compared separately.
The five-seed entry is the original paired comparison; the current cone
contains ten H100 seeds at that cell.}
\label{tab:cone-replication}
\begin{tabular}{rllll}
\toprule
Prime & Widths & Seeds per cell & A100 minima & H100 minima \\
\midrule
97 & 54, 56, 58 & 10 & 2, 2, 5 & 2, 2, 5 \\
107 & 62, 64, 66 & 10 & 0, 1, 2 & 0, 1, 2 \\
107 & 78, 80, 82 & 10 & 5, 4, 5 & 5, 4, 5 \\
113 & 62, 64, 66 & 10 & 2, 3, 2 & 2, 3, 2 \\
127 & 78, 80, 82 & 10 & 1, 4, 2 & 1, 4, 2 \\
139 & 94, 96, 98 & 10 & 1, 2, 1 & 1, 2, 1 \\
139 & 110, 112, 114 & 10 & 3, 2, 2 & 2, 4, 3 \\
139 & 220, 224, 230 & 10, 5, 10 & 12, 12, 9 & 8, 16, 8 \\
\bottomrule
\end{tabular}
\end{table}

\begin{figure}[!htbp]
    \centering
    \includegraphics[width=0.9\linewidth]{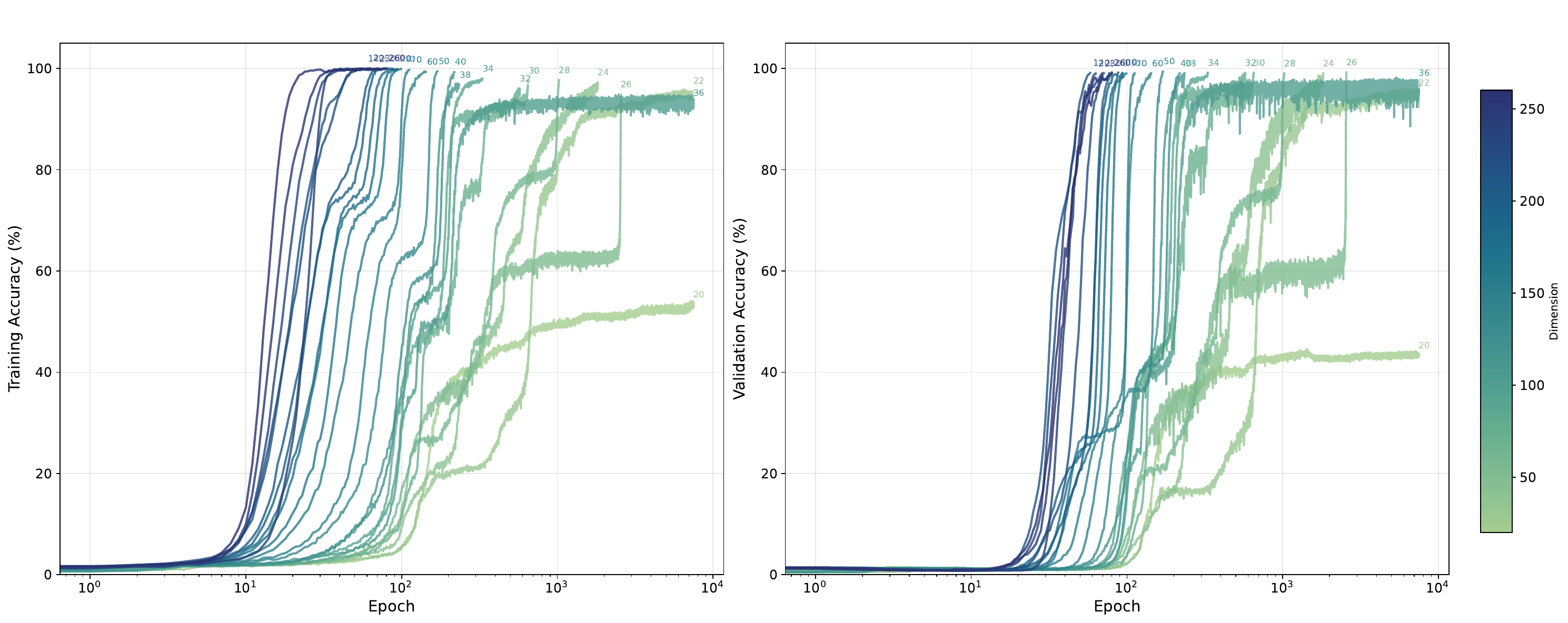}
    \caption{Training (left) and validation (right) accuracy curves for modular division with $p=127$ across model sizes.
    Small models underfit, intermediate models generalise immediately, and larger models exhibit grokking.}
    \label{fig:grokking-curves}
\end{figure}

\cref{fig:grokking-curves} shows representative training and validation curves for modular division at $p=127$ as we vary model size.

\subsection{Hyperparameter, architectural, and task sweeps}
\label{sec:exp-sweeps}
The secondary sweeps repeat the random-label and modular-task measurements while varying one hyperparameter, model depth, or task. Unless stated otherwise, they retain the training protocol in \cref{sec:exp-arch}, including the $5{,}000$-epoch cap. The task dataset size and the information-matched random dataset size are recomputed for each training fraction and operation. For a task with $N$ possible pairs, both datasets contain $\lfloor\alpha N\rfloor$ training examples. The nominal complexity used by the analysis is $K_{\mathrm{mem}}=\alpha N\log_2(p+2)$, before this integer rounding; its difference from the sampled random-label information is less than one label. We keep $C_{\mathrm{model}}=0.784933$ as a common normalisation; this does not assume that achieved capacity is unchanged under every intervention.

For the non-reference hyperparameter and task settings, random-label measurements use widths $\{20,24,28,32,40,48,56,64,80,96,112,128,160,200,256\}$. Modular-task measurements use $\{20,24,28,32\}\cup\{40,48,56,64\}\cup\{80,96,112,128\}\cup\{160,192,224,256\}$. The reference setting reuses the denser main-sweep grids. The depth sweep instead selects widths to approximate parameter-count targets $\{5{,}000,10{,}000,20{,}000,50{,}000,100{,}000,200{,}000,500{,}000\}$, using the actual parameter count in the analysis.

\paragraph{Seed coverage and hardware.}
Each comparison retains a common pool of four or five seeds at every setting, as specified in \cref{sec:hparam-invariance,sec:arch-invariance,sec:task-invariance}. Measurements missing a required seed are excluded before analysing outcomes; a completed run that has not reached an accuracy threshold remains in the cohort as a censored observation. Some comparisons use H100 runs alongside an A100 reference. The hardware replications in \cref{sec:exp-cone} support approximate reproducibility, but do not eliminate this possible source of differences between settings.

\section{Quantitative tests of the intersection hypothesis}
\label{sec:stats-methods}

We turn the qualitative claim ``the intersection of $T_{\text{mem}}$ and $T_{\text{gen}}$ predicts the empirical grokking onset'' into three falsifiable sub-claims and report a test for each. The paired comparisons operate on $\{(\widehat{P}_{\text{cross}}(p),\,P_{\text{onset}}(p))\}_{p}$ in $\log_{10}$ coordinates, where $\widehat{P}_{\text{cross}}(p)$ is the lower intersection of the seed-averaged $T_{\text{mem}}$ and $T_{\text{gen}}$ curves, and $P_{\text{onset}}(p)$ is defined in \cref{sec:theory-grokking}.

\paragraph{Constructing the timescale curves.}
A cell contains runs with the same prime, model size, and training settings, but different random seeds. For each timescale separately, we record the first epoch at which the relevant accuracy reaches $99\%$: training accuracy on random-label data for $T_{\mathrm{mem}}$, and validation accuracy on the modular task for $T_{\mathrm{gen}}$. At each cell, we add the observed event times and divide by the number of seeds that reached the threshold. These arithmetic means form the timescale curves across model sizes. Runs that do not reach the threshold within their budget are excluded from the average; their epoch caps are not substituted as event times. A partially observed cell contributes its observed seeds, while an all-censored cell has no timescale mean. The averages therefore describe completion within the training budget. When fitting the empirical laws, we take logarithms after averaging and give each contributing cell equal weight. This timescale average is distinct from the minimum paired delay across seeds used to define consistent grokking.

A statistical observation is one pair of parameter counts: the measured intersection and the first grokking onset for a given prime and setting. Both must be identified to enter the point statistics. The coverage tables also count cases where either cannot be determined; individual seeds and model widths are not separate observations in these tests. An intersection must be an actual decreasing root of the piecewise-linear log-time difference in log parameter count. An absent root is not replaced by the nearest approach. The primary ten-prime test and the global parametric-law comparison are reported separately.

\paragraph{Onset domain and unresolved outcomes.}
The onset is the first measured size with an identified positive minimum delay across the stated seed pool. Every smaller measured minimum must be identified and zero. If any smaller minimum could still be positive, onset remains unresolved, even when a larger model is known to grok in every seed. A known zero-delay seed fixes the minimum at zero even if other seed outcomes remain unknown. An unresolved minimum is never assigned zero delay or counted as resolved. Later zero or unresolved minima do not invalidate an already identified first onset. If every measured minimum is zero, no onset is observed; if no positive minimum is identified and some minima are unknown, onset remains unresolved. A positive minimum at the narrowest measured width is left-censored and excluded from point statistics. The main comparison uses $d\le256$; the five-seed dense study uses its full sampled grid without trimming wider models (\cref{sec:exp-ood}).

All ten main-sweep onsets are identified under this rule. Secondary sweeps use their stated four- or five-seed pools; unresolved earlier minima reduce the number of identifiable onsets. A zero minimum in a larger seed pool can be determined by a seed absent from a smaller pool, so secondary reference coverage need not match the ten-seed main sweep. The final row of \cref{tab:threshold-sensitivity} compares the current definition with a sustained-positive-tail criterion, requiring positive minimum delay at every larger measured size.

\paragraph{Predictiveness.}
We test for any monotone relationship between predicted and empirical onsets via Spearman's $\rho$ (with both an analytic $p$-value and a $10^4$-shuffle permutation $p$-value, the latter to guard against optimistic analytic tails with few cases) and Kendall's $\tau$ (different small-sample tail behaviour to Spearman). The main-sweep values are $\rho=0.988$ and $\tau=0.956$.

\paragraph{Calibration.}
Rank correlation alone does not commit to predicted and empirical coordinates being equal: a strongly correlated predictor with the wrong slope or a constant offset can pass predictiveness but fail calibration. We test calibration with three statistics:
\begin{itemize}
    \item Lin's concordance correlation coefficient $\text{CCC}=2r\sigma_p\sigma_e/(\sigma_p^2+\sigma_e^2+(\mu_p-\mu_e)^2)$, where $r$ is the Pearson correlation of the log coordinates, $\mu_p,\sigma_p$ are the mean and standard deviation of the predicted coordinates, and $\mu_e,\sigma_e$ those of the empirical coordinates (the subscript $p$ here means predicted, not prime). This is a one-number summary of ``predicted equals empirical'' since it penalises both decorrelation and offset from $y=x$. We report a $95\%$ percentile interval from $10^4$ resamples of these onset/intersection pairs, excluding degenerate draws for which CCC is undefined.
    \item OLS fit $\log_{10}P_{\text{onset}}=a_{\mathrm{cal}}+b_{\mathrm{cal}}\log_{10}\widehat{P}_{\text{cross}}$, where $a_{\mathrm{cal}}$ and $b_{\mathrm{cal}}$ are the calibration intercept and slope, with a joint $F$-test for $(a_{\mathrm{cal}}=0,b_{\mathrm{cal}}=1)$. Reporting $(\hat a_{\mathrm{cal}},\hat b_{\mathrm{cal}})$ separately makes the direction of a departure from identity legible; the slope and intercept must be interpreted jointly over the observed range.
    \item Wilcoxon signed-rank on the log-residuals $\log_{10}(P_{\text{onset}}/\widehat{P}_{\text{cross}})$ against zero, a rank-based test of the location of the residual distribution under its symmetry assumption.
\end{itemize}
On the main sweep, Lin's CCC is $0.699$ with interval $[0.468,0.763]$. The fitted slope is $\hat b_{\mathrm{cal}}=0.948$ and intercept $\hat a_{\mathrm{cal}}=0.108$, but the joint $F$-test rejects $(0,1)$ ($F=1007.40$, $p=2.45\times10^{-10}$). The Wilcoxon test also rejects zero location (median log-residual $-0.170$, $p=0.002$). The OLS intercept and the median are not in tension: the residual model gives $\log_{10}(P_{\text{onset}}/\widehat{P}_{\text{cross}})\approx0.108-0.052\log_{10}\widehat{P}_{\text{cross}}$, or about $-0.15$ at $\log_{10}\widehat{P}_{\text{cross}}=5$. A slope near one does not by itself establish absence of proportional bias. The median offset corresponds to onset about $32\%$ below the prediction, or a prediction about $48\%$ above onset.

\paragraph{Incremental information over baselines.}
We operationalise the comparison with simpler predictors using nested OLS models for $\log_{10}P_{\text{onset}}$:
\begin{description}
    \item[$M_0$ (constant):] the same fitted log onset for every observation.
    \item[$M_1$ (intersection):] a linear fit to $\log_{10}\widehat{P}_{\text{cross}}$, with a fitted constant and slope.
    \item[$M_2$ (baseline):] a fitted constant plus prime $p$ in the main sweep, or prime and the varied hyperparameter, depth, or task in secondary sweeps.
    \item[$M_3$ (combined):] a fitted constant and all predictors from $M_1$ and $M_2$.
\end{description}
$M_1$ vs.\ $M_0$ tests whether the intersection has signal. $M_3$ vs.\ $M_2$ tests whether the intersection adds information over and above the baseline, including the possibility that onset and intersection are correlated because both depend on $p$. $M_3$ vs.\ $M_1$ asks whether the baseline adds information beyond the intersection. Numeric baseline variables enter on their stated raw scale; task categories are dummy coded. Constant predictors are omitted and degrees of freedom use design-matrix rank.

The main-sweep intersection adds information over prime ($F=155.79$, $p=4.88\times10^{-6}$) and over the capacity-based predictor $\log_{10}P_{\mathrm{mem}}$, with $P_{\mathrm{mem}}=K_{\mathrm{mem}}/C_{\mathrm{model}}$ ($F=170.35$, $p=3.61\times10^{-6}$). Prime adds no detectable improvement over the intersection in this sample ($F=0.816$, $p=0.396$). With ten observations, a nonsignificant improvement does not establish that the additional predictor contains no useful information. Dropout is constant and cannot be tested here.

\paragraph{Robustness.}
The supplementary robustness analysis regresses the log onset/prediction ratio against each varied axis and Holm-correct the resulting $p$-values across axes. The intent is to flag systematic mispredictions along any axis (e.g.\ residuals trending with weight decay or depth). Correlation and calibration summaries are conditional on identified pairs and do not score missing predictions as successes.

\paragraph{Uncertainty and scope.}
The main comparison contains ten observations, one per prime. The nested-model tests assess fit to those observations; they do not use cross-validation or estimate prediction error on new primes. Permutations shuffle observed onsets across the identified pairs, and the CCC bootstrap resamples pairs without clustering by prime or seed. These procedures treat the pairs as exchangeable. Reusing primes, seeds, or reference measurements across secondary settings can violate that assumption, so pooled significance levels should be interpreted cautiously. The regression $F$-tests additionally assume independent, constant-variance Gaussian residuals conditional on the predictors; their measurement uncertainty is not modelled. Rank and CCC summaries require at least three pairs; calibration and nested regression tests require at least four, with positive residual degrees of freedom. The Wilcoxon test requires at least six nonzero residuals. The tabulated tests are two-sided where applicable and are not corrected across the suite of comparisons; only the separate axis-residual tests use the stated Holm correction.

There is also dependence between the measured predictor and outcome. For each task run, $T_{\mathrm{gen}}$ uses the first validation-$99\%$ epoch, while delay uses the first validation-$98\%$ epoch from the same accuracy history, together with the training-$99\%$ epoch. Measuring $T_{\mathrm{mem}}$ on random labels does not remove that dependence. These comparisons therefore assess an empirical diagnostic, rather than an independent forecast of task learning. The fitted-law extrapolations in \cref{sec:exp-ood} are a separate test.

Intervals describe their specified resampling unit: onset/intersection pairs for CCC, prime clusters for the fitted-law fold, and paired seed labels within each model/prime combination for dense measured intersections. They do not recover unobserved completion times. Finite training budgets, unequal width grids, and exclusion of unresolved onsets restrict the conclusions to the cases that could be measured. Failure to reject a null hypothesis with these small samples does not establish equality or absence of additional predictive information.

\begin{table}[!htbp]
\centering\small
\caption{Sensitivity on the same ten-prime primary cohort. Timescale means exclude unobserved events. All variants refind actual intersections and retain unresolved delays. The last row instead requires positive minimum delay at every larger measured size. Median residual is $\log_{10}(P_{\mathrm{onset}}/\widehat{P}_{\mathrm{cross}})$.}
\label{tab:threshold-sensitivity}
\begin{tabular}{lcccc}
\toprule
Train / gen / delay-val (\%) & Delay statistic & Pairs / 10 & Median residual & $\rho$ \\
\midrule
99 / 99 / 98 & min & 10/10 & $-0.170$ & $0.988$ \\
95 / 95 / 95 & min & 10/10 & $-0.199$ & $0.988$ \\
98 / 98 / 98 & min & 10/10 & $-0.249$ & $0.988$ \\
99 / 99 / 99 & min & 9/10 & $-0.308$ & $0.967$ \\
99 / 99 / 98 & median & 0/10 & --- & --- \\
99 / 99 / 98 & mean & 0/10 & --- & --- \\
99 / 99 / 98 & min; sustained tail & 10/10 & $-0.160$ & $0.976$ \\
\bottomrule
\end{tabular}
\end{table}

Under matched $99\%$ thresholds, nine of ten first onsets remain identifiable. Using seed-median or seed-mean delay leaves all ten first onsets unresolved because smaller-width aggregates depend on unobserved seed outcomes; these variants therefore have no point statistics in \cref{tab:threshold-sensitivity}. This does not imply absence of positive delay.

\paragraph{Reproducible cohort selection.}
The primary measured-curve cohort uses the ten calibration primes and seeds 42--51 from one A100 host. Secondary analyses use each sweep's declared prime set and a common seed pool at every setting: seeds 42--45 for initialisation, learning rate, training fraction, depth, and task comparisons; seeds 42--46 for weight decay. The otherwise matched central reference is selected explicitly from A100 data. Cells lacking the declared seed pool are omitted by coverage before observing outcomes; censoring within retained cells is preserved. No seeds are filled using results from another hardware cohort. The exact UUIDs, coverage exclusions, raw events, and complete point tables are recorded with the analysis.

\clearpage
\begingroup
\renewenvironment{figure}[1][]{\par\addvspace{\intextsep}\noindent\begin{minipage}{\linewidth}\captionsetup{type=figure}}{\end{minipage}\par\addvspace{\intextsep}}
\renewenvironment{table}[1][]{\par\addvspace{\intextsep}\noindent\begin{minipage}{\linewidth}\captionsetup{type=table}}{\end{minipage}\par\addvspace{\intextsep}}

\section{Robustness across hyperparameters}
\label{sec:hparam-invariance}

\paragraph{Scope and reading guide.}
We test whether the association between timescale intersection and grokking onset persists when a training hyperparameter changes. A setting means one value of that hyperparameter, such as weight decay $0.3$. An identified pair means that both the intersection and the first onset can be located for one prime at that setting, using \cref{sec:stats-methods}.

The pooled tables combine all identified pairs in a sweep. The within-setting tables hold the hyperparameter fixed and compare primes, showing both the number of usable pairs and the total number tested. Pooled correlation does not establish correlation at every setting. We also report $\log_{10}(P_{\mathrm{onset}}/\widehat P_{\mathrm{cross}})$, the log-ratio of observed to predicted onset parameter count: zero means agreement, and a negative value means the intersection overestimates onset. Variation in this offset indicates that quantitative predictions need recalibration. These tables describe the observed variation; they do not establish that adjusting one constant is sufficient at every setting.

\subsection{Initialisation scale}
\label{sec:hparam-init-scale}

\paragraph{Scope.} We vary initialisation scale over $\{0.5,1,2\}$, using five primes $\{97,107,113,127,139\}$ and four seeds (42--45) per retained cell. Non-reference settings use H100 runs, and the reference uses A100 runs. The hardware comparison in \cref{sec:exp-cone} supports approximate reproducibility, but does not remove this difference between settings. Timescale means exclude censored times; onset uses minimum paired delay and the fixed $d\le256$ domain. The reference uses the denser main-sweep width grid (\cref{sec:exp-sweeps}). We retain $C_{\mathrm{model}}=0.784933$ as a common normalisation.

\begin{figure}[!htbp]
    \centering
    \begin{subfigure}{0.48\textwidth}
        \centering
        \includegraphics[width=\linewidth]{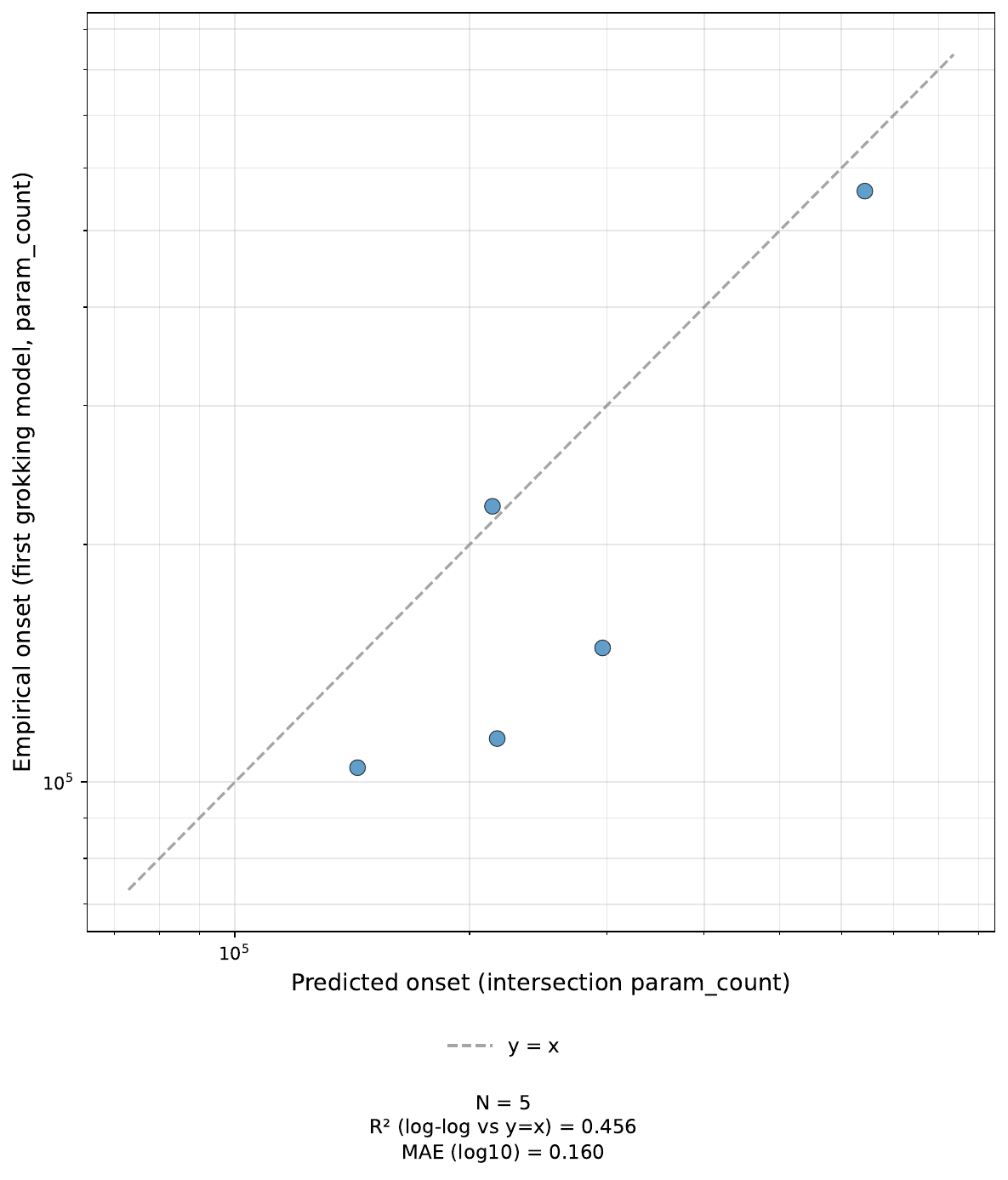}
        \caption{Predicted vs.\ empirical onset for identified prime/setting pairs.}
        \label{fig:init-scale-pred-vs-emp}
    \end{subfigure}\hfill
    \begin{subfigure}{0.48\textwidth}
        \centering
        \includegraphics[width=\linewidth]{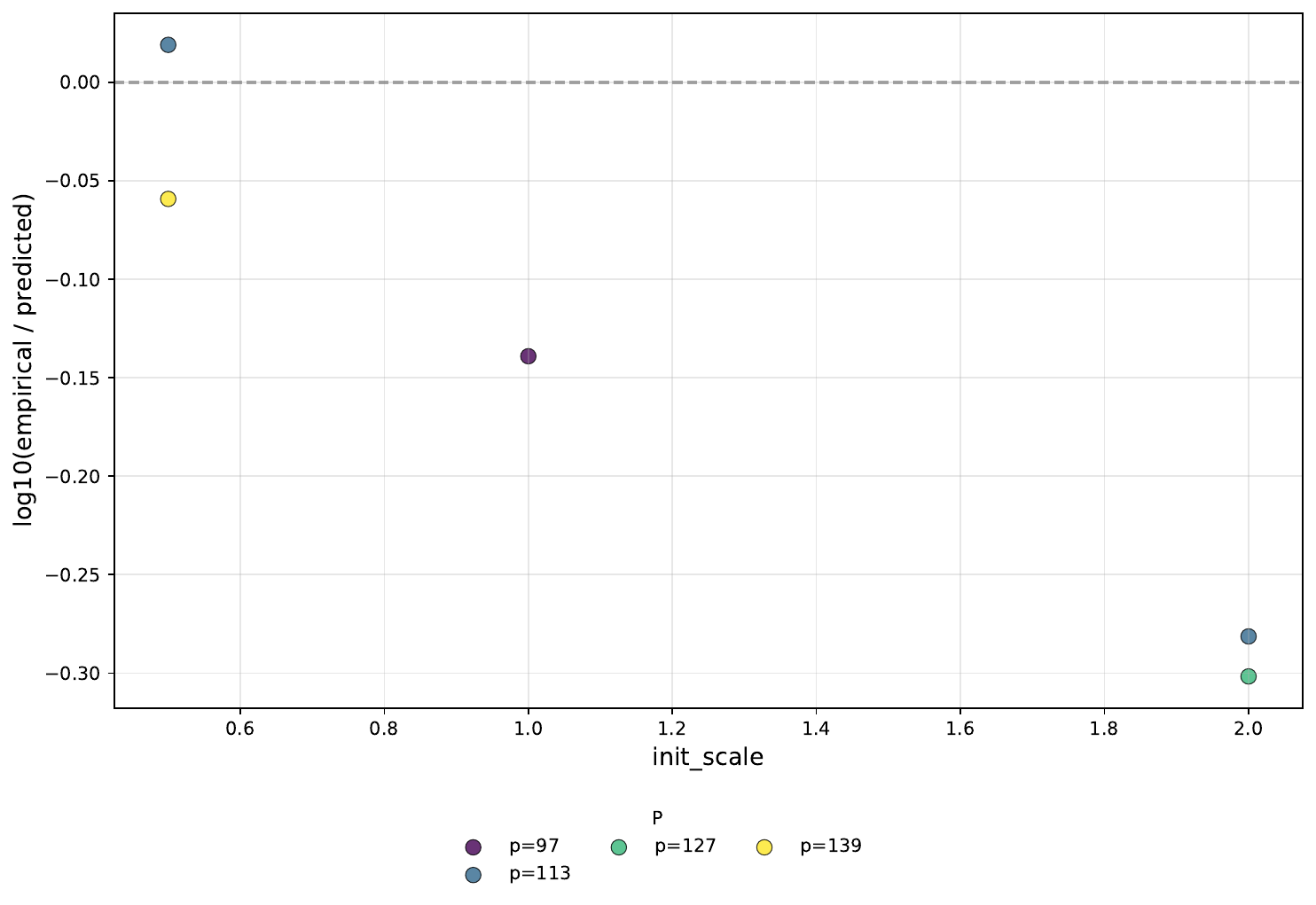}
        \caption{Log onset/prediction ratio against initialisation scale.}
        \label{fig:init-scale-error}
    \end{subfigure}
    \caption{Initialisation-scale sweep: predicted-vs-empirical scatter and residual trend with the swept axis.}
    \label{fig:init-scale-overview}
\end{figure}

\begin{table}[!htbp]
\centering\small
\caption{Initialisation scale: association and agreement between the measured timescale intersection and observed grokking onset, using 5 pairs from 15 tested combinations of prime and setting. A pair contributes only when both parameter counts can be determined. The regression models predict log onset: $M_0$ predicts the same value for every observation, $M_1$ uses log parameter count at the intersection, $M_2$ uses prime and the varied setting, and $M_3$ uses both sets of predictors. Each comparison asks whether the added predictors improve the fit. The last row compares with $P_{\mathrm{mem}}=K_{\mathrm{mem}}/C_{\mathrm{model}}$, the capacity-based estimate. CCC measures agreement in values; the calibration test asks whether predicted and observed values coincide. Median residual is $\log_{10}(P_{\mathrm{onset}}/\widehat P_{\mathrm{cross}})$. Dashes indicate omitted or unestimable tests. Resampling, test assumptions, and sample-size rules are given in \cref{sec:stats-methods}; the model comparisons use fits to the observed sample.}
\label{tab:init-scale-pooled}
\begin{tabular}{lll}
\toprule
Test & Statistic & $p$-value \\
\midrule
Spearman $\rho$ & $0.700$ & $0.235$ \\
Kendall $\tau$ & $0.600$ & $0.233$ \\
Lin's CCC (95\% interval) & $0.727\;[-0.125,0.964]$ & --- \\
Log--log slope & $1.086$ & --- \\
Intercept & $-0.618$ & --- \\
Calibration $F$ test & $2.34$ & $0.244$ \\
Median log-residual; Wilcoxon & $-0.139$ & --- \\
$M_1$ vs. $M_0$ & $11.13$ & $0.045$ \\
$M_3$ vs. $M_2$ & $1.33$ & $0.455$ \\
$M_3$ vs. $M_1$ & $21.69$ & $0.150$ \\
Intersection adds over $P_{\mathrm{mem}}$ & $2.89$ & $0.231$ \\
\bottomrule
\end{tabular}
\end{table}

\begin{table}[!htbp]
\centering\small
\caption{Initialisation scale: coverage and agreement at each setting. The numerator counts primes for which both onset and intersection can be determined; the denominator counts all tested primes. The median and sample standard deviation (SD) summarise $\log_{10}(P_{\mathrm{onset}}/\widehat P_{\mathrm{cross}})$: negative values mean the intersection overestimates onset. SD requires two pairs, and rank correlation requires three. Seed pools, hardware, and reference measurements are specified in the text.}
\label{tab:init-scale-within}
\begin{tabular}{lcccc}
\toprule
Setting & Pairs / eligible & Median log-ratio & SD & Spearman $\rho$ \\
\midrule
0.5 & 2/5 & $-0.020$ & $0.055$ & --- \\
1.0 & 1/5 & $-0.139$ & --- & --- \\
2.0 & 2/5 & $-0.292$ & $0.014$ & --- \\
\bottomrule
\end{tabular}
\end{table}

\paragraph{Results.} Only 5 of 15 eligible cases have an identified onset and intersection. Their pooled correlation is $\rho=0.700$ (permutation $p=0.235$), and each individual setting has only one or two pairs. These data are insufficient to establish within-setting predictiveness or a stable calibration. Failure to reject equality of predicted and observed parameter counts ($p=0.244$) does not establish agreement; the sample is small and the CCC interval is wide.

\subsection{Weight decay}
\label{sec:hparam-weight-decay}

\paragraph{Scope.} We vary weight decay over $\{0,0.01,0.03,0.1,0.3,1,3\}$, using six primes $\{97,107,113,127,139,149\}$ and five seeds (42--46) at every setting, including the central reference. All selected settings use the same A100 host. Timescale means exclude censored times; onset uses minimum paired delay and the fixed $d\le256$ domain. The reference uses the denser main-sweep width grid (\cref{sec:exp-sweeps}). We retain $C_{\mathrm{model}}=0.784933$ as a common normalisation.

\begin{figure}[!htbp]
    \centering
    \begin{subfigure}{0.48\textwidth}
        \centering
        \includegraphics[width=\linewidth]{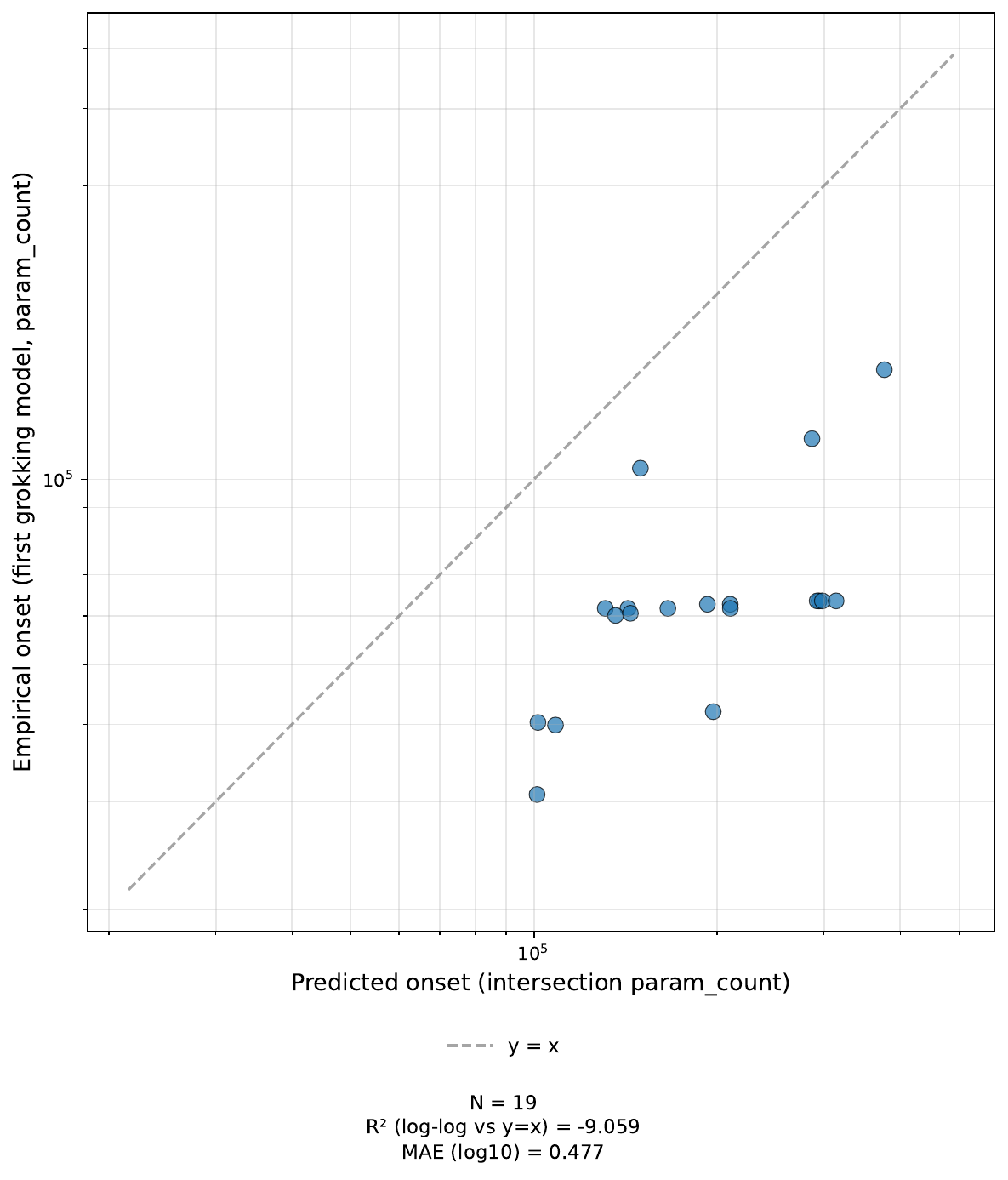}
        \caption{Predicted vs.\ empirical onset for identified prime/setting pairs.}
        \label{fig:wd-pred-vs-emp}
    \end{subfigure}\hfill
    \begin{subfigure}{0.48\textwidth}
        \centering
        \includegraphics[width=\linewidth]{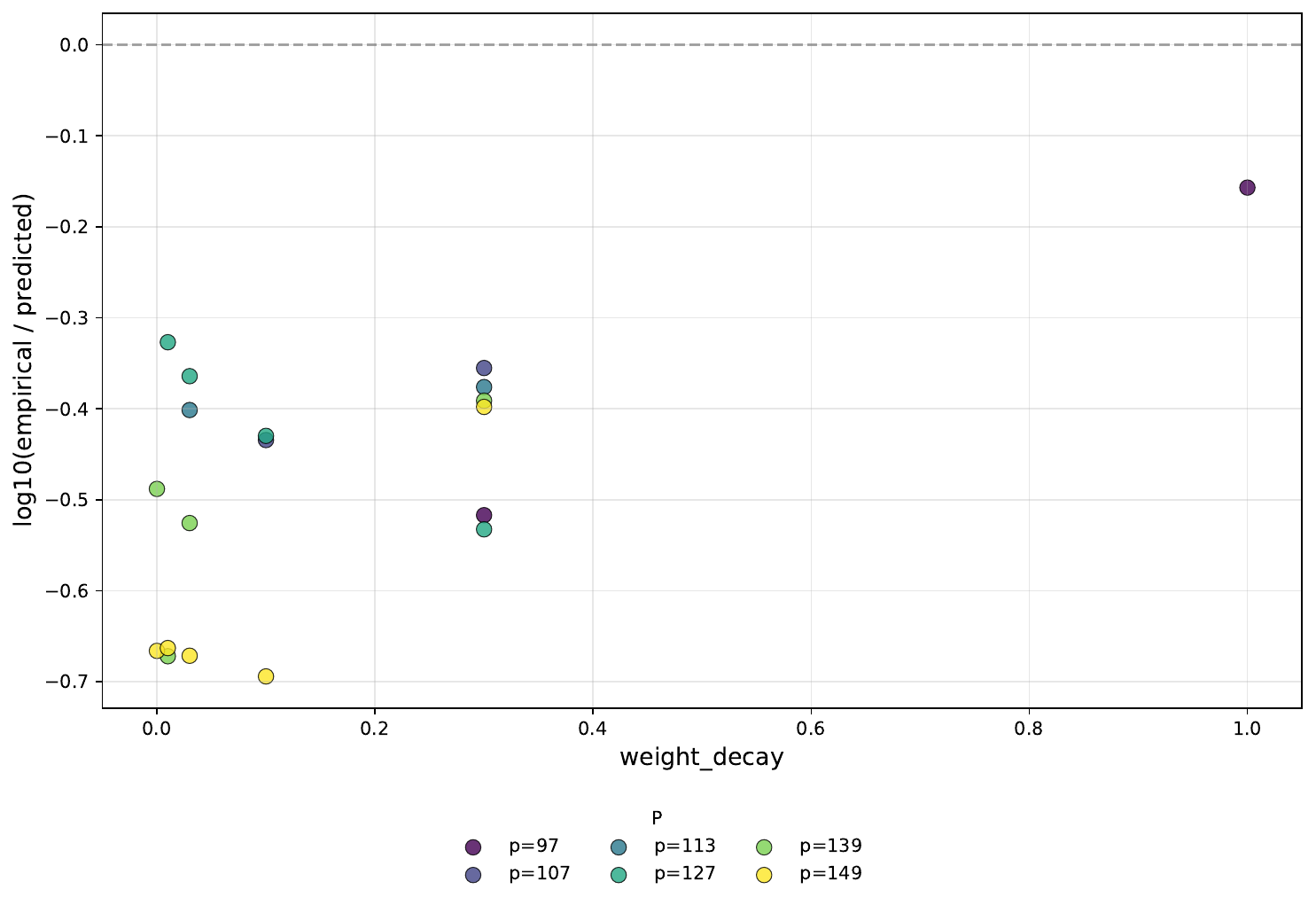}
        \caption{Log onset/prediction ratio against weight decay.}
        \label{fig:wd-error}
    \end{subfigure}
    \caption{Weight-decay sweep: predicted-vs-empirical scatter and residual trend with the swept axis.}
    \label{fig:wd-overview}
\end{figure}

\begin{table}[!htbp]
\centering\small
\caption{Weight decay: association and agreement between the measured timescale intersection and observed grokking onset, using 19 pairs from 42 tested combinations of prime and setting. A pair contributes only when both parameter counts can be determined. The regression models predict log onset: $M_0$ predicts the same value for every observation, $M_1$ uses log parameter count at the intersection, $M_2$ uses prime and the varied setting, and $M_3$ uses both sets of predictors. Each comparison asks whether the added predictors improve the fit. The last row compares with $P_{\mathrm{mem}}=K_{\mathrm{mem}}/C_{\mathrm{model}}$, the capacity-based estimate. CCC measures agreement in values; the calibration test asks whether predicted and observed values coincide. Median residual is $\log_{10}(P_{\mathrm{onset}}/\widehat P_{\mathrm{cross}})$. Dashes indicate omitted or unestimable tests. Resampling, test assumptions, and sample-size rules are given in \cref{sec:stats-methods}; the model comparisons use fits to the observed sample.}
\label{tab:wd-pooled}
\begin{tabular}{lll}
\toprule
Test & Statistic & $p$-value \\
\midrule
Spearman $\rho$ & $0.806$ & $2.00\times10^{-4}$ \\
Kendall $\tau$ & $0.657$ & $1.33\times10^{-4}$ \\
Lin's CCC (95\% interval) & $0.123\;[0.025,0.227]$ & --- \\
Log--log slope & $0.570$ & --- \\
Intercept & $1.788$ & --- \\
Calibration $F$ test & $134.63$ & $3.77\times10^{-11}$ \\
Median log-residual; Wilcoxon & $-0.434$ & $3.81\times10^{-6}$ \\
$M_1$ vs. $M_0$ & $11.46$ & $0.004$ \\
$M_3$ vs. $M_2$ & $0.47$ & $0.504$ \\
$M_3$ vs. $M_1$ & $8.57$ & $0.003$ \\
Intersection adds over $P_{\mathrm{mem}}$ & $13.15$ & $0.002$ \\
\bottomrule
\end{tabular}
\end{table}

\begin{table}[!htbp]
\centering\small
\caption{Weight decay: coverage and agreement at each setting. The numerator counts primes for which both onset and intersection can be determined; the denominator counts all tested primes. The median and sample standard deviation (SD) summarise $\log_{10}(P_{\mathrm{onset}}/\widehat P_{\mathrm{cross}})$: negative values mean the intersection overestimates onset. SD requires two pairs, and rank correlation requires three. Seed pools, hardware, and reference measurements are specified in the text.}
\label{tab:wd-within}
\begin{tabular}{lcccc}
\toprule
Setting & Pairs / eligible & Median log-ratio & SD & Spearman $\rho$ \\
\midrule
0.0 & 2/6 & $-0.577$ & $0.126$ & --- \\
0.01 & 3/6 & $-0.663$ & $0.197$ & $0.500$ \\
0.03 & 4/6 & $-0.464$ & $0.139$ & $1.000$ \\
0.1 & 3/6 & $-0.434$ & $0.151$ & $1.000$ \\
0.3 & 6/6 & $-0.395$ & $0.076$ & $1.000$ \\
1.0 & 1/6 & $-0.157$ & --- & --- \\
3.0 & 0/6 & --- & --- & --- \\
\bottomrule
\end{tabular}
\end{table}

\paragraph{Results.} There are 19 identified pairs among 42 eligible cases, with pooled $\rho=0.806$ (permutation $p=0.0002$). Quantitative agreement is poor (CCC $0.123$), and equality of predicted and observed parameter counts is rejected. Decay $0.3$ has six pairs with $\rho=1$; other settings have zero to four pairs, including only two at zero decay and none at decay $3$. The intersection does not improve a model using prime and weight decay in this sample ($p=0.504$). The pooled association therefore does not establish predictiveness at every decay setting.

\subsection{Learning rate}
\label{sec:hparam-lr}

\paragraph{Scope.} We vary learning rate over $\{10^{-4},3\times10^{-4},10^{-3},3\times10^{-3},10^{-2}\}$, using five primes $\{97,107,113,127,139\}$ and four seeds (42--45) per retained cell. Non-reference settings use H100 runs, and the reference uses A100 runs. The hardware comparison in \cref{sec:exp-cone} supports approximate reproducibility, but does not remove this difference between settings. Timescale means exclude censored times; onset uses minimum paired delay and the fixed $d\le256$ domain. The reference uses the denser main-sweep width grid (\cref{sec:exp-sweeps}). We retain $C_{\mathrm{model}}=0.784933$ as a common normalisation.

\begin{figure}[!htbp]
    \centering
    \begin{subfigure}{0.48\textwidth}
        \centering
        \includegraphics[width=\linewidth]{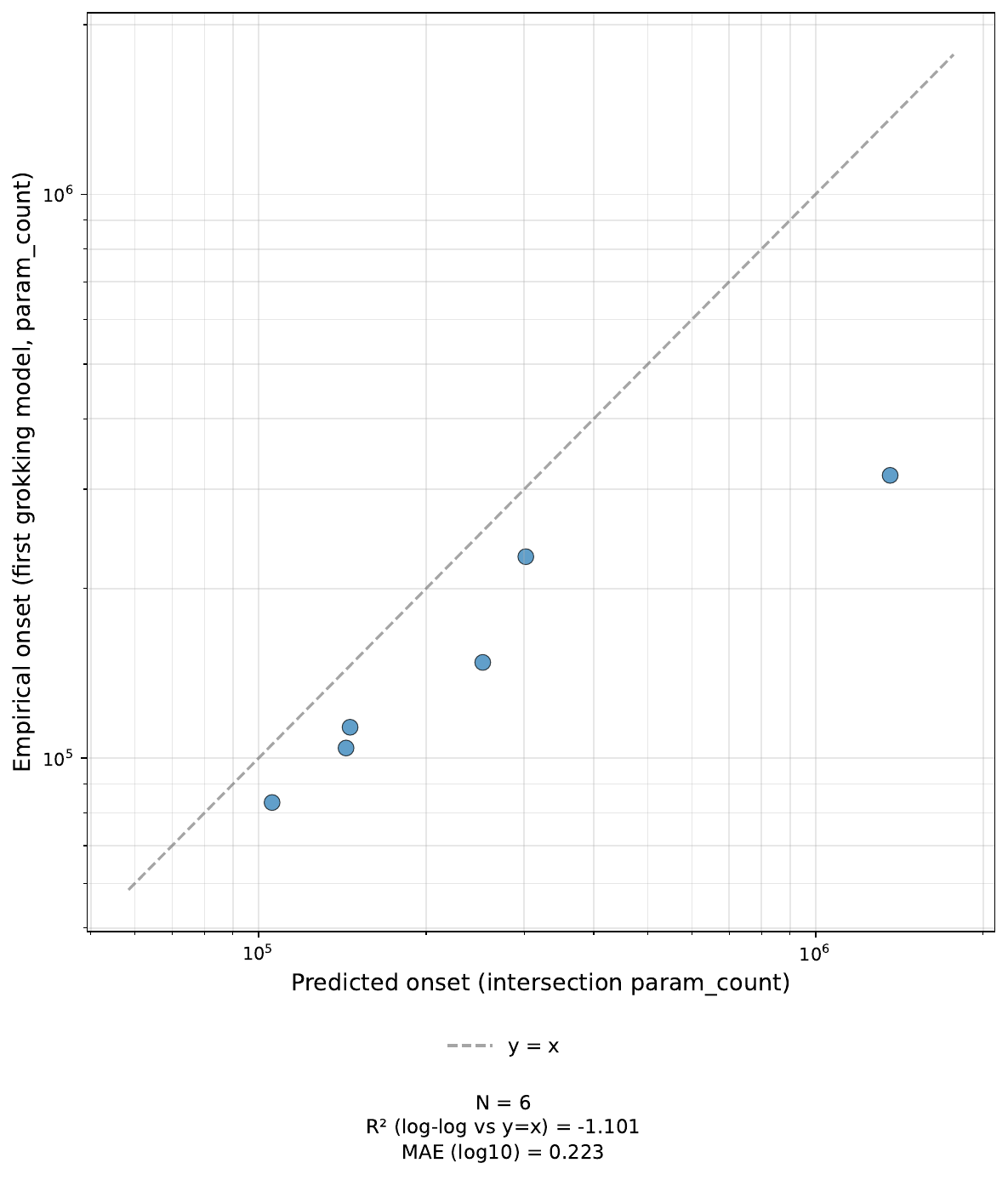}
        \caption{Predicted vs.\ empirical onset for identified prime/setting pairs.}
        \label{fig:lr-pred-vs-emp}
    \end{subfigure}\hfill
    \begin{subfigure}{0.48\textwidth}
        \centering
        \includegraphics[width=\linewidth]{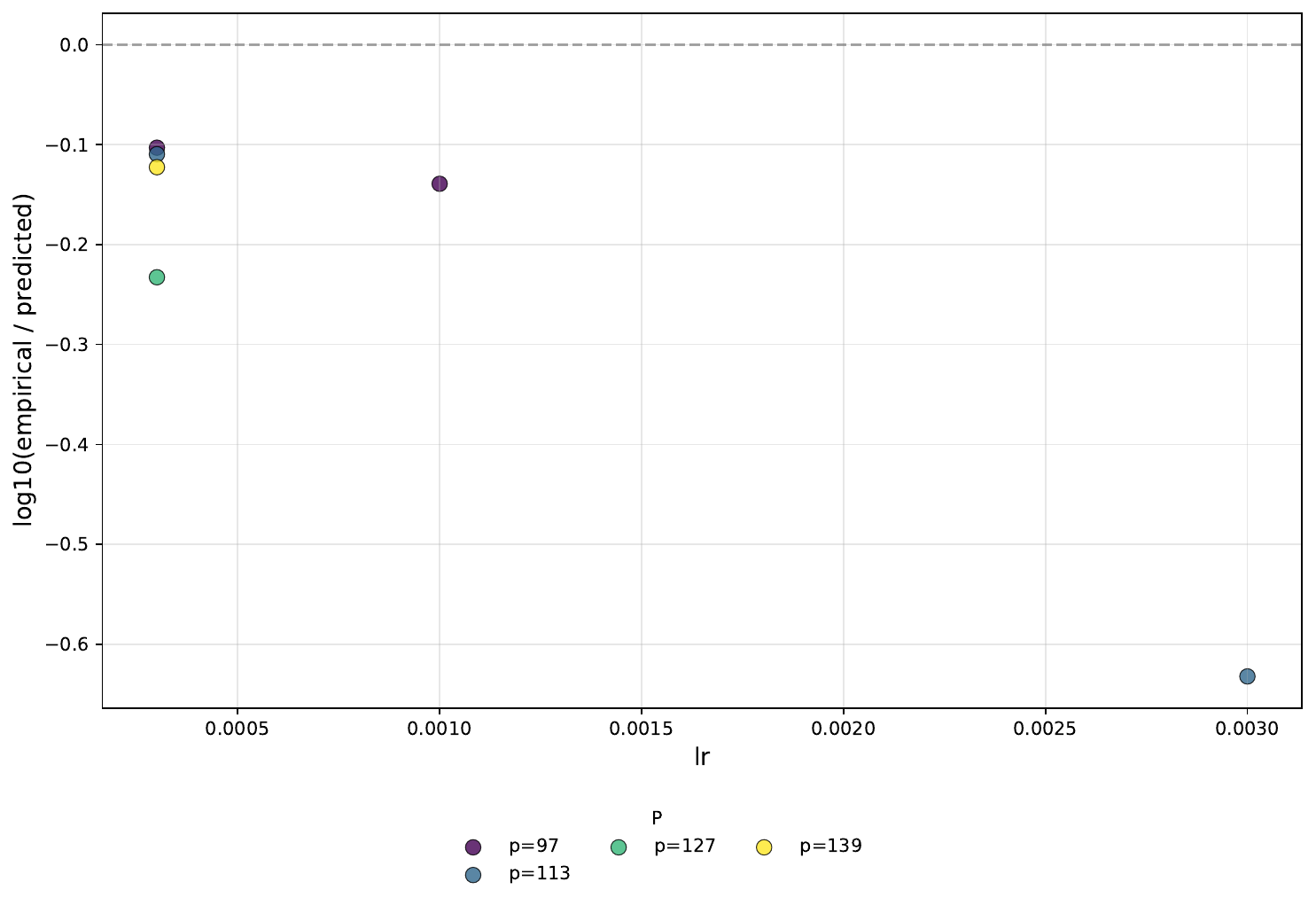}
        \caption{Log onset/prediction ratio against learning rate.}
        \label{fig:lr-error}
    \end{subfigure}
    \caption{Learning-rate sweep: predicted-vs-empirical scatter and residual trend with the swept axis.}
    \label{fig:lr-overview}
\end{figure}

\begin{table}[!htbp]
\centering\small
\caption{Learning rate: association and agreement between the measured timescale intersection and observed grokking onset, using 6 pairs from 25 tested combinations of prime and setting. A pair contributes only when both parameter counts can be determined. The regression models predict log onset: $M_0$ predicts the same value for every observation, $M_1$ uses log parameter count at the intersection, $M_2$ uses prime and the varied setting, and $M_3$ uses both sets of predictors. Each comparison asks whether the added predictors improve the fit. The last row compares with $P_{\mathrm{mem}}=K_{\mathrm{mem}}/C_{\mathrm{model}}$, the capacity-based estimate. CCC measures agreement in values; the calibration test asks whether predicted and observed values coincide. Median residual is $\log_{10}(P_{\mathrm{onset}}/\widehat P_{\mathrm{cross}})$. Dashes indicate omitted or unestimable tests. Resampling, test assumptions, and sample-size rules are given in \cref{sec:stats-methods}; the model comparisons use fits to the observed sample.}
\label{tab:lr-pooled}
\begin{tabular}{lll}
\toprule
Test & Statistic & $p$-value \\
\midrule
Spearman $\rho$ & $1.000$ & $0.002$ \\
Kendall $\tau$ & $1.000$ & $0.003$ \\
Lin's CCC (95\% interval) & $0.619\;[0.309,0.809]$ & --- \\
Log--log slope & $0.519$ & --- \\
Intercept & $2.369$ & --- \\
Calibration $F$ test & $35.52$ & $0.003$ \\
Median log-residual; Wilcoxon & $-0.131$ & $0.031$ \\
$M_1$ vs. $M_0$ & $31.71$ & $0.005$ \\
$M_3$ vs. $M_2$ & $0.23$ & $0.681$ \\
$M_3$ vs. $M_1$ & $10.46$ & $0.087$ \\
Intersection adds over $P_{\mathrm{mem}}$ & $59.24$ & $0.005$ \\
\bottomrule
\end{tabular}
\end{table}

\begin{table}[!htbp]
\centering\small
\caption{Learning rate: coverage and agreement at each setting. The numerator counts primes for which both onset and intersection can be determined; the denominator counts all tested primes. The median and sample standard deviation (SD) summarise $\log_{10}(P_{\mathrm{onset}}/\widehat P_{\mathrm{cross}})$: negative values mean the intersection overestimates onset. SD requires two pairs, and rank correlation requires three. Seed pools, hardware, and reference measurements are specified in the text.}
\label{tab:lr-within}
\begin{tabular}{lcccc}
\toprule
Setting & Pairs / eligible & Median log-ratio & SD & Spearman $\rho$ \\
\midrule
0.0001 & 0/5 & --- & --- & --- \\
0.0003 & 4/5 & $-0.116$ & $0.061$ & $1.000$ \\
0.001 & 1/5 & $-0.139$ & --- & --- \\
0.003 & 1/5 & $-0.632$ & --- & --- \\
0.01 & 0/5 & --- & --- & --- \\
\bottomrule
\end{tabular}
\end{table}

\paragraph{Results.} There are 6 identified pairs among 25 eligible cases, with pooled $\rho=1$ (permutation $p=0.0023$). Four pairs come from learning rate $0.0003$, one from $0.001$, and one from $0.003$; rates $0.0001$ and $0.01$ contribute none. Unresolved smaller minima, as well as missing intersections, limit coverage. Equality of predicted and observed parameter counts is rejected, and adding the intersection to prime and learning rate gives no detectable improvement ($p=0.681$). These results do not establish robustness across the full learning-rate range.

\subsection{Training fraction}
\label{sec:hparam-train-fraction}

\paragraph{Scope.} We vary training fraction over $\{0.3,0.4,0.5,0.6,0.7\}$, using five primes $\{97,107,113,127,139\}$ and four seeds (42--45) at every setting. All selected settings use the same A100 host. Timescale means exclude censored times; onset uses minimum paired delay and the fixed $d\le256$ domain. The reference uses the denser main-sweep width grid (\cref{sec:exp-sweeps}). We retain $C_{\mathrm{model}}=0.784933$ as a common normalisation.

\begin{figure}[!htbp]
    \centering
    \begin{subfigure}{0.48\textwidth}
        \centering
        \includegraphics[width=\linewidth]{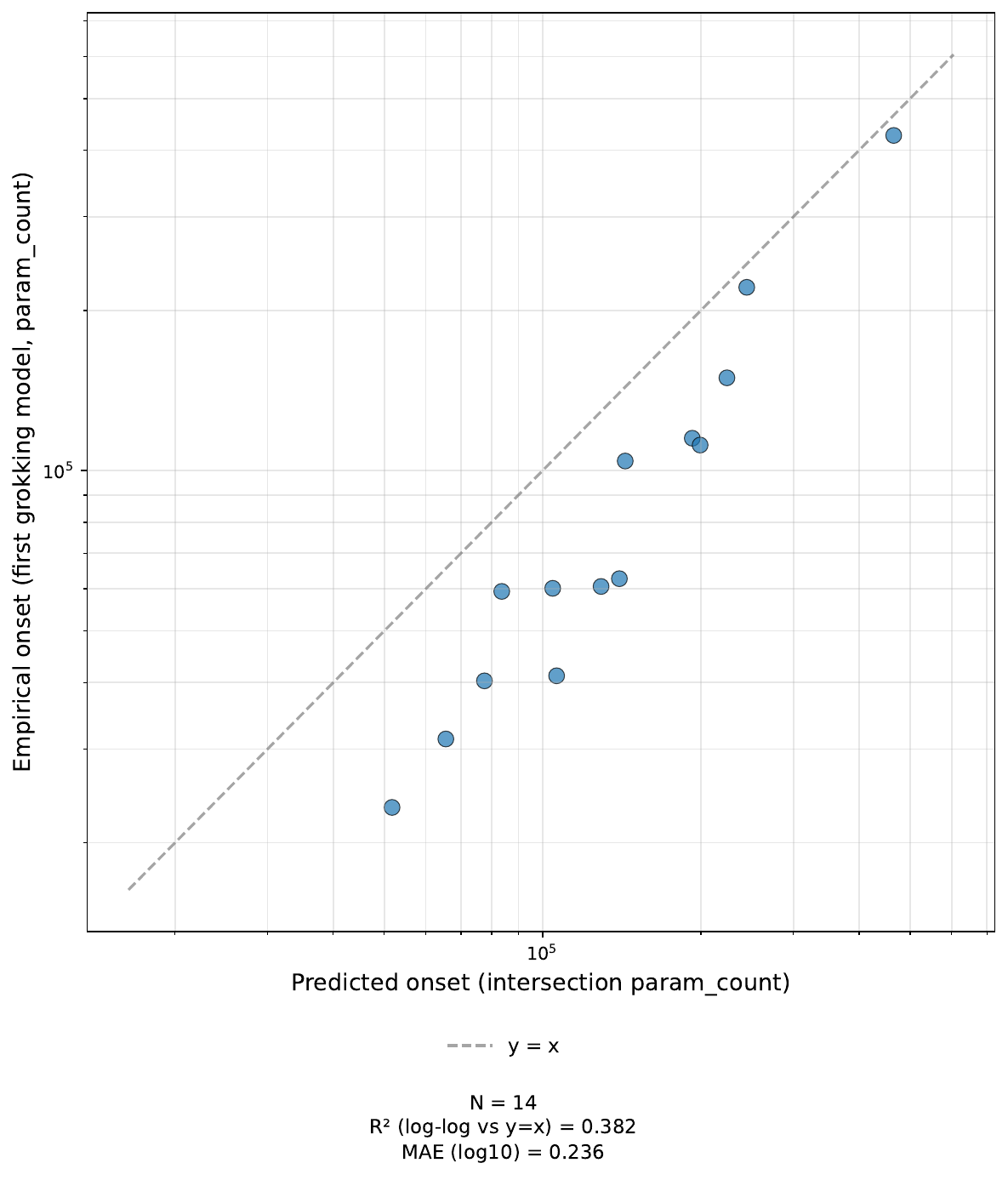}
        \caption{Predicted vs.\ empirical onset for identified prime/setting pairs.}
        \label{fig:alpha-pred-vs-emp}
    \end{subfigure}\hfill
    \begin{subfigure}{0.48\textwidth}
        \centering
        \includegraphics[width=\linewidth]{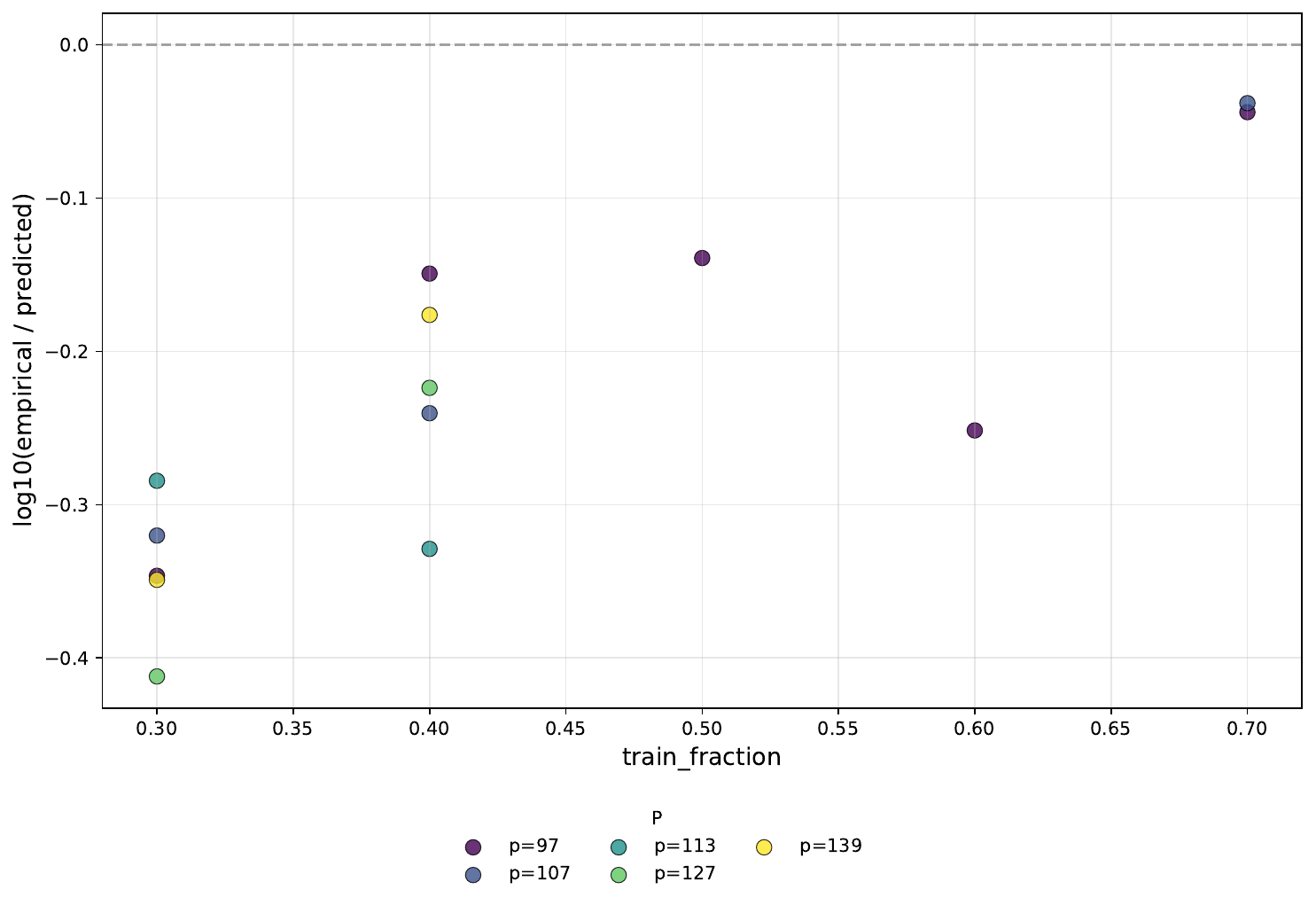}
        \caption{Log onset/prediction ratio against training fraction.}
        \label{fig:alpha-error}
    \end{subfigure}
    \caption{Training-fraction sweep: predicted-vs-empirical scatter and residual trend with the swept axis.}
    \label{fig:alpha-overview}
\end{figure}

\begin{table}[!htbp]
\centering\small
\caption{Training fraction: association and agreement between the measured timescale intersection and observed grokking onset, using 14 pairs from 25 tested combinations of prime and setting. A pair contributes only when both parameter counts can be determined. The regression models predict log onset: $M_0$ predicts the same value for every observation, $M_1$ uses log parameter count at the intersection, $M_2$ uses prime and the varied setting, and $M_3$ uses both sets of predictors. Each comparison asks whether the added predictors improve the fit. The last row compares with $P_{\mathrm{mem}}=K_{\mathrm{mem}}/C_{\mathrm{model}}$, the capacity-based estimate. CCC measures agreement in values; the calibration test asks whether predicted and observed values coincide. Median residual is $\log_{10}(P_{\mathrm{onset}}/\widehat P_{\mathrm{cross}})$. Dashes indicate omitted or unestimable tests. Resampling, test assumptions, and sample-size rules are given in \cref{sec:stats-methods}; the model comparisons use fits to the observed sample.}
\label{tab:alpha-pooled}
\begin{tabular}{lll}
\toprule
Test & Statistic & $p$-value \\
\midrule
Spearman $\rho$ & $0.982$ & $10.00\times10^{-5}$ \\
Kendall $\tau$ & $0.934$ & $1.25\times10^{-8}$ \\
Lin's CCC (95\% interval) & $0.700\;[0.406,0.830]$ & --- \\
Log--log slope & $1.299$ & --- \\
Intercept & $-1.771$ & --- \\
Calibration $F$ test & $54.37$ & $9.64\times10^{-7}$ \\
Median log-residual; Wilcoxon & $-0.246$ & $1.22\times10^{-4}$ \\
$M_1$ vs. $M_0$ & $183.26$ & $1.25\times10^{-8}$ \\
$M_3$ vs. $M_2$ & $5.82$ & $0.037$ \\
$M_3$ vs. $M_1$ & $3.76$ & $0.061$ \\
Intersection adds over $P_{\mathrm{mem}}$ & $62.74$ & $7.18\times10^{-6}$ \\
\bottomrule
\end{tabular}
\end{table}

\begin{table}[!htbp]
\centering\small
\caption{Training fraction: coverage and agreement at each setting. The numerator counts primes for which both onset and intersection can be determined; the denominator counts all tested primes. The median and sample standard deviation (SD) summarise $\log_{10}(P_{\mathrm{onset}}/\widehat P_{\mathrm{cross}})$: negative values mean the intersection overestimates onset. SD requires two pairs, and rank correlation requires three. Seed pools, hardware, and reference measurements are specified in the text.}
\label{tab:alpha-within}
\begin{tabular}{lcccc}
\toprule
Setting & Pairs / eligible & Median log-ratio & SD & Spearman $\rho$ \\
\midrule
0.3 & 5/5 & $-0.346$ & $0.047$ & $1.000$ \\
0.4 & 5/5 & $-0.224$ & $0.069$ & $1.000$ \\
0.5 & 1/5 & $-0.139$ & --- & --- \\
0.6 & 1/5 & $-0.252$ & --- & --- \\
0.7 & 2/5 & $-0.041$ & $0.004$ & --- \\
\bottomrule
\end{tabular}
\end{table}

\paragraph{Results.} There are 14 identified pairs among 25 eligible cases, with pooled $\rho=0.982$ (permutation $p=10^{-4}$). Fractions $0.3$ and $0.4$ each have five pairs and $\rho=1$, while the other fractions have only one or two pairs. Equality of predicted and observed parameter counts is rejected. The intersection improves the model using prime and training fraction ($p=0.037$), but sparse coverage at the larger fractions limits conclusions about calibration throughout the sweep.

\subsection{Untested hyperparameters}
\label{sec:hparam-future}
Dropout remains fixed in these comparisons. Its effect, and behaviour between or beyond the sampled learning rates, require further experiments.

\section{Robustness across model depths}
\label{sec:arch-invariance}

\paragraph{Scope.} We vary depth while retaining a single attention head and gated feed-forward blocks. The parameter-count targets in \cref{sec:exp-sweeps} allow comparisons at approximately matched total model size. As in the hyperparameter sweeps, we distinguish pooled association from association across primes at a fixed depth.

\subsection{Depth scaling}
\label{sec:arch-depth}

\paragraph{Scope.} We vary depth over $\{2,4,6,8,10\}$ at the parameter-count targets in \cref{sec:exp-sweeps}, using five primes $\{97,107,113,127,139\}$ and four seeds (42--45) per retained cell. Sparse observed timescale curves limit intersection identification at larger depths. Non-reference settings use H100 runs, and the reference uses A100 runs. The hardware comparison in \cref{sec:exp-cone} supports approximate reproducibility, but does not remove this difference between settings. Timescale means exclude censored times; onset uses minimum paired delay and the fixed $d\le256$ domain. The reference uses the denser main-sweep width grid (\cref{sec:exp-sweeps}). We retain $C_{\mathrm{model}}=0.784933$ as a common normalisation.

\begin{figure}[!htbp]
    \centering
    \begin{subfigure}{0.48\textwidth}
        \centering
        \includegraphics[width=\linewidth]{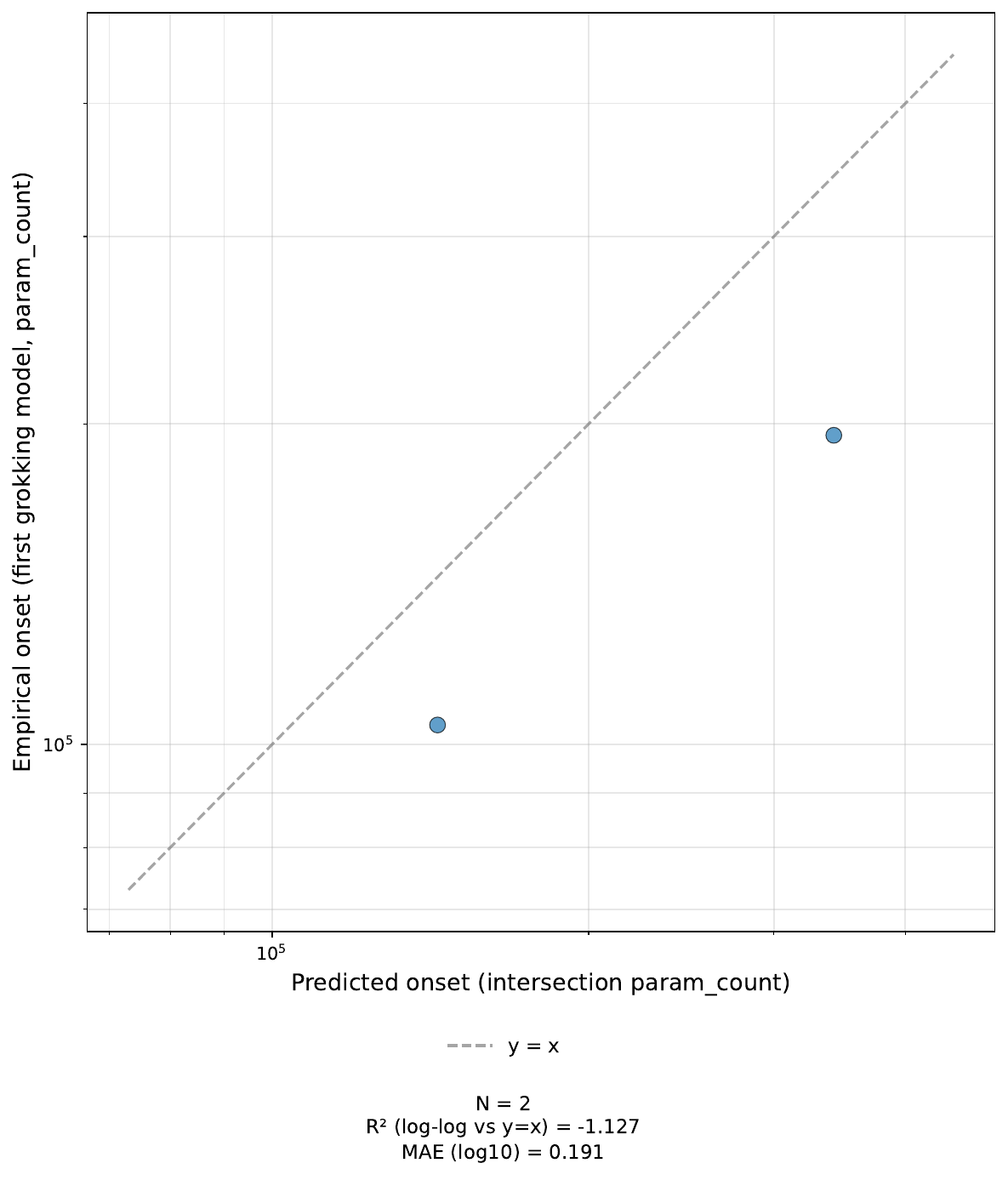}
        \caption{Predicted vs.\ empirical onset for identified prime/setting pairs.}
        \label{fig:depth-pred-vs-emp}
    \end{subfigure}\hfill
    \begin{subfigure}{0.48\textwidth}
        \centering
        \includegraphics[width=\linewidth]{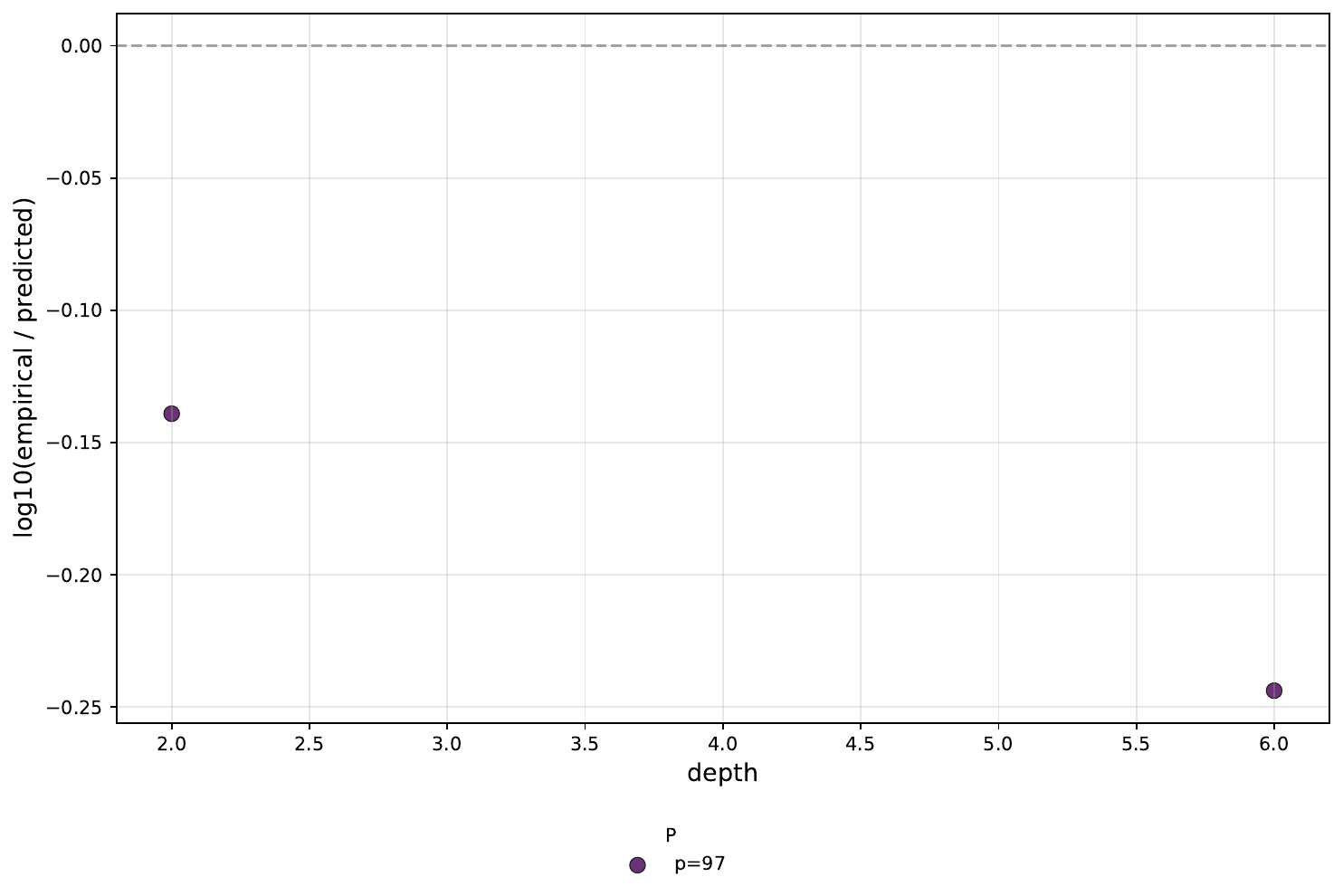}
        \caption{Log onset/prediction ratio against depth.}
        \label{fig:depth-error}
    \end{subfigure}
    \caption{Depth-scaling sweep: predicted-vs-empirical scatter and residual trend with the swept axis.}
    \label{fig:depth-overview}
\end{figure}

\begin{table}[!htbp]
\centering\small
\caption{Depth: only two combinations of prime and depth have both a measured timescale intersection and an identified first grokking onset. Correlation, calibration, and regression tests are omitted because the sample is too small. Coverage at each depth is shown in \cref{tab:depth-within}; unresolved-outcome rules are given in \cref{sec:stats-methods}.}
\label{tab:depth-pooled}
\begin{tabular}{lr}
\toprule
Quantity & Count \\
\midrule
Tested combinations of prime and depth & 25 \\
Both onset and intersection identified & 2 \\
\bottomrule
\end{tabular}
\end{table}

\begin{table}[!htbp]
\centering\small
\caption{Depth: coverage and agreement at each setting. The numerator counts primes for which both onset and intersection can be determined; the denominator counts all tested primes. The median and sample standard deviation (SD) summarise $\log_{10}(P_{\mathrm{onset}}/\widehat P_{\mathrm{cross}})$: negative values mean the intersection overestimates onset. SD requires two pairs, and rank correlation requires three. Seed pools, hardware, and reference measurements are specified in the text.}
\label{tab:depth-within}
\begin{tabular}{lcccc}
\toprule
Setting & Pairs / eligible & Median log-ratio & SD & Spearman $\rho$ \\
\midrule
2 & 1/5 & $-0.139$ & --- & --- \\
4 & 0/5 & --- & --- & --- \\
6 & 1/5 & $-0.244$ & --- & --- \\
8 & 0/5 & --- & --- & --- \\
10 & 0/5 & --- & --- & --- \\
\bottomrule
\end{tabular}
\end{table}

\paragraph{Results.} Only 2 of 25 eligible cases have both an identified first onset and a measured intersection: one at depth two and one at depth six. Correlation and regression tests are therefore omitted. Sixteen cases lack a measured intersection, and 23 have unresolved first onsets; these categories overlap. Larger models can have positive delay while an unresolved smaller minimum prevents locating the first onset. The available comparisons do not establish predictiveness or calibration across depth.

\subsection{Untested architectural variations}
\label{sec:arch-future}
The number of attention heads, feed-forward gating, normalisation operation, and positional encoding are not varied here. Testing these choices, together with MLP baselines, would be needed to assess transfer beyond this Transformer family. In particular, the present one-head implementation has no effective RoPE rotation (\cref{sec:exp-arch}).

\section{Robustness across tasks}
\label{sec:task-invariance}

\paragraph{Scope.} We repeat the timescale and grokking measurements for modular addition and multiplication, comparing each with division. The within-task tables assess association across primes; pooled tables combine the task with its division reference. The same division observations appear in both comparisons, so these are not independent replications.

\subsection{Modular addition}
\label{sec:task-modular-add}

\paragraph{Scope.} We compare modular addition with division, using five primes $\{97,107,113,127,139\}$ and four seeds (42--45) per retained cell. Non-reference settings use H100 runs, and the reference uses A100 runs. The hardware comparison in \cref{sec:exp-cone} supports approximate reproducibility, but does not remove this difference between settings. Timescale means exclude censored times; onset uses minimum paired delay and the fixed $d\le256$ domain. The reference uses the denser main-sweep width grid (\cref{sec:exp-sweeps}). We retain $C_{\mathrm{model}}=0.784933$ as a common normalisation.

For addition and multiplication, the dataset enumerates all $p^2$ pairs (rather than $p(p-1)$ for $/$), so $K_{\text{mem}}(p,\alpha)$ and $n_{\text{equiv}}(p,\alpha)$ are recomputed accordingly. The $\circ=/$ cells reuse the main-sweep timescale and grokking runs and serve as an in-sweep baseline.

\begin{figure}[!htbp]
    \centering
    \begin{subfigure}{0.48\textwidth}
        \centering
        \includegraphics[width=\linewidth]{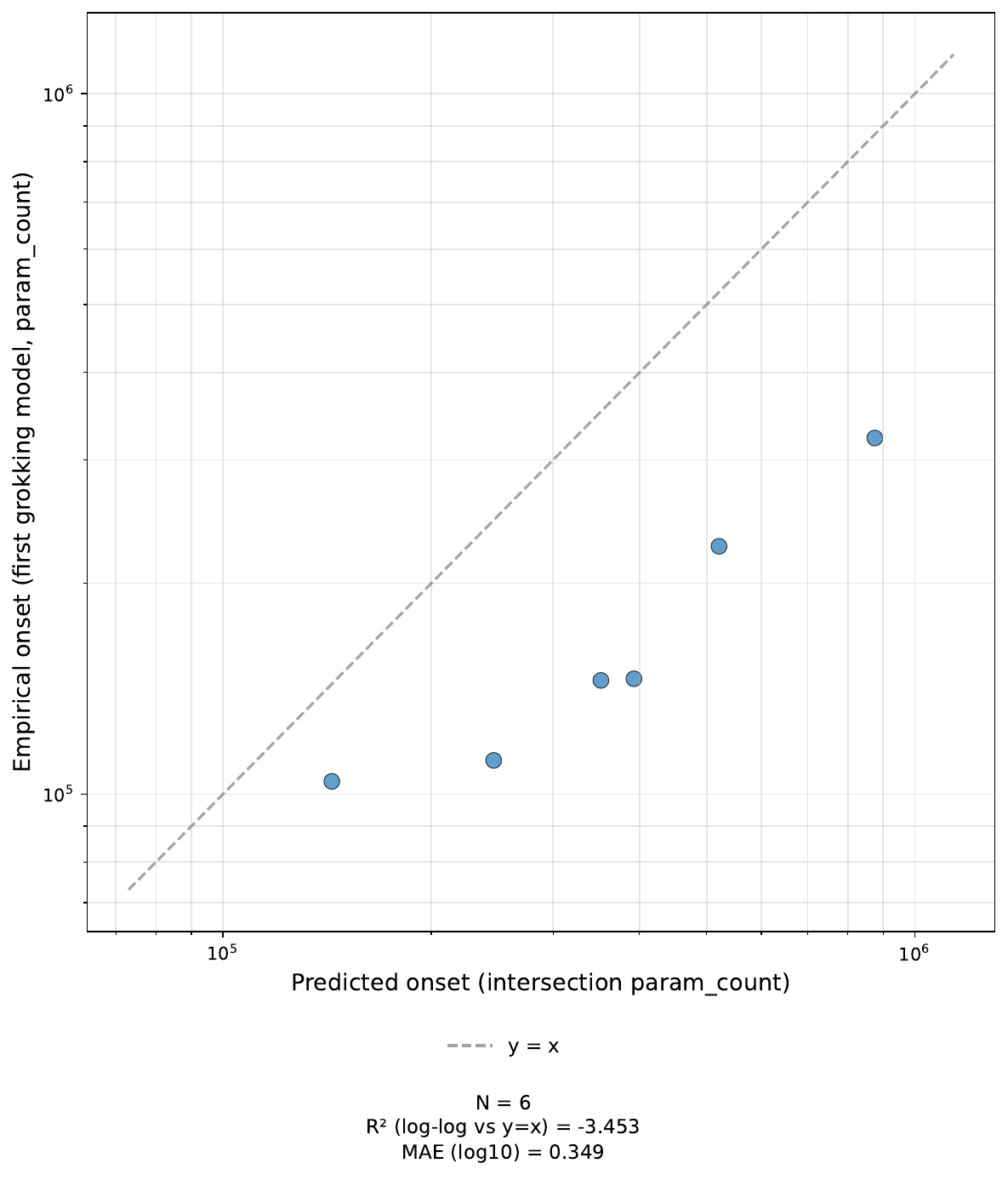}
        \caption{Predicted vs.\ empirical onset for identified prime/setting pairs.}
        \label{fig:task-add-pred-vs-emp}
    \end{subfigure}\hfill
    \begin{subfigure}{0.48\textwidth}
        \centering
        \includegraphics[width=\linewidth]{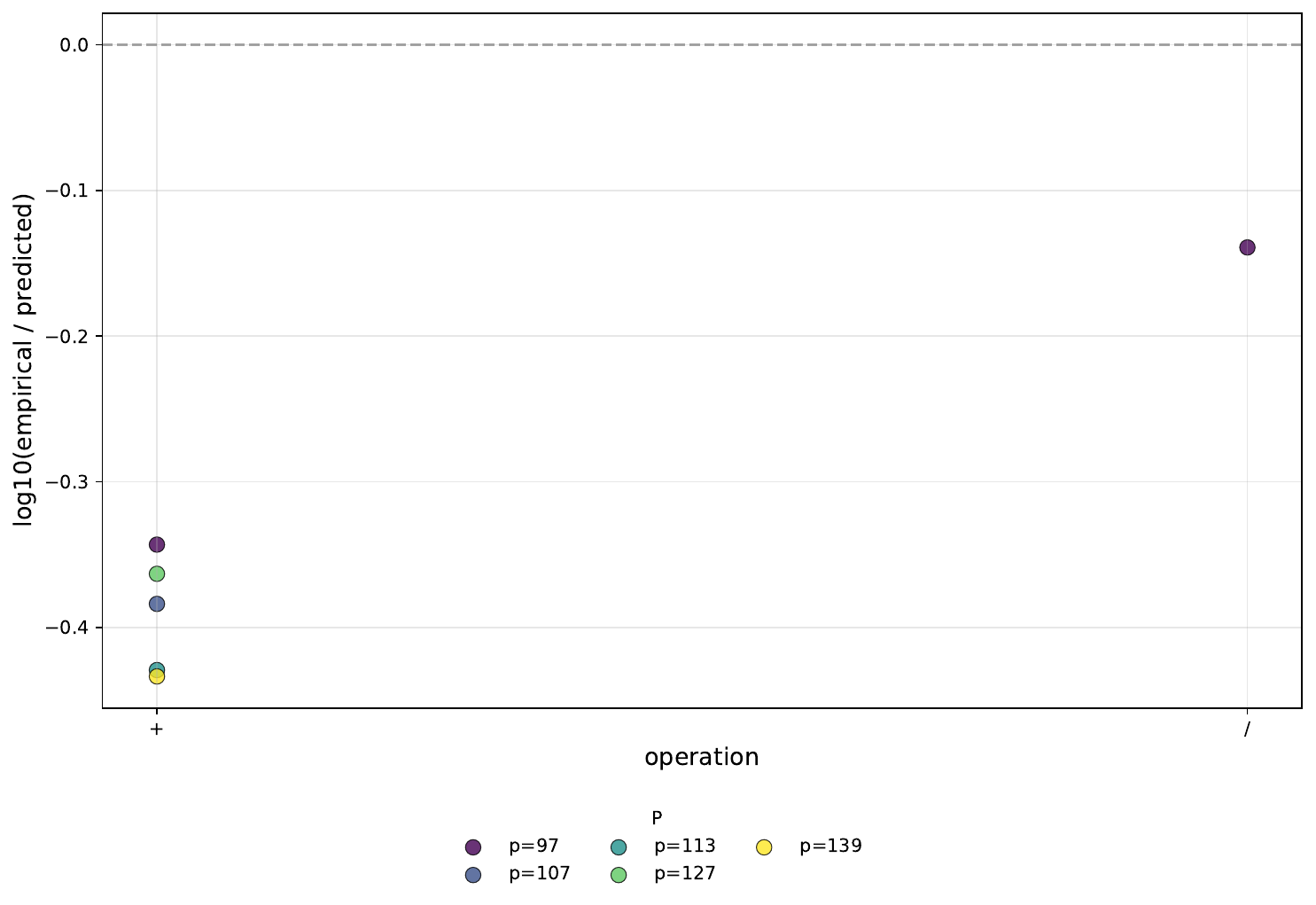}
        \caption{Log onset/prediction ratio for each operation.}
        \label{fig:task-add-error}
    \end{subfigure}
    \caption{Task-addition sweep: predicted-vs-empirical scatter and residual trend with the swept axis.}
    \label{fig:task-add-overview}
\end{figure}

\begin{table}[!htbp]
\centering\small
\caption{Addition and division: association and agreement between the measured timescale intersection and observed grokking onset, using 6 pairs from 10 tested combinations of prime and setting. A pair contributes only when both parameter counts can be determined. The regression models predict log onset: $M_0$ predicts the same value for every observation, $M_1$ uses log parameter count at the intersection, $M_2$ uses prime and the varied setting, and $M_3$ uses both sets of predictors. Each comparison asks whether the added predictors improve the fit. The last row compares with $P_{\mathrm{mem}}=K_{\mathrm{mem}}/C_{\mathrm{model}}$, the capacity-based estimate. CCC measures agreement in values; the calibration test asks whether predicted and observed values coincide. Median residual is $\log_{10}(P_{\mathrm{onset}}/\widehat P_{\mathrm{cross}})$. Dashes indicate omitted or unestimable tests. Resampling, test assumptions, and sample-size rules are given in \cref{sec:stats-methods}; the model comparisons use fits to the observed sample.}
\label{tab:task-add-pooled}
\begin{tabular}{lll}
\toprule
Test & Statistic & $p$-value \\
\midrule
Spearman $\rho$ & $1.000$ & $0.002$ \\
Kendall $\tau$ & $1.000$ & $0.003$ \\
Lin's CCC (95\% interval) & $0.379\;[0.075,0.547]$ & --- \\
Log--log slope & $0.663$ & --- \\
Intercept & $1.526$ & --- \\
Calibration $F$ test & $86.37$ & $5.12\times10^{-4}$ \\
Median log-residual; Wilcoxon & $-0.373$ & $0.031$ \\
$M_1$ vs. $M_0$ & $35.69$ & $0.004$ \\
$M_3$ vs. $M_2$ & $0.33$ & $0.625$ \\
$M_3$ vs. $M_1$ & $6.37$ & $0.136$ \\
Intersection adds over $P_{\mathrm{mem}}$ & $0.15$ & $0.721$ \\
\bottomrule
\end{tabular}
\end{table}

\begin{table}[!htbp]
\centering\small
\caption{Addition and division: coverage and agreement at each setting. The numerator counts primes for which both onset and intersection can be determined; the denominator counts all tested primes. The median and sample standard deviation (SD) summarise $\log_{10}(P_{\mathrm{onset}}/\widehat P_{\mathrm{cross}})$: negative values mean the intersection overestimates onset. SD requires two pairs, and rank correlation requires three. Seed pools, hardware, and reference measurements are specified in the text.}
\label{tab:task-add-within}
\begin{tabular}{lcccc}
\toprule
Setting & Pairs / eligible & Median log-ratio & SD & Spearman $\rho$ \\
\midrule
+ & 5/5 & $-0.384$ & $0.040$ & $1.000$ \\
/ & 1/5 & $-0.139$ & --- & --- \\
\bottomrule
\end{tabular}
\end{table}

\paragraph{Results.} There are 6 identified pairs among 10 eligible cases: five for addition and one for the division reference. Their pooled correlation is $\rho=1$ (permutation $p=0.0021$), but equality of predicted and observed parameter counts is rejected. Addition alone retains five pairs with $\rho=1$. Adding the intersection to prime and task gives no detectable improvement ($p=0.625$); the single identified division reference limits comparisons of calibration between tasks.

\subsection{Modular multiplication}
\label{sec:task-modular-mul}

\paragraph{Scope.} We compare modular multiplication with division, using five primes $\{97,107,113,127,139\}$ and four seeds (42--45) per retained cell. Non-reference settings use H100 runs, and the reference uses A100 runs. The hardware comparison in \cref{sec:exp-cone} supports approximate reproducibility, but does not remove this difference between settings. Timescale means exclude censored times; onset uses minimum paired delay and the fixed $d\le256$ domain. The reference uses the denser main-sweep width grid (\cref{sec:exp-sweeps}). We retain $C_{\mathrm{model}}=0.784933$ as a common normalisation.

\begin{figure}[!htbp]
    \centering
    \begin{subfigure}{0.48\textwidth}
        \centering
        \includegraphics[width=\linewidth]{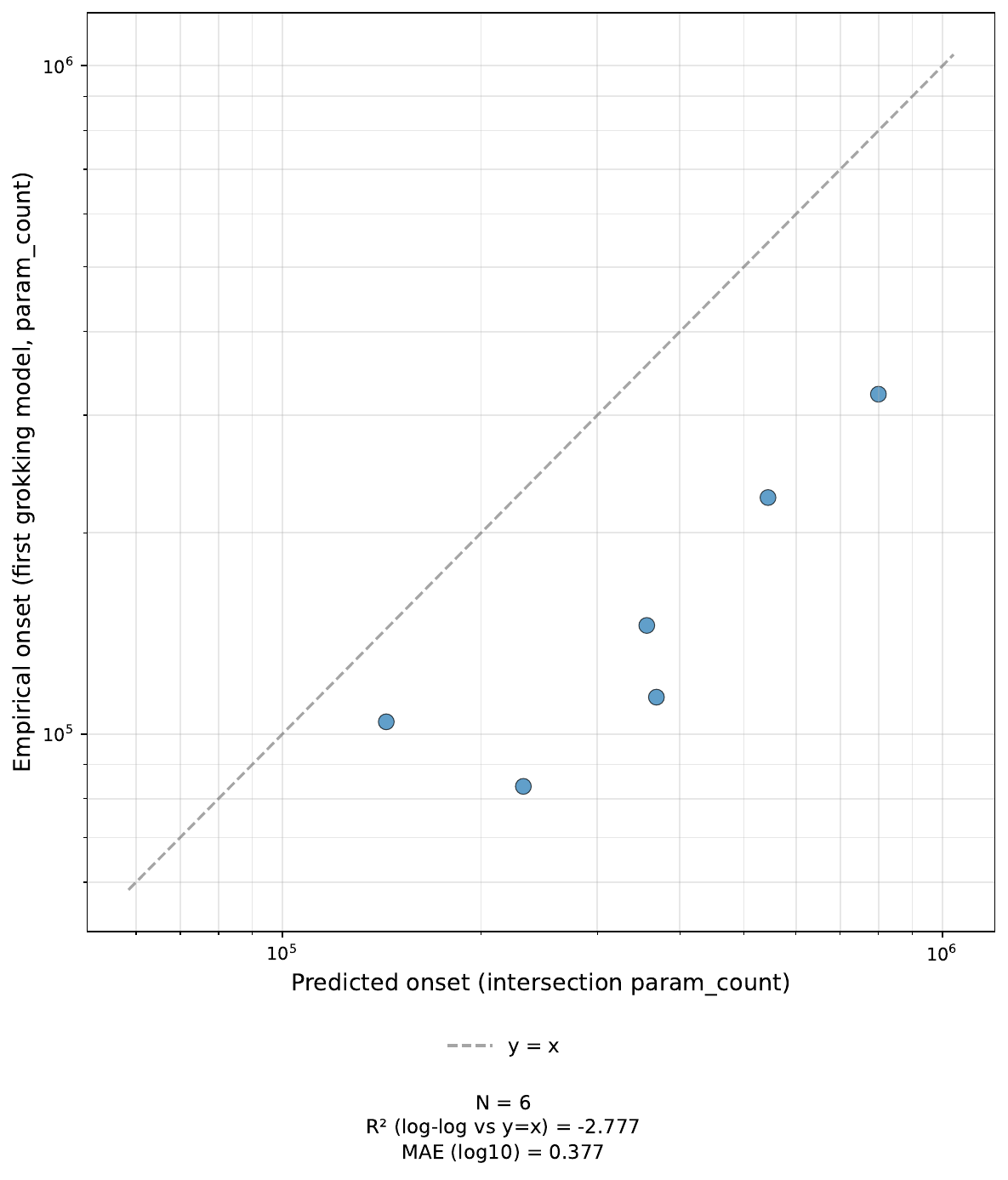}
        \caption{Predicted vs.\ empirical onset for identified prime/setting pairs.}
        \label{fig:task-mul-pred-vs-emp}
    \end{subfigure}\hfill
    \begin{subfigure}{0.48\textwidth}
        \centering
        \includegraphics[width=\linewidth]{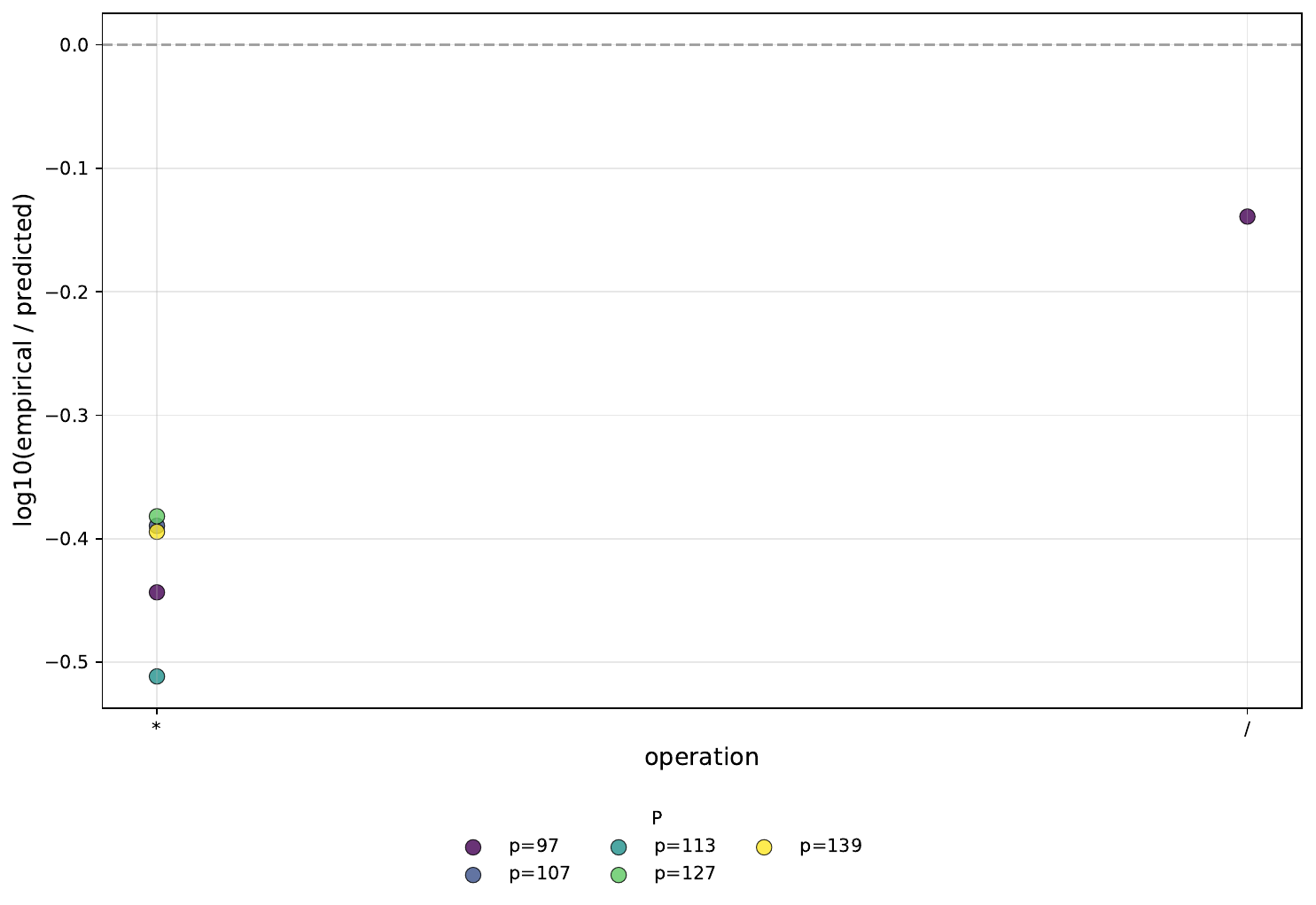}
        \caption{Log onset/prediction ratio for each operation.}
        \label{fig:task-mul-error}
    \end{subfigure}
    \caption{Task-multiplication sweep: predicted-vs-empirical scatter and residual trend with the swept axis.}
    \label{fig:task-mul-overview}
\end{figure}

\begin{table}[!htbp]
\centering\small
\caption{Multiplication and division: association and agreement between the measured timescale intersection and observed grokking onset, using 6 pairs from 10 tested combinations of prime and setting. A pair contributes only when both parameter counts can be determined. The regression models predict log onset: $M_0$ predicts the same value for every observation, $M_1$ uses log parameter count at the intersection, $M_2$ uses prime and the varied setting, and $M_3$ uses both sets of predictors. Each comparison asks whether the added predictors improve the fit. The last row compares with $P_{\mathrm{mem}}=K_{\mathrm{mem}}/C_{\mathrm{model}}$, the capacity-based estimate. CCC measures agreement in values; the calibration test asks whether predicted and observed values coincide. Median residual is $\log_{10}(P_{\mathrm{onset}}/\widehat P_{\mathrm{cross}})$. Dashes indicate omitted or unestimable tests. Resampling, test assumptions, and sample-size rules are given in \cref{sec:stats-methods}; the model comparisons use fits to the observed sample.}
\label{tab:task-mul-pooled}
\begin{tabular}{lll}
\toprule
Test & Statistic & $p$-value \\
\midrule
Spearman $\rho$ & $0.886$ & $0.033$ \\
Kendall $\tau$ & $0.733$ & $0.056$ \\
Lin's CCC (95\% interval) & $0.356\;[0.052,0.542]$ & --- \\
Log--log slope & $0.740$ & --- \\
Intercept & $1.066$ & --- \\
Calibration $F$ test & $31.20$ & $0.004$ \\
Median log-residual; Wilcoxon & $-0.392$ & $0.031$ \\
$M_1$ vs. $M_0$ & $13.58$ & $0.021$ \\
$M_3$ vs. $M_2$ & $2.89$ & $0.231$ \\
$M_3$ vs. $M_1$ & $6.55$ & $0.132$ \\
Intersection adds over $P_{\mathrm{mem}}$ & $0.22$ & $0.672$ \\
\bottomrule
\end{tabular}
\end{table}

\begin{table}[!htbp]
\centering\small
\caption{Multiplication and division: coverage and agreement at each setting. The numerator counts primes for which both onset and intersection can be determined; the denominator counts all tested primes. The median and sample standard deviation (SD) summarise $\log_{10}(P_{\mathrm{onset}}/\widehat P_{\mathrm{cross}})$: negative values mean the intersection overestimates onset. SD requires two pairs, and rank correlation requires three. Seed pools, hardware, and reference measurements are specified in the text.}
\label{tab:task-mul-within}
\begin{tabular}{lcccc}
\toprule
Setting & Pairs / eligible & Median log-ratio & SD & Spearman $\rho$ \\
\midrule
* & 5/5 & $-0.394$ & $0.054$ & $0.900$ \\
/ & 1/5 & $-0.139$ & --- & --- \\
\bottomrule
\end{tabular}
\end{table}

\paragraph{Results.} There are 6 identified pairs among 10 eligible cases: five for multiplication and one for the division reference. The pooled correlation is $\rho=0.886$ (permutation $p=0.033$), but equality of predicted and observed parameter counts is rejected. Multiplication alone has $\rho=0.9$ across five pairs. Adding the intersection to prime and task gives no detectable improvement ($p=0.231$); the single identified division reference limits comparisons of calibration between tasks.

\subsection{Other task sweeps}
\label{sec:task-future}
Subtraction would use the same information-matched random-label dataset size as addition, but its generalisation and grokking measurements are not included in these task sweeps. The subtraction toy model in \cref{sec:mechanism-protocols} is a separate experiment. Non-arithmetic tasks, such as permutation composition, remain untested.


\section{Out-of-distribution tests: protocols and figures}
\label{sec:exp-ood}

The experiments use the main-sweep training protocol unless stated otherwise. The fitted-law and measured-intersection comparisons below are retrospective analyses.

\subsection{Fitting details and scope of the extrapolation}
\label{sec:extrapolation-details}

The primary laws are fitted only on ten primes from $97$ through $139$. We evaluate their extrapolation using the ten-seed cone and a separate five-seed study of directly measured timescales.

On $p\in\{97,101,103,107,109,113,127,131,137,139\}$, both timescale laws use the arithmetic mean of observed seed times in each cell, followed by equal-cell-weight ordinary least squares (OLS) in log time. The retained width limit is $d\le256$, and the fixed capacity-fraction selection is $f\in[0.220146,2.751828]$ at $C_{\mathrm{model}}=0.784933$ (the same cells as the original $[0.08,1]$ convention at $C_{\mathrm{model}}=2.16$). Missing events are excluded before averaging. Memorisation contributes $70$ cells from $674$ observed runs, excluding $966$ censored runs; generalisation contributes $326$ cells from $3058$ observed runs, excluding $212$ censored runs. There are $3$ and $57$ partially observed contributing cells, respectively. All fit runs use A100.

The interval resamples the ten prime clusters in $2000$ draws, keeping observed cell means and eligibility fixed. It does not include unobserved event-time or seed-population uncertainty. The fold is a prediction of the fitted functions, not an exact empirical cutoff. The fold fraction $0.07168$ and several predicted intersections lie below the selected fitting window, so even some predictions within the prime range extrapolate in capacity fraction.

Across calibration primes, the fitted lower-intersection curve has a log--log slope $3.531$, compared with the empirical onset slope $3.149$. Agreement after fitting a single calibration offset is an in-sample descriptive check. The geometric consequences of the fitted forms are developed in \cref{sec:consequences}.

\subsection{Observations at $p=149$ and $p=197$}
\label{sec:theory-prereg}

At $p=149$, the ten-seed cone first confirms positive minimum delay at $d=112$ ($\log_{10}P=5.64$). However, one smaller measured model has an unresolved minimum, so the grokking point remains unresolved under \cref{sec:theory-grokking}; $d=112$ supplies an upper bound, not an identified onset. The zero minimum at the larger width $d=124$ does not affect the first-model criterion. At $d=96$, six seeds have positive delay and four have zero delay.

At $p=197>p^*$, the fitted laws predict no intersection. The eighteen-width grid in \cref{sec:exp-ood} spans $P=1.1\times10^5$ to $10^7$, with ten seeds per width, and has zero minimum delay at every width. A power law refitted to the ten identified main-sweep onsets predicts $\log_{10}P=5.91$ at $p=197$, with a 95\% prime-bootstrap fit interval $[5.75,6.05]$. The grid spans this interval without an onset. The interval uses $2000$ prime resamples and describes fit uncertainty only, excluding seed-population and censoring uncertainty. This is a finite-grid comparison, rather than evidence of disappearance at every width or in every seed. Of $177$ scored runs, $31$ have delays of $1$ to $6$ epochs at $d\ge188$, with seed-mean delay at most $1.4$ on this grid. All eighteen cell minima are determined, although three of the $180$ runs reach neither threshold within budget. The additional $d=316$ cell has delays $(1,1,1,2,1,1,0,1,1,0)$ epochs: its minimum is zero, but eight seeds have positive delay.

Lowering the training fraction $\alpha$ shrinks $K_{\mathrm{mem}}$ and moves $R(p)$ back above its threshold, so the intersection equation predicts that grokking can return at $p=197$. The five-seed training-fraction sweeps give measured onsets at $\log_{10}P=5.88$, $5.33$, and $5.03$ for $\alpha=0.4$, $0.3$, and $0.25$, respectively (\cref{sec:exp-ood}).

\subsection{The intersection measured directly}
\label{sec:theory-direct}

Random-label $T_{\mathrm{mem}}$ is measured independently of the task. Direct $T_{\mathrm{gen}}$ and the delay use the same task-validation accuracy history at different thresholds (\cref{sec:stats-methods}); their association is therefore not an independent validation of the predictor. The dense experiments cover eleven fixed-rate primes, with five seeds per timescale cell (\cref{sec:exp-ood}). We recompute both timescales as arithmetic means of observed events and identify actual lower roots in log--log coordinates.

At $p=149,151,157,163,167$, the revised fitted law and measured intersection differ by at most $0.093$ in $\log_{10}P$ (a factor of $1.24$). At $173,179,181,191,193,197$, the revised law has no root while the observed-event means have one. These missing predictions are extrapolation failures, not successful zero-error comparisons. In particular, $p=197$ now has a conditional measured intersection at $\log_{10}P=7.086$; only $83.8\%$ of bootstrap draws yield an intersection, with a broad conditional interval $[6.507,7.462]$.

Across the ten fixed-rate primes with identified intersections and first-model onsets, the mean onset-minus-intersection offset is $-0.089$ in $\log_{10}P$, with sample standard deviation (SD) $0.225$ in the same units. There is no basis for claiming that the offset vanishes. The cell-level timescale-ordering agreement is $77/106$ scoreable cells ($72.6\%$), with four additional cells excluded because timescale means are unavailable. The full denominators and root-existence frequencies are reported in \cref{tab:crossing}.

\paragraph{Additional cone diagnostics.}
\Cref{fig:cone} contains matched ten-seed cells, with minimum paired delay; cells containing unresolved seeds are omitted from the figure. Its positive-minimum region thins as prime increases, but five cells remain positive beyond the fitted terminal prime: three at $173$, one at $181$, and one at $191$. None of the sampled cells at $193$, $197$, or $199$ has positive delay in all ten seeds. At $p=197,d=316$, eight seeds have positive delay and two have zero delay. The cone therefore tests a finite-seed criterion and supports only an approximate terminal trend, not an exact population boundary. Wide-width zero minima do not imply zero mean delays: in the five-seed measurements, for example, $p=151,d=744$ has mean delay $4.6$ epochs while its minimum is zero. The upper branch of the fitted timescale equation is outside the reliable fit regime and is not empirically established.

\paragraph{The $p = 197$ column.} Eighteen widths $d \in [52, 556]$
($P = 1.1\times10^{5}$ to $10^{7}$), ten seeds, cap $5000$ epochs, training
fraction $0.5$; the onset estimator of \cref{sec:theory-grokking} evaluated
on the full sampled grid, retaining wider zero-delay models. Return
of grokking is tested with five seeds per width: ten widths $d\in[52,364]$ at $\alpha=0.4$ and $0.3$, and eight widths $d\in[40,148]$ at $\alpha=0.25$.

\paragraph{The direct intersection study.} Eleven fixed-rate primes $149,\allowbreak151,\allowbreak157,\allowbreak163,\allowbreak167,\allowbreak173,\allowbreak179,\allowbreak181,\allowbreak191,\allowbreak193,\allowbreak197$ each have ten widths and five seeds per timescale, giving $1{,}100$ H100 runs, including those that finish at the training cap without reaching the threshold. This five-seed study is separate from the ten-seed cone.

For both timescales, we average observed first-$99\%$ event times within each cell before taking logs. There are $21$ censored memorisation runs and $3$ censored generalisation runs in the fixed-rate study. The measured lower intersection is the first decreasing zero of the piecewise-linear log-timescale difference in $\log_{10}P$, restricted to the observed domain. No nearest approach or epoch cap is treated as an intersection. Bootstrap resampling draws five paired seed labels within each cell, recomputes observed means, and refinds the root in $1000$ draws. Reported percentile intervals condition on root existence, whose frequency is reported separately. All-censored cells and partial means are marked in the figures.

Delay uses task training-$99\%$ and validation-$98\%$ events. The onset is the first identified positive minimum across all five seeds, provided every smaller measured minimum is identified and zero. Otherwise it remains unresolved. Later zero minima do not invalidate an identified onset. Positive at the narrowest sampled width is left-censored and excluded from point comparisons. The full delay curves and all eleven prime rows remain available. Under this retrospective analysis, agreement between mean-timescale ordering and positive minimum delay is $77/106$ scoreable cells out of $110$ total cells.

\begin{figure}[!htbp]
  \centering
  \includegraphics[width=.98\linewidth]{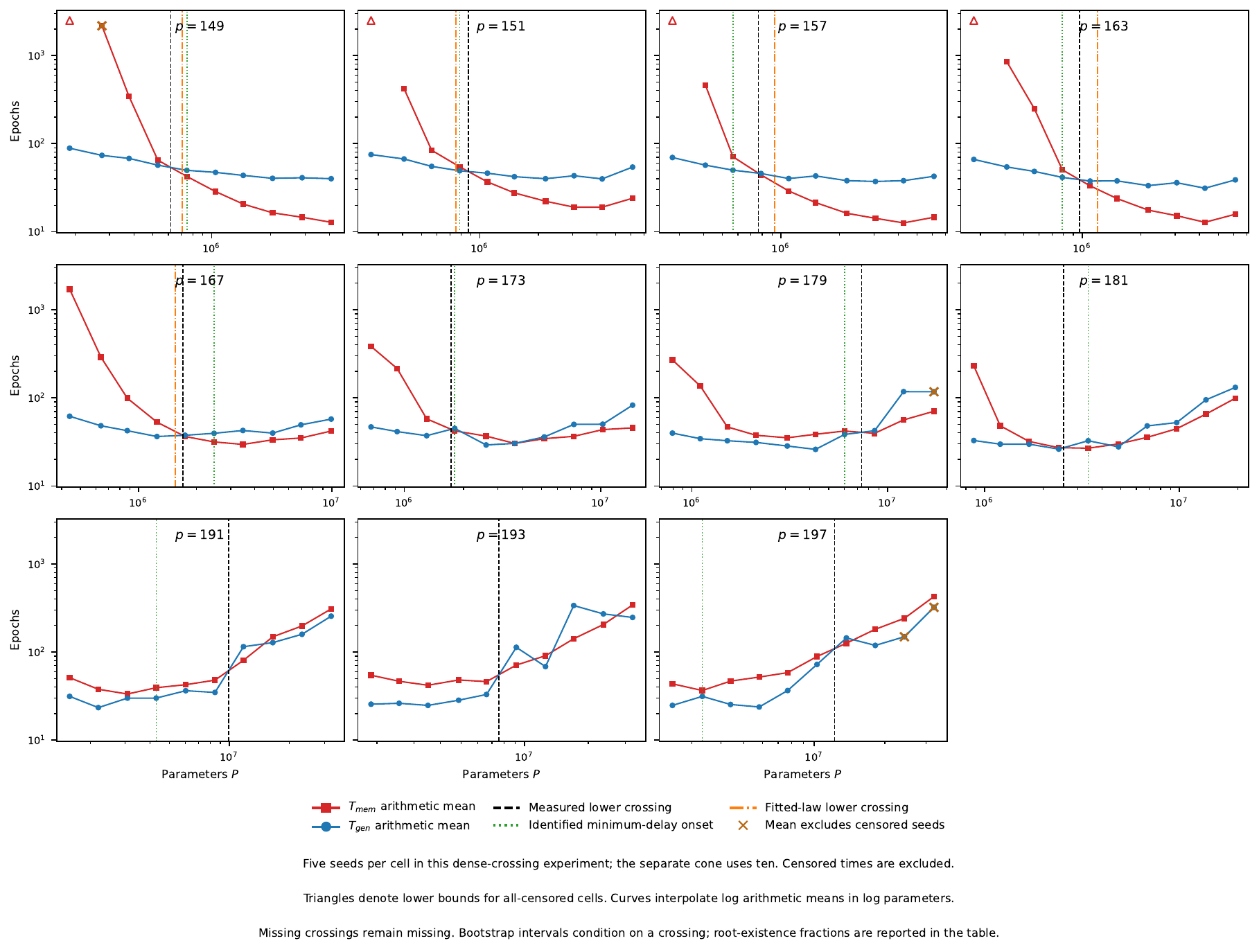}
  \caption{Direct fixed-rate measurements at eleven primes, with five seeds per timescale cell. Lines join arithmetic means of observed event times. Crosses mark means that exclude censored seeds; triangles mark follow-up lower bounds for all-censored cells. Dashed lines show measured lower intersections, dash-dot lines the revised global-law prediction where it exists, and dotted lines identified first-model onsets on the full sampled grid. The complete numerical comparison and bootstrap root fractions appear in \cref{tab:crossing}.}

  \label{fig:crossing}
\end{figure}

\begin{table}[!htbp]
\centering\small
\caption{Direct timescale intersections under fixed-rate, reported as $\log_{10}P$, recomputed from arithmetic means of observed events (five seeds per timescale cell). Intervals are percentile intervals among bootstrap draws with an intersection; root fraction reports how often one exists in 1,000 draws. A dash denotes no fitted-law root; unidentified onset is not zero. Agreement counts model sizes where mean-timescale ordering correctly indicates positive minimum delay, divided by all scoreable sizes; unresolved measurements are excluded. The onset is the first positive minimum on the full sampled grid and remains unresolved if a smaller minimum could still qualify. This retrospective reanalysis is separate from the ten-seed cone.}
\label{tab:crossing}
\begin{tabular}{lcccccc}
\toprule
$p$ & Fitted law & Measured & 95\% interval & Root fraction & Onset & Agreement \\
\midrule
149 & $5.85$ & $5.79$ & $[5.75,5.82]$ & $1.000$ & $5.87$ & 9/9 \\
151 & $5.88$ & $5.94$ & $[5.88,5.97]$ & $1.000$ & $5.90$ & 7/9 \\
157 & $5.97$ & $5.88$ & $[5.87,5.91]$ & $1.000$ & $5.75$ & 8/9 \\
163 & $6.08$ & $5.99$ & $[5.97,6.00]$ & $1.000$ & $5.90$ & 8/9 \\
167 & $6.19$ & $6.23$ & $[6.20,6.31]$ & $1.000$ & $6.39$ & 6/10 \\
173 & --- & $6.24$ & $[6.21,6.77]$ & $0.999$ & $6.26$ & 5/10 \\
179 & --- & $6.87$ & $[6.74,7.01]$ & $1.000$ & $6.78$ & 6/10 \\
181 & --- & $6.40$ & $[6.21,6.46]$ & $0.999$ & $6.53$ & 5/10 \\
191 & --- & $7.00$ & $[6.47,7.29]$ & $0.934$ & $6.63$ & 8/10 \\
193 & --- & $6.88$ & $[6.84,7.17]$ & $1.000$ & unidentified & 7/10 \\
197 & --- & $7.09$ & $[6.51,7.46]$ & $0.838$ & $6.52$ & 8/10 \\
\bottomrule
\end{tabular}
\end{table}

\endgroup

\input{theory/consequences}

\input{theory/extra}

\input{theory/principled}


\newpage
\input{checklist-revised.tex}

\end{document}

%% file: theory/cone-panels.tex
\ifdefined\conetimescaleheight\else\newlength{\conetimescaleheight}\fi
\setlength{\conetimescaleheight}{0.25\linewidth}
\begin{subfigure}{\linewidth}
    \centering
    \includegraphics[width=\linewidth]{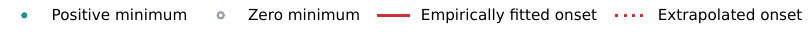}

    \smallskip
    \includegraphics[width=\linewidth]{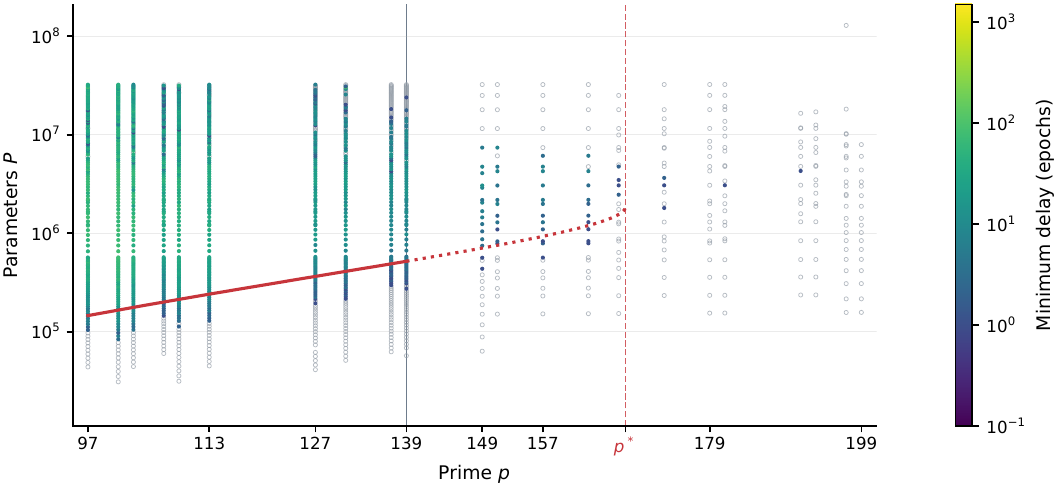}
    \caption{Observed generalisation delay}
    \label{fig:cone-observed}
\end{subfigure}

\medskip
\begin{subfigure}[t]{0.43\linewidth}
    \centering
    \includegraphics[height=\conetimescaleheight]{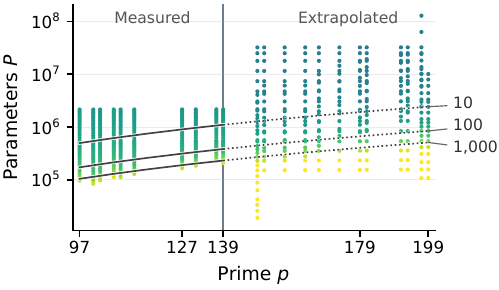}
    \caption{Memorisation timescale $T_{\mathrm{mem}}$}
    \label{fig:cone-mem}
\end{subfigure}\hfill
\begin{subfigure}[t]{0.43\linewidth}
    \centering
    \includegraphics[height=\conetimescaleheight]{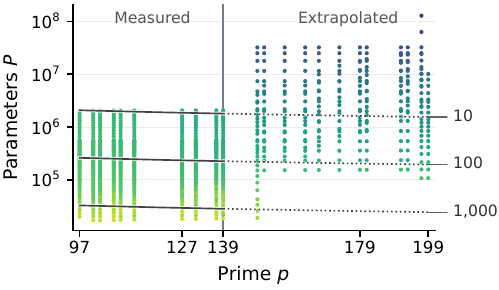}
    \caption{Generalisation timescale $T_{\mathrm{gen}}$}
    \label{fig:cone-gen}
\end{subfigure}\hfill
\begin{minipage}[t]{0.12\linewidth}
    \centering
    \includegraphics[height=\conetimescaleheight]{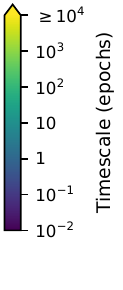}
\end{minipage}

%% file: theory/theory-section.tex
\section{Empirical Laws for the Two Timescales}
\label{sec:theory-account}

\subsection{Empirical laws and predicted endpoint of grokking onset}
\label{sec:theory-crossing}
\label{sec:theory-mem-law}

We observe that empirically, $T_{\mathrm{gen}}$ and $T_{\mathrm{mem}}$ scale across model and dataset sizes as follows:
\begin{align}
 T_{\mathrm{gen}}&=Ap^aP^{-b}, \label{eq:tgen-law} \\
 T_{\mathrm{mem}}&=B e^{\gamma f},\qquad f=\frac{K_{\mathrm{mem}}(p,\alpha)}{C_{\mathrm{model}}P}, \label{eq:tmem-law}
\end{align}
where $A$ and $B$ are positive constant factors, $a$ and $b$ are the prime and parameter-count exponents of the generalisation law, and $\gamma$ controls how memorisation time grows with the capacity fraction $f$.

In \cref{sec:mechanisms}, we report additional experiments that probe possible mechanisms behind these laws and assess competing explanations. Motivated by these observations, in \cref{sec:principled}, we develop simplified theoretical models. In particular, we show that the memorisation law may be consistent with a model of  repeated interference between examples that share model parameters. Meanwhile, we motivate the generalisation law by examining the learning of structured arithmetic features, where the dynamics may depend on weight decay and normalisation, although these models do not fully recover the observed scaling with prime and model size.

However, for the time being, we use these timescale laws as empirical models only. Fitting the laws to data collected from ten primes in $[97, 139]$, we obtain:
\[
 A\approx8.02\times10^8,\quad a\approx-0.452,\quad b\approx1.11,\quad B\approx2.94,\quad \gamma\approx15.5,
\]
with $R^2$ values of $0.8738$ and $0.8680$ in log-space for the $T_{\mathrm{gen}}$ and $T_{\mathrm{mem}}$ laws, respectively.

The empirical laws give us a way to predict grokking behaviour as models and datasets scale.
Specifically, we are interested in finding the intersection of $T_{\mathrm{gen}}$ and $T_{\mathrm{mem}}$ across different parameter counts and primes.
Equating the two timescale laws and substituting the definition of $f$ gives
\[
 \gamma f-b\ln(\gamma f)=R(p),
\]
where $R(p)=\ln A-\ln B+a\ln p-b\ln\!\bigl(\gamma K_{\mathrm{mem}}(p,\alpha)/C_{\mathrm{model}}\bigr)$.
For positive $b$, the left-hand side is strictly convex in $\gamma f$, with minimum $b(1-\ln b)$ at $\gamma f=b$. There are two solutions for $\gamma f$ when $R(p)$ exceeds this minimum, one when they are equal, and none otherwise. $R(p)$ decreases with $p$ for our coefficients, so the largest $p$ for which there is a solution for $\gamma f$ occurs at the critical value satisfying $R(p^*)=b(1-\ln b)$, giving
\begin{equation}
 p^*=167.89\qquad[148.84,193.38].\label{eq:pstar-tests}
\end{equation}
The brackets give a 95\% bootstrap interval obtained by resampling the ten fitted primes. Beyond this critical value, the fitted laws have no intersection and predict no region of consistent grokking. See \cref{sec:extrapolation-details} for fitting details and uncertainty, and refer to \cref{sec:consequences} for a more detailed discussion of the intersection geometry with derivations.

\subsection{Comparison with observed grokking region}
\label{sec:theory-region}

\Cref{fig:cone} compares the prediction that $p^*=167.89$ with empirically measured $\Delta E(P)$ across a grid of larger primes and different model sizes. As predicted by the extrapolated empirical laws, the region of consistent grokking thins as prime $p$ increases; moreover, no sampled model size at $p=193$, $197$, or $199$ has positive delay in all ten seeds. However, the predicted critical value of $p^*=167.89$ is only approximate: five cases of grokking are observed beyond it at $p=173$, $181$, and $191$. Overall, these results support the broad trend predicted by the empirical laws on out-of-distribution model and dataset sizes, but are not perfect quantitative matches. Observations on larger primes and additional cone diagnostics are reported in \cref{sec:theory-prereg,sec:theory-direct}.

%% file: theory/consequences.tex
\section{Consequences of the empirical laws: the geometry of the implicit onset condition}
\label{sec:consequences}

Throughout, the two fitted timescale laws are taken as given:
\begin{equation}
    T_{\text{gen}}(P,p) = A\, p^{a} P^{-b}, \qquad
    T_{\text{mem}}(P,p) = B\, e^{\gamma f}, \qquad
    f = \frac{K_{\mathrm{mem}}(p,\alpha)}{C_{\mathrm{model}}P},
    \label{eq:laws}
\end{equation}
with $A, B, a, b, \gamma$ the window fits of \cref{sec:theory-crossing} and
$K_{\mathrm{mem}}(p,\alpha)=\alpha p(p-1)\log_2(p+2)$ at fixed $\alpha=1/2$. We denote the lower intersection of these fitted laws by $P_{\mathrm{cross}}(p)$, distinct from the measured-curve intersection $\widehat P_{\mathrm{cross}}(p)$ and empirical onset $P_{\mathrm{onset}}(p)$. For the intersection geometry we extend $p$ continuously beyond its prime-valued observations. The intersection geometry below follows from \cref{eq:laws} without additional fitted parameters. The later mechanism comparisons and auxiliary approximations require their separately stated assumptions.

\subsection{Reduction to a one-variable intersection equation}

\begin{proposition}[Normal form]
\label{prop:normal-form}
Define the scaled capacity fraction $x = \gamma K_{\mathrm{mem}}(p,\alpha)/(C_{\mathrm{model}} P)$ (so $x = \gamma f$, and
$P = \gamma K_{\mathrm{mem}}/(C_{\mathrm{model}} x)$). Then $T_{\text{mem}}(P,p) = T_{\text{gen}}(P,p)$ if and only if
\begin{equation}
    F(x) \;:=\; x - b \ln x \;=\; R(p),
    \qquad
    R(p) := \ln\frac{A p^{a}}{B} - b \ln \frac{\gamma K_{\mathrm{mem}}(p,\alpha)}{C_{\mathrm{model}}}.
    \label{eq:normal-form}
\end{equation}
\end{proposition}

\begin{proof}
Take logs of $B e^{x} = A p^a P^{-b}$ and substitute
$\ln P = \ln(\gamma K_{\mathrm{mem}}/C_{\mathrm{model}}) - \ln x$:
$x + \ln B = \ln(A p^a) - b \ln(\gamma K_{\mathrm{mem}}/C_{\mathrm{model}}) + b \ln x$. Rearranged, this is
\cref{eq:normal-form}.
\end{proof}

All $p$-dependence is confined to the single scalar $R(p)$, which is strictly
decreasing: writing $\kappa_p := d\ln K_{\mathrm{mem}}/d\ln p = 1 + \frac{p}{p-1} +
\frac{p}{(p+2)\ln(p+2)} \approx 2.21$ over the measured range,
\begin{equation}
    \frac{dR}{d\ln p} \;=\; a - b\, \kappa_p \;\approx\; -0.45 - 1.109 \times 2.21
    \;\approx\; -2.90 \;<\; 0 .
    \label{eq:R-slope}
\end{equation}

\subsection{Existence and the critical prime}

$F(x) = x - b\ln x$ on $x>0$ is strictly convex with a unique minimum at
$x = b$, $F(b) = b(1 - \ln b) =: R^*$. Hence:

\begin{proposition}[Fold structure]
\label{prop:fold}
(i) Intersections exist if and only if $R(p) \ge R^*$; since $R$ is strictly decreasing, this
defines a unique critical value $p^*$ by
\begin{equation}
    R(p^*) = b\,(1 - \ln b).
    \label{eq:pstar}
\end{equation}
(ii) For $p < p^*$ there are exactly two intersections $x_-(p) < b < x_+(p)$: the model's
lower intersection is the \emph{larger} root $x_+$ (smaller $P$), and $x_-$ is a
second, re-entrant boundary at larger $P$ beyond which
$T_{\text{mem}} > T_{\text{gen}}$ again.
(iii) At $p = p^*$ the two roots merge at $x = b$: the fitted lower-intersection branch terminates not by
running to infinite width but at the finite fold point
\begin{equation}
    f_{\text{fold}} = \frac{b}{\gamma}, \qquad
    P_{\text{fold}} = \frac{\gamma\, K_{\mathrm{mem}}(p^*,\alpha)}{C_{\mathrm{model}}\, b}.
    \label{eq:fold-point}
\end{equation}
\end{proposition}

\begin{proof}
(i)–(iii) follow from convexity of $F$, monotonicity of $R$
(\cref{eq:R-slope}), and $F'(b) = 0$, $F''(b) = 1/b > 0$. That $x_+$
is the lower intersection: as $P$ increases from small values, $x$ decreases from large values,
and $h(P) := \ln T_{\text{mem}} - \ln T_{\text{gen}} = F(x) - R$ first changes
sign at the largest root, $x_+$.
\end{proof}

Numerically, the selected arithmetic-cell-mean fits give $p^*=167.8858$, $f_{\mathrm{fold}}=0.071679$, and $P_{\mathrm{fold}}=1.8446\times10^6$. This fraction is below the fitted selection window $[0.220146,2.751828]$. These are roots and extrapolations of the fitted functions, not observed phase boundaries.

Two structural consequences deserve emphasis. First, \emph{the fitted lower intersection does not
diverge continuously}: as $p \to p^*$ the fitted lower intersection approaches
$P_{\text{fold}} \approx 1.8$M parameters and then the predicted grokking region closes at a saddle-node (fold) bifurcation. Second, by
(ii) \emph{the fitted model predicts a closed cone in the $(P, p)$ plane}: bounded
below by the lower intersection branch $x_+$, above by the re-entrant branch $x_-$, with apex
$(P_{\text{fold}}, p^*)$.

\subsection{Local scaling near the fold}

Expanding $F$ about its minimum, $F(x) = R^* + (x-b)^2/(2b) + O((x-b)^3)$,
so with $\Delta R(p) := R(p) - R^*$,
\begin{equation}
    x_\pm(p) \;=\; b \pm \sqrt{2b\,\Delta R(p)}+\frac{2}{3}\Delta R(p)+O(\Delta R^{3/2}),
    \qquad
    \Delta R(p) \approx |a - b\kappa_{p}|\, \ln\frac{p^*}{p},
    \label{eq:sqrt-scaling}
\end{equation}
i.e.\ the two boundaries of the cone close like a square root in $\ln(p^*/p)$
(verified: at $p = 167$, $x_+ - b = 0.1945$ vs the leading square-root approximation
$0.1841$; the common linear correction is $0.0102$).

\begin{proposition}[Local exponent of the lower intersection]
\label{prop:seff}
The local power-law exponent of the lower intersection branch,
$s_{\mathrm{eff}}(p) := d\ln P_{\text{cross}}/d\ln p$, is
\begin{equation}
    s_{\mathrm{eff}}(p) \;=\; \kappa_p \;+\; \frac{b \kappa_p - a}{\,x_+(p) - b\,}.
    \label{eq:seff}
\end{equation}
\end{proposition}

\begin{proof}
$\ln P = \ln(\gamma K_{\mathrm{mem}}/C_{\mathrm{model}}) - \ln x_+$, so $s_{\mathrm{eff}} = \kappa_p - d\ln
x_+/d\ln p$. Differentiating \cref{eq:normal-form} at fixed root branch,
$F'(x_+)\, dx_+ = dR$, with $F'(x_+) = 1 - b/x_+ = (x_+-b)/x_+$, gives
$d\ln x_+ / d\ln p = (a - b\kappa_p)/(x_+ - b)$, which is
\cref{eq:seff}.
\end{proof}

The local exponent is a consequence of the fitted timescale functions: $s_{\mathrm{eff}}=3.282$ at $p=105$, $3.410$ at $113$, $4.150$ at $139$, and $8.055$ at $163$. The fitted lower-intersection curve has OLS log--log slope $3.531$ across the ten calibration primes; the measured minimum-delay onset slope is $3.149$. Their similarity is a consistency observation, not an exact derivation of the measured exponent. The exponent of the fitted branch diverges as the two roots merge, but the finite data need not follow that extrapolation.

\subsection{The upper edge: what the second root predicted and what the data show}

The upper root exists in the fitted exponential/power-law model, but lies below the reliable capacity-fraction range of the timescale fits. Its empirical location is not established. An intersection of mean times also does not imply a zero minimum or mean paired delay without assumptions linking the random-label proxy to task training.

Wide-width minima can be zero while means remain positive. In the five-seed fixed-rate data, fully observed examples include $(p,d)=(149,1000)$ with mean delay $36.4$ epochs, $(151,744)$ with $4.6$, and $(163,1000)$ with $3.2$, all with zero minimum. Thus a zero-minimum upper edge cannot be described as disappearance of mean delay. Task training and validation both slow at large fixed-rate widths; this does not validate the fitted upper intersection. The lower-branch extrapolation also has counterexamples beyond $p^*=167.89$ in the ten-seed cone.

\subsection{Comparing the timescale exponent with a storage-counting cost}
\label{sec:c3}

Storage-volume and activated-dynamics models can involve extensive exponents, written schematically as $\ln T\sim N\Delta F(\alpha_{\mathrm G})$, where $N$ is system size, $\alpha_{\mathrm G}=n/N$ is pattern load, and $\Delta F$ is a cost per parameter. The following comparison tests the specific identification of a counting cost with a learning-time exponent; it does not rule out these broader mechanism classes.

\begin{proposition}[Leading-order storage entropy and mean-volume counting]
\label{prop:counting}
For the zero-margin spherical perceptron at small load $\alpha_{\mathrm G}$, the replica-symmetric free entropy relative to the unconstrained volume has the small-load expansion \citep{gardner1988space}
\begin{equation}
 s(\alpha_{\mathrm G})=-\alpha_{\mathrm G}\ln2+O(\alpha_{\mathrm G}^2).
\end{equation}
The factor of two per binary constraint is a leading-order statement about typical log volume. Separately, for independent uniform $V$-ary labels, the expected fraction of parameter space that classifies all $n$ examples correctly is $V^{-n}$, provided ties have zero measure and the parameter measure is independent of the labels. Its negative log per parameter is $(n/P)\ln V=(K_{\mathrm{mem}}/P)\ln2=C_{\mathrm{model}}f\ln2$. This is the log of an expected volume, not an exact expression for the expected log volume or for a dynamical barrier.
\end{proposition}

\begin{proof}[Proof sketch]
For the binary replica-symmetric expansion, write $H(z)=\int_z^\infty e^{-t^2/2}\,dt/\sqrt{2\pi}$. At overlap $q=0$, the functional equals $\alpha_{\mathrm G}\ln H(0)=-\alpha_{\mathrm G}\ln2$; the small-load saddle $q^*=O(\alpha_{\mathrm G})$ changes this at second order. For the separate mean-volume identity, each fixed classifier matches $n$ independent uniform labels with probability $V^{-n}$. Integrating over a label-independent normalised parameter measure gives the same expectation. This identity does not transfer the binary quenched-entropy expansion to a general multiclass model.
\end{proof}

\begin{corollary}[Constraint on a direct counting-cost interpretation]
\label{cor:c3}
If an escape-time model with reference time $T_0$ identifies its exponent directly with the counting cost, $\ln(T/T_0)=K_{\mathrm{mem}}\ln2=PC_{\mathrm{model}}f\ln2$, it predicts $14962$ nats at $P=10^5$, $f=0.275$, and $C_{\mathrm{model}}=0.784933$. At the same fraction, the fitted timescale gives $\ln(T_{\mathrm{mem}}/B)=\gamma f=4.254$ nats. This direct identification is incompatible with the measured timescale scale. The conclusion does not exclude every activated-dynamics model with an independently specified barrier or prefactor.

\end{corollary}

The empirical collapse motivates mechanisms whose dominant exponent depends on information per parameter. The comparison above constrains a specific direct counting-cost interpretation; it is not a general theorem excluding all barrier models. Residual prime and width dependence, finite-budget selection, and the difference between full-range and window fits remain part of the empirical uncertainty.

\subsection{Two auxiliary derivations}

\paragraph{The extreme-value approximation for initial overlap.}
Let $F_k$ be the initial fraction of power in character pair $k$, with $m$ such pairs and embedding width $d$. For complex conjugate character pairs of an independent Gaussian initialisation, the power has a chi-square form with $2d$ degrees of freedom, motivating the approximation
\begin{equation}
 \mathbb E[\max_k F_k]\approx\frac1m\left(1+\sqrt{\frac{2\ln m}{d}}\right).
\end{equation}
This is an extreme-value approximation, not an exact identity for normalised shares. The nontrivial real character in the multiplicative group is exceptional and does not have the same complex-mode power law. In the stored comparisons, the approximation differs from real-initialisation means by up to $6.54\%$ and from Gaussian-null means by up to $3.89\%$; real and Gaussian means differ by up to $2.49\%$. These three comparisons should not be conflated.

\paragraph{Censoring and curvature.}
Excluding uncompleted runs removes long times from an observed-event mean. The resulting timescale estimates condition on completion within the budget. This fact alone does not determine how the selection bias varies with capacity fraction, nor the sign of its second derivative. Positive fitted curvature therefore is not generally a lower bound on population curvature, and the fitted slope is not automatically conservative. Additional assumptions or extended follow-up would be needed for those claims.

\subsection{What the cone predicts at $p = 197$}

The point estimate $p^*=167.89$ puts $197$ beyond the fold of the fitted functions. The eighteen-width grid has no positive ten-seed minimum, whereas a power law refitted to the identified first-model onsets predicts $\log_{10}P=5.91$, with a 95\% prime-bootstrap fit interval $[5.75,6.05]$. This comparison applies to the stated seed pool and width domain. It does not turn the model fold into a theorem about all widths or seeds; the positive-minimum cells at $173,181,191$ already depart from the point prediction.

The eighteen-width grid has zero minimum delay at all eighteen
widths. The additional $p=197$, $d=316$ cell has delays
$(1,1,1,2,1,1,0,1,1,0)$ epochs, with zero minimum but eight observed positive
delays. The eighteen-width result remains valid; it
cannot be extended to every width or every seed (\cref{sec:theory-prereg}).

%% file: theory/extra.tex
\section{Mechanistic evidence and competing explanations}
\label{sec:mechanisms}

\input{theory/mechanism-protocols}

\subsection{The memorisation exponent $\gamma$}
\label{sec:mech-mem}

The batch experiments test how the fitted memorisation exponent changes with optimiser noise and update frequency. The auxiliary individual-run fits give a decreasing exponent as batch size grows: the two-batch observed-only OLS comparison gives approximately $B_{\mathrm{batch}}^{-0.47}$, while censored regression across all three batches gives $B_{\mathrm{batch}}^{-0.378}$. These are different estimators, and neither is the primary fit to mean timescales. Learning-rate probes and longer-budget measurements provide additional constraints (\cref{sec:probe-protocol}). The table contrasts specific candidate predictions with these observations.

\begin{center}\small
\begin{tabular}{p{.25\linewidth}p{.28\linewidth}p{.36\linewidth}}
\toprule
Mechanism & Prediction & Verdict \\
\midrule
Thermal Arrhenius ($T_{\mathrm{eff}}\!\propto\!\eta/B_{\mathrm{batch}}$) & $\gamma\propto B_{\mathrm{batch}}/\eta$ & incompatible batch trend: $\gamma$ falls as $B_{\mathrm{batch}}$ grows \\
Direct counting-cost timescale & $\ln(T/T_0)=K_{\mathrm{mem}}\ln2$ & incompatible numerical scale; other barriers not excluded \\
Critical slowing $(f_c-f)^{-k}$ & divergence at a finite fraction $f_c$ & finite-budget observations do not locate a population divergence \\
Directional interference & intensive $e^{\gamma f}$ under rate assumptions & candidate; effective geometry unmeasured \\
\bottomrule
\end{tabular}
\end{center}
Here $T$ is a generic learning time, $T_0$ a reference time, $T_{\mathrm{eff}}$ an effective noise temperature, $f_c$ a putative critical fraction, and $k>0$ the critical-slowing exponent.

\paragraph{Interference as a candidate explanation.}
The conditional projection model in \cref{sec:principled-mem} motivates an exponential in information per parameter through repeated local repair. Its exact result concerns expected attenuation under independent random projections; the conversion to a fitting time and the effects of noise, decay, and optimiser choice require further assumptions. The auxiliary batch analysis gives $\gamma(B_{\mathrm{batch}})=3.98+319.34/\sqrt{B_{\mathrm{batch}}}$. The observed reduction in $\gamma$ as batch size grows and its increase with decay are constraints on an interference model, rather than consequences already proved by the projection identity. These auxiliary censored-regression estimates are distinct from the primary arithmetic-cell-mean fits.

\subsection{The generalisation law $T_{\mathrm{gen}} = g(p,\alpha,\lambda)\,P^{-b}$}
\label{sec:mech-gen}

Here $g(p,\alpha,\lambda)$ is the generalisation-time prefactor at the given prime, training fraction, and decay. The matched weight-decay comparison gives $T_{\mathrm{gen}}\propto\lambda^{-0.512}$ over its selected complete widths. Character ablations and traces probe which features matter and when they become prominent. These measurements constrain proposed mechanisms, but do not by themselves derive the generalisation law.

\begin{center}\small
\begin{tabular}{p{.25\linewidth}p{.28\linewidth}p{.36\linewidth}}
\toprule
Mechanism & Prediction & Verdict \\
\midrule
Init-overlap saddle escape & $T\!\sim\!\ln(1/m_0)/\mathrm{gap}$ & initial bias alone is insufficient; width-dependent gap remains possible \\
Norm-minimisation timescale & $T \propto 1/(\eta\lambda)$ & incompatible with the matched slope $-0.51$ as a complete timescale law \\
Kernel/lazy regime & sample requirements depend on task and assumptions & the present measurements do not isolate a kernel regime \\
Normalisation-mediated growth & half-power under additional equilibrium assumptions & candidate; not derived by the bare bilinear system \\
\bottomrule
\end{tabular}
\end{center}
Here $m_0$ is the initial overlap with the generalising mode and $\mathrm{gap}$ its linear instability growth rate.

\paragraph{Observed selection and later concentration.}
The character product identity motivates probing embedding and unembedding weights. Output-component ablations establish functional importance without reconstructing the entire circuit. The per-character trace
measurement at $p=113$, $d=128$, and seed $42$ follows multiplicative-character power shares every five epochs for $2{,}500$ epochs (\cref{sec:character-protocol}). It distinguishes early enrichment of the eventual winners from their later concentration:

\emph{Early enrichment.} The eventual winners, defined using the final checkpoint, have elevated share
before validation reaches $99\%$: the winning cohort's mean share exceeds $1.5\times$
uniform by epoch 25 and $2\times$ by epoch 40, against training-$99\%$ at
epoch 55 and validation-$99\%$ at epoch 95. The eventual winners are therefore already enriched before training accuracy saturates. Separate initialisation measurements show a bias (elevated initial share in eight of ten runs, Wilcoxon $p=0.02$, mean initial percentile $0.63$), and the effective character number in the trace has already fallen
$55\to16$ by validation-$99\%$.

\emph{Later concentration.} Post-grok concentration ($16\to11$
effective characters over the next $2{,}400$ epochs) accompanies a stronger decline in losing-bin shares: mean log-share slopes are $-2.1\times10^{-4}$/epoch
for losing characters (the most-pruned reach $-1.5\times10^{-3}$) versus
$+5\times10^{-5}$/epoch for winners (aggregate loser mass decays at $-3.5\times10^{-4}$/epoch). The change in relative shares motivates a pruning interpretation. Shares alone do not establish absolute growth or decay, since they are renormalised by total power; the absolute-power measurements are specified in \cref{sec:character-protocol}. Attributing the observed concentration specifically to weight decay requires comparison across interventions.

The bare coupled-factor system and the separate normalisation-equilibrium argument are developed in \cref{sec:principled-gen}. Neither the width exponent nor the observed decay dependence follows without additional assumptions.

\paragraph{The $\lambda$-sweep of the traces.}
For $\lambda\in\{0.1,0.3,1.0\}$, one seed per setting has winner spectral-power growth rates $1.05,1.17,3.55\times10^{-3}$ per epoch, giving a fitted exponent $+0.533$. These rates are measured in the interval from grokking to $300$ epochs afterwards. They describe post-transition growth and cannot directly identify the preceding plateau-escape timescale. The middle point lies below the fitted trend, and the sample is too small to establish a universal rate law. Early winner selection and later loser pruning remain distinguishable observations; assigning either stage a causal timescale requires additional interventions or phase-specific measurements.

\subsection{The calibration offset $\hat\Delta$}
\label{sec:mech-offset}

The calibration offset is empirical. We define $\hat\Delta:=\operatorname{median}_p\log_{10}(P_{\mathrm{onset}}/\widehat P_{\mathrm{cross}})$ for the central measured-mean-curve test, which gives $\hat\Delta=-0.170$; the globally fitted law compared with the ten-seed cone has median $-0.271$. The dense fixed-rate comparison gives mean $-0.089$ with sample SD $0.225$ over ten identified pairs. These are different predictors and cohorts.

A positive minimum paired delay requires every seed to have positive delay. Under monotone individual onset curves this selects the latest individual onset, not the earliest. Moreover, lowering the validation threshold from $99\%$ to $98\%$ cannot increase a run's clipped delay. An earliest-seed order statistic and threshold lowering therefore do not explain the negative offset. The measured offsets are not parameter-free theoretical predictions; their sensitivity to thresholds, seed count, grid, and censoring is reported explicitly.

\subsection{Additional mechanism tests}
\label{sec:mech-closure}

The following results describe the mechanism section's empirical accounting:

\begin{itemize}
\item \textbf{Changing the training fraction.} Reducing $\alpha$ reduces the information assigned to memorisation in the intersection equation. At $p=197$, the five-seed follow-ups identify onsets at $\log_{10}P=5.88$, $5.33$, and $5.03$ for $\alpha=0.4$, $0.3$, and $0.25$, respectively (\cref{sec:theory-prereg}). This supports the predicted direction of change on the tested grids. It does not establish a sharply bounded region in $(P,p,\alpha)$, or determine how the generalisation law changes with training fraction.
\item \textbf{The constants.} $\gamma_\infty = 3.98$, $c' = 319.34$
(\cref{sec:mech-mem}); together with the measured decay dependence $T_{\mathrm{gen}}\propto\lambda^{-0.512}$, these provide quantitative targets for candidate mechanisms.
\end{itemize}

%% file: theory/mechanism-protocols.tex
\subsection{Protocols for the mechanism experiments}
\label{sec:mechanism-protocols}

These experiments probe possible explanations of the measured timescales. Their estimators and observation windows are specified separately from the primary timescale-law fits in \cref{sec:extrapolation-details}.

\paragraph{Matched weight-decay comparison.}
We use $400$ A100 modular-division runs at $p=113$, with the central architecture, optimiser, and split protocol in \cref{sec:exp-arch}. The grid is
\begin{align*}
 d&\in\{20,24,28,32,40,48,56,64,80,96,112,128,160,192,224,256\},\\
 \lambda&\in\{0.01,0.03,0.1,0.3,1.0\},\qquad
 s\in\{42,43,44,45,46\},
\end{align*}
where $s$ denotes the random seed. The event is first validation accuracy $\ge99\%$, with a $5{,}000$-epoch cap. Twelve runs are censored. The primary analysis retains the twelve widths $d\ge40$ in this grid, for which all $25$ seed/decay events are observed. At each width we first average event times over the five seeds, then fit log mean time against log decay. The median of the twelve width-specific slopes is $-0.512$, with a $95\%$ paired-seed bootstrap interval $[-0.559,-0.472]$. The bootstrap enumerates all $5^5$ resamples of seed labels, using the same labels jointly across widths and decay settings. Eligibility is fixed, so the interval is conditional on the complete-width grid. Capped-time sensitivity analyses are separate from observed-event estimates. The selection and calculation are implemented in \texttt{scripts/weight\_decay\_scaling.py}.

\paragraph{Paired normalisation-gain intervention.}
We use $15$ matched H100 pairs at $p=113$, seeds $42$--$44$, and
\[
 (d,\lambda)\in\{(128,0.1),(128,0.3),(128,1),(64,1),(256,1)\}.
\]
Each pair starts with identical weights, data, and random-number-generator states. One arm freezes all five RMSNorm gain vectors at their initial values; the other trains them. Both retain the RMS normalisation operation. Training uses the central protocol and a $5{,}000$-epoch cap, with no additional post-transition training. All $30$ validation-$99\%$ events are observed. We fit log arithmetic cell-mean time against log decay at $d=128$, and against log total architecture parameter count at $\lambda=1$. Both arms use the same total parameter count, including frozen gains. The decay slopes are $-0.611$ frozen and $-0.659$ unfrozen. Uncertainty resamples the three seed labels jointly across both arms and all settings, enumerating $3^3$ resamples; with only three seeds it is descriptive. These experiments test the contribution of trainable gains, not the removal of normalisation. Source and analysis are \texttt{scripts/run\_normfrozen\_h100\_paired.py} and \texttt{scripts/normfrozen\_analysis.py}.

\paragraph{Auxiliary batch and full-batch fits.}
The separately labelled batch analyses use individual random-label runs, rather than logarithms of arithmetic cell means. The H100 grid uses primes $97,113,139$, widths $40,48,56,64,80,96,112,128,152,176,216,256$, and seeds $52$--$54$, with the main architecture, dropout $0.2$, learning rate $10^{-3}$, and a $5{,}000$-epoch cap. There are $324$ minibatch runs and $108$ full-batch runs. Each batch arm contributes $45$ runs in the selected fraction window; observed-event counts are $5$, $18$, $23$, and $31$ for batch sizes $128$, $512$, $2048$, and full batch, respectively. Full batch removes minibatch sampling but retains dropout. At $B_{\mathrm{batch}}\in\{128,512,2048\}$, \texttt{scripts/gamma\_by\_batch.py} fits $\log T=\log B+\gamma_{B_{\mathrm{batch}}}f+\epsilon$ with Gaussian residual $\epsilon$ of fitted constant variance. Observed events contribute a Gaussian density; unresolved runs contribute the probability of exceeding their recorded budget. Ordinary least squares on observed runs is reported separately. The fixed selected fraction window is $[0.220146,2.751828]$ in the current capacity units. Fits are made in both epochs and optimiser steps, using $\lceil n/B_{\mathrm{batch}}\rceil$ steps per epoch; full-batch runs have one step per epoch. The reported $\gamma_\infty+c'/\sqrt{B_{\mathrm{batch}}}$ curve is a least-squares fit to the three step-unit censored-regression slopes. The individual-run bootstrap uses $800$ resamples for censored regression and $4{,}000$ for observed-only OLS, and does not preserve clusters of shared seed labels. These results depend on the auxiliary residual model and are not substituted for the primary observed-event means or their uncertainty.

\paragraph{Learning-rate and extended-budget probes.}
\label{sec:probe-protocol}
These $18$ H100 random-label runs use $p=113$, $n=6{,}328$, seeds $52$ and $53$, and the main architecture, dropout, decay, and batch size. Twelve runs compare learning rates $10^{-4}$, $3\times10^{-4}$, and $10^{-3}$ at widths $216$ and $64$, with a $5{,}000$-epoch cap. All reach training accuracy $99\%$. At $d=216$ ($f=0.03575$), arithmetic mean times are $120.5$, $42$, and $17$ epochs, giving log--log learning-rate slope $-0.849$. At $d=64$ ($f=0.37770$), the corresponding means are $1507$, $558$, and $814$ epochs. The latter response is non-monotonic, so it does not establish learning-rate invariance.

Six additional runs extend the budget to $30{,}000$ epochs at learning rate $10^{-3}$. At $d=56$ ($f=0.48618$), the two events occur at $10{,}416$ and $11{,}724$ epochs. Neither seed reaches the threshold at $d=52$ ($f=0.55887$) or $d=50$ ($f=0.60150$). These observations demonstrate slow fitting beyond the original budget, but do not determine a population divergence or the sign of censoring bias in a fitted curvature.

\paragraph{Spectral quantities and character basis.}
\label{sec:character-protocol}
Here ``spectral'' refers to a decomposition over numerical residue values, not token positions or RoPE. Choose a primitive root $g$ modulo $p$ and order nonzero residues as $g^j$, $j=0,\ldots,p-2$. Multiplicative characters satisfy
\[
 \chi_k(g^j)=\exp\!\left(\frac{2\pi\mathrm{i}kj}{p-1}\right),
 \qquad \chi_k(a/b)=\chi_k(a)\overline{\chi_k(b)}
 \quad(a,b\ne0).
\]
This product identity motivates examining these features in division models. For an embedding or unembedding matrix, we reorder its nonzero-residue rows, subtract their column means, and take an unnormalised real Fourier transform along the residue axis. Let $Z_{k\ell}$ denote the resulting coefficient at frequency $k$ and feature coordinate $\ell$. The scripts retain bins $k=1,\ldots,m$, where $m=(p-1)/2$, and define
\[
 S_k=\sum_\ell |Z_{k\ell}|^2,\qquad
 a_k=\sqrt{S_k},\qquad q_k=\frac{S_k}{\sum_{j=1}^{m}S_j}.
\]
Thus $S_k$ is spectral power, $a_k$ its amplitude norm, and $q_k$ its share. A bin represents a conjugate pair except for the final, self-conjugate bin; the stored one-sided powers do not double the paired bins. We use that convention throughout the trace and ablation summaries. The effective number of bins is $(\sum_k q_k^2)^{-1}$. An increasing share can reflect shrinking power in other bins rather than absolute growth of the selected bin.

\paragraph{Character traces and rate fits.}
The trace experiments use $p=113$, seed $42$, the central architecture and optimiser, and no early stopping. The decay comparison uses $d=128$, $\alpha=0.5$, and $\lambda\in\{0.1,0.3,1\}$. The training-fraction comparison uses $d=128$, $\lambda=1$, and $\alpha\in\{0.3,0.5,0.7\}$; the width comparison uses $\alpha=0.5$, $\lambda=1$, and $d\in\{64,128,256,512\}$. Each setting has one seed. The script \texttt{scripts/character\_growth.py} trains for $2{,}500$ epochs in these comparisons, extended to $4{,}000$ for $\alpha=0.3$ and for $d=64$, and records powers and evaluation-mode training/validation accuracies every five epochs and at the final epoch. Its stored epoch index is zero-based and is recorded after an update epoch: index zero is already after one training epoch. These sampled evaluation-mode times differ from the primary online training-accuracy convention. Trace event locations are resolved only to the recording interval.

For trace analyses, ``winners'' are selected retrospectively by final unembedding share $q_k>2/m$. Takeoff is the first sampled time when their mean share exceeds $1.5/m$ (or $2/m$ for the second threshold). The decay-rate comparison fits each winner's $\log S_k$ against epoch in the interval from the first sampled validation-$99\%$ event through $300$ epochs afterwards, then averages the slopes across winners. The three mean power-growth rates are $(1.05,1.17,3.55)\times10^{-3}$ per epoch, with log--log decay exponent $+0.533$. They are post-transition measurements, not direct estimates of the preceding escape rate. The training-fraction and width analyses also fit the takeoff-to-validation-$99\%$ interval; settings with too few samples or nonpositive rates do not enter the corresponding log-rate regression.

The analyses fit power, despite the field names \texttt{amplitude} and \texttt{winner\_amp\_rate\_escape} in their saved outputs. Specifically,
\[
 \frac{d\log S_k}{dt}=2\frac{d\log a_k}{dt}.
\]
Amplitude-growth rates would be half the reported rates; scaling exponents across decay, width, or training fraction are unchanged by this conversion. We use spectral-power terminology in the text. Winner selection from the final checkpoint and the single-seed design limit inferences about selection timing and population growth laws.

\paragraph{Character ablations.}
The ablation cohort comprises ten saved models: $p\in\{97,113\}$, $d=128$, and seeds $42$--$46$. The intervention in \texttt{scripts/character\_ablation.py} changes only the unembedding matrix. Here winners use the separate threshold $q_k>3/m$, selecting four to nine bins. Exclusion removes their coefficients; restriction retains only them among the nonconstant character bins. Both preserve the row mean, the residue-zero row, and the special-token rows. Twenty controls per model remove an equal number of randomly chosen non-winner bins, sampled without replacement within each control using a NumPy generator seeded at zero. The controls match bin count, not removed spectral energy. Accuracy is evaluated with dropout disabled on all $p(p-1)$ task pairs, including training pairs. Median accuracies are $99.53\%$ before intervention, $53.36\%$ after exclusion, and $93.97\%$ after restriction; the median control loss is $0.011$ percentage points. These interventions establish functional importance of the selected output components in trained models; they do not establish how those components were learned or identify the training timescale.

\paragraph{Solvable character-network comparisons.}
The auxiliary model, based on \citet{gromov2023grokking}, is a quadratic two-layer network on modular subtraction with modulus $97$ and a random half-data training split. It is a separate model, rather than a reduction proved for the Transformer. The primary toy decay comparison uses $512$ hidden units, AdamW with learning rate $10^{-3}$, batch size $512$, cross-entropy loss, and seeds $0,1,2$. Unnormalised and hidden-only RMS-normalised arms use $\lambda\in\{0.05,0.1,0.2,0.3,0.5,1\}$. A fully normalised variant also normalises the readout direction and adds a decay-exempt scalar gain; it uses seeds $0,1$ and $\lambda\in\{0.1,0.2,0.3,0.5,1\}$. Runs have a $30{,}000$-epoch cap, record every five epochs, and continue $400$ epochs after validation reaches $99\%$, subject to the cap.

Reported toy times are arithmetic means of observed threshold times; escape delay subtracts the first training-$99\%$ time from the validation-$99\%$ time. Log--log fits give exponents $-0.940$ (validation time) and $-1.039$ (escape delay) in the unnormalised minibatch arm, versus $-0.451$ and $-0.457$ in the fully normalised arm. Hidden-only normalisation gives $-0.823$ and $-0.903$. Consequently the quoted $0.46$ concerns fully normalised escape delay, not every normalised model or every time statistic. A dropout-$0.2$ variant gives $-0.767$ and $-0.906$. Full-batch gradient-descent comparisons use a different protocol, switching after $100$ AdamW epochs; their exponent depends on the validation threshold ($-0.542$ at $99\%$, versus $-0.938$ at $50\%$). They do not support a claim that every unnormalised variant has exponent $-1$.

The unnormalised toy width sweep uses widths $128,256,512,1024,2048$, decays $0.03,0.1,0.3$, seeds $0,1$, and a $40{,}000$-epoch cap; its validation-time width exponents range from $0.072$ to $0.103$. The fully normalised comparison at decay $0.3$ uses widths $128,512,2048$ and gives $0.270$. The toy AdamW arms use the PyTorch default moment coefficients $(0.9,0.999)$, distinct from the Transformer setting. Toy evaluations include the initial state at epoch zero and occur before the next update epoch. The scripts and per-run settings accompany the derivation in \texttt{theory/principled/gen-solvable-circuit/}; these toy quantities are not pooled into the Transformer fits.

%% file: theory/principled.tex
\section{A stylised theory of the two timescales}
\label{sec:principled}

The measured laws are $T_{\mathrm{gen}}=Ap^aP^{-b}$ and $T_{\mathrm{mem}}=B\exp(\gamma f)$, with $f=K_{\mathrm{mem}}/(C_{\mathrm{model}}P)$. Their intersection geometry follows from these functional forms (\cref{sec:consequences}). Explaining why training produces the forms is a separate question. We develop two simplified accounts: repeated interference during memorisation, and a possible role for normalisation in the generalisation timescale. The assumptions connecting each model to the Transformer are stated explicitly. Neither account derives all fitted constants or the complete training dynamics.

\subsection{Memorisation: repeated interference between examples}
\label{sec:principled-mem}

An update that improves one training example can disturb others because they share parameters. Suppose that maintaining each stored example removes part of an attempted update. Repeated attenuation can make progress exponentially slower as the amount of stored information per parameter increases. The following construction makes this intuition precise under an idealised constraint geometry.

\paragraph{Projection model and exact expectation.}
Let $u\in\mathbb R^P$ be a nonzero attempted update and let $c$ be an integer with $0<c<P$. For each of $n$ stored examples, let $U_j$ be an independently and isotropically oriented $c$-dimensional subspace of constrained directions, and let $\Pi_j$ project onto $U_j^\perp$. Define $u_j=\Pi_j u_{j-1}$, with $u_0=u$, and $\rho=\|u_n\|^2/\|u\|^2$. Independence of $U_j$ from the preceding update gives
\[
 \mathbb E[\|u_j\|^2\mid u_{j-1}]=(1-c/P)\|u_{j-1}\|^2.
\]
Iterating this identity yields
\begin{equation}
 \mathbb E[\rho]=(1-c/P)^n\approx e^{-\nu},
 \qquad \nu=nc/P.
 \label{eq:projection}
\end{equation}
The exponential approximation requires $c/P$ to be small, with $nc^2/P^2$ controlling the log error. This is an expectation for sequential projections. Later projections generally reintroduce components along earlier constrained directions, so the resulting update need not satisfy all stored constraints simultaneously. Repeated local repair is the intended interpretation. The expected projection onto the original update obeys the same product law under these assumptions.

\paragraph{From retained update to memorisation time.}
Three additional assumptions connect the identity to a timescale: late fitting is dominated by progress on the last unfit examples; an epoch of local repair is approximated by the sequential projections; and the mean progress rate is proportional to the retained update fraction, with other factors absorbed into a prefactor. Approximating completion time by the inverse of this mean rate then gives $T_{\mathrm{mem}}\propto e^{nc/P}$. This step is a mean-rate approximation, not an identity between an expected hitting time and the reciprocal of expected retained power. With $K_{\mathrm{mem}}=n\log_2 V$, it becomes
\begin{equation}
 T_{\mathrm{mem}}\propto e^{\gamma f},\qquad
 \gamma=\frac{c\,C_{\mathrm{model}}}{\log_2 V}.
 \label{eq:gamma-pred}
\end{equation}
Thus the exponential depends on an intensive load, rather than total dataset information alone.

Taking $c=V-1$ is a further modelling choice, motivated by the number of logit gaps in a $V$-class classifier. It assumes that all these directions contribute to maintenance. At $V=p+2$ and the measured capacity constant, this gives $\gamma\in[11.60,15.39]$ over $p=97$ through $139$, of similar scale to the pooled empirical exponents $11.109$ (full range) and $15.470$ (selected window). This comparison was made after observing the empirical law. It is not a parameter-free derivation for the Transformer: its effective constraint dimension and orientation distribution have not been measured. Even the linear classifier has structured constraint subspaces, including a common logit-shift invariant block, so its exact geometry is not the independent isotropic model used in \cref{eq:projection}.

\paragraph{Noise, optimiser, and decay dependence.}
A proposed extension assumes that minibatch fluctuations impose an extra attenuation factor $\exp(-c'f/\sqrt{B_{\mathrm{batch}}})$ per repair cycle. Together with a deterministic exponent $\gamma_\infty$, this gives $\gamma(B_{\mathrm{batch}})=\gamma_\infty+c'/\sqrt{B_{\mathrm{batch}}}$. The $B_{\mathrm{batch}}^{-1/2}$ fluctuation scale alone does not establish this attenuation rule. The separately fitted censored-regression analysis gives $\gamma_\infty=3.98$ and $c'=319.34$ in the current capacity units, with residuals at most $0.47$. These are empirical inputs to the extension, not consequences of the projection identity.

Likewise, an increase in maintenance cost under weight decay and cancellation of update-magnitude dependence under AdamW require additional assumptions about margins and optimiser dynamics. The measured batch, decay, and learning-rate responses in \cref{sec:mech-mem} constrain such models; the projection calculation alone does not fix their signs or numerical values. The deterministic floor $\gamma_\infty$, deviations from an exponential, and the effective constraint geometry remain open. Specific counting-cost comparisons do not exclude other landscape-based explanations.

\subsection{Generalisation: arithmetic features and normalisation}
\label{sec:principled-gen}

\paragraph{What the character measurements establish.}
Fourier analyses of modular addition identify structured generalising circuits \citep{nanda2023progress}. For division, multiplicative characters supply the corresponding product identity (\cref{sec:character-protocol}). In our trained models, removing the four to nine dominant unembedding bins reduces median accuracy from $99.53\%$ to $53.36\%$, whereas retaining them preserves $93.97\%$. The controls and full-task evaluation are specified in \cref{sec:mechanism-protocols}. This is evidence that these output components contribute to the learned solution. It does not by itself derive how they were selected or how long learning took.

\paragraph{Selection and subsequent concentration.}
Retrospective traces show that eventual winners acquire an above-uniform share before validation reaches $99\%$, and that later concentration coincides with shrinking relative shares of losing bins. The initial spectrum is close to a Gaussian-matrix null in the tested summaries (mean differences up to $2.49\%$), and eventual winners have an approximately $6\%$ initial-share advantage ($p=0.02$). These observations support an initialisation bias, not determination of the winning set by initialisation alone.

In a simplified masked-target calculation the diagonal task-drive coefficient is the same for every character, proportional to training fraction $\alpha$. Equal diagonal coefficients do not eliminate off-diagonal interactions, finite-sample effects, or nonlinear feature learning. Nor do three single-seed decay traces establish decay-independent selection: their $1.5\times$-uniform takeoff indices are $20,30,25$, while their $2\times$ indices are $30,80,40$ at $\lambda=0.1,0.3,1$. Early selection is an observation; a universal selection law remains unestablished.

\paragraph{Bare factor growth.}
For factor amplitudes $u,v$, task coupling $s$, and decay $\lambda$, the linear system
\begin{equation}
 \dot u=sv-\lambda u,\qquad \dot v=su-\lambda v,
 \qquad \mu_\pm=-\lambda\pm s
 \label{eq:bilinear}
\end{equation}
has growth eigenvalues $\mu_\pm$. If one additionally assumes $s=c_s\lambda$ with fixed $c_s>1$, the growing rate is $(c_s-1)\lambda$. This gives an inverse-decay time for a fixed amplification factor, rather than a half-power. Without a model for $s$, even that decay dependence is not determined. The bare system therefore does not derive the matched Transformer estimate $T_{\mathrm{gen}}\propto\lambda^{-0.512}$.

\paragraph{A candidate normalisation-mediated timescale.}
For scale-invariant weights, decoupled decay and stochastic updates can reach a rotational equilibrium \citep{vanlaarhoven2017l2,wan2021spherical}. Under the AdamW noise-dominated equilibrium approximation, the characteristic angular update per optimiser step is
\[
 \delta\theta\simeq
 \sqrt{2\eta\lambda\frac{1-\beta_1}{1+\beta_1}}
 \qquad\text{\citep{kosson2024rotational}}.
\]
This is an angular-update magnitude, not a directed velocity towards the arithmetic solution. Obtaining a passage time proportional to $(\eta\lambda)^{-1/2}$ additionally requires approximately coherent progress through a fixed angular distance, with alignment and distance independent of the varied settings. An unbiased angular random walk instead accumulates squared displacement, giving a fixed-distance time proportional to $1/(\delta\theta)^2$. The equilibrium, scale-invariance, and directed-progress assumptions have not been demonstrated for the relevant weights during escape in this residual pre-normalised Transformer. Conversion from steps to epochs also depends on the number of minibatches; that factor is fixed in the matched decay comparison.

A fully normalised quadratic character network provides an illustrative comparison: the measured decay exponent is $-0.451$ for validation time and $-0.457$ for escape delay, whereas the corresponding unnormalised minibatch arm gives $-0.940$ and $-1.039$. The toy architecture and other variants are described in \cref{sec:mechanism-protocols}. These comparisons motivate the candidate mechanism but do not isolate it in the Transformer. Freezing RMSNorm gains preserves the normalisation operation and therefore does not test its removal.

The separate three-point trace comparison gives a post-transition spectral-power growth exponent $+0.533$. Its rates are measured after validation reaches $99\%$ and cannot directly identify the preceding escape timescale. A relation between those rates and directed pre-transition progress requires additional measurement.

\paragraph{What remains underived.}
The Transformer width exponent $b\simeq1.109$ is not obtained by this account. The auxiliary toy width exponents are approximately $0.072$--$0.103$ without normalisation and $0.270$ in the fully normalised comparison. Explaining the stronger measured width dependence requires a model of how the task drive, representations, and optimiser dynamics vary with width. The prime exponent $a$ is also not derived. The fitted constants of $T_{\mathrm{gen}}=Ap^aP^{-b}$ remain empirical; the normalisation argument supplies a candidate explanation of decay dependence under additional assumptions.

\subsection{Consequences and remaining empirical tests}
\label{sec:principled-ledger}

Setting the two laws equal gives a one-variable intersection equation whose
convexity predicts a fold at $p^{*} = 167.89$ $[148.84,193.38]$ beyond which the
fitted model has no grokking region, and, taken literally, a second root at
large $P$ (\cref{sec:consequences}). The $p = 197$ sweep found
the per-seed-minimum delay zero at all $18$ widths spanning
$P = 1.1\times10^{5}$--$10^{7}$, showing no onset across the refitted power-law interval
under that criterion. However, the ten-seed extension at $d=316$ has eight
positive delays and a zero minimum, so absence of delayed generalisation in
every seed is not established. At $p=149$, the first positive-minimum cell is $d=112$, but an unresolved smaller minimum leaves the first onset unidentified (\cref{sec:theory-prereg}). The upper root is not empirically established. Zero minima can coexist with positive mean delays (\cref{sec:consequences}).

The following auxiliary measurements further constrain the proposed mechanisms. They use the separate protocols and estimators in \cref{sec:mechanism-protocols}.

\begin{itemize}
\item \emph{Full-batch comparison.} Replacing minibatches with full-batch updates reduces $\gamma$ from $17.62$ to $7.19$ $[6.57, 7.66]$. Dropout remains active, so this does not remove all stochasticity. Moreover $\gamma(B_{\mathrm{batch}}) = 3.98 + 319.34/\sqrt{B_{\mathrm{batch}}}$
  \emph{evaluated at} $B_{\mathrm{batch}} = n$ predicts $7.27$--$8.72$: the $\sqrt{B_{\mathrm{batch}}}$ law
  gives values near, but above, the full-batch point estimate. These auxiliary values use the separately labelled Tobit analysis; they are not the primary arithmetic-cell-mean fits. The true $B_{\mathrm{batch}}\to\infty$ floor remains unmeasured.
\item \emph{Learning-rate dependence.} At $f=0.03575$, the probe gives $T_{\mathrm{mem}}\propto\eta^{-0.849}$ across three learning rates. At $f=0.37770$, the response is non-monotonic. These two-seed measurements constrain an invariance claim rather than confirming it (\cref{sec:probe-protocol}).
\item \emph{Longer follow-up.} At $f=0.48618$, both probe seeds reach the threshold only after $10{,}000$ epochs; at $f=0.55887$ and $0.60150$, neither does so within $30{,}000$ epochs. These results show why the training budget matters, without identifying a divergence or a bound on population curvature.
\item \emph{Training-fraction trace comparison.} Across $\alpha=0.3,0.5,0.7$, the fitted spectral-power growth rate between takeoff and validation-$99\%$ scales approximately as $\alpha^{4.28}$. Each setting has one seed, and the fitted intervals differ. This is a descriptive slope, not evidence sufficient to select a unique causal explanation of generalisation time.
\item \emph{Width trace comparison.} Across $d=64,128,256,512$, the corresponding power-growth rates have fitted width exponent $+0.215$, while validation time has exponent $-0.122$. The four measurements are non-monotonic and use one seed each. Neither exponent derives the parameter-count dependence $P^{-b}$ of the primary law.
\item \emph{Norm-gain freezing.} Freezing the RMSNorm gains preserves
  the qualitative decay dependence, while modestly changing generalisation
  times and the fitted decay exponent ($-0.61$ frozen versus $-0.66$
  unfrozen). Removing the normalisation operation remains untested.
\end{itemize}

%% file: checklist-revised.tex
\section*{NeurIPS Paper Checklist}

\begin{enumerate}

\item {\bf Claims}
    \item[] Question: Do the main claims made in the abstract and introduction accurately reflect the paper's contributions and scope?
    \item[] Answer: \answerYes{}
    \item[] Justification: The abstract and introduction place the paper within the broader literature and provide concrete contribution statements.
    \item[] Guidelines:
    \begin{itemize}
        \item The answer \answerNA{} means that the abstract and introduction do not include the claims made in the paper.
        \item The abstract and/or introduction should clearly state the claims made, including the contributions made in the paper and important assumptions and limitations. A \answerNo{} or \answerNA{} answer to this question will not be perceived well by the reviewers.
        \item The claims made should match theoretical and experimental results, and reflect how much the results can be expected to generalise to other settings.
        \item It is fine to include aspirational goals as motivation as long as it is clear that these goals are not attained by the paper.
    \end{itemize}

\item {\bf Limitations}
    \item[] Question: Does the paper discuss the limitations of the work performed by the authors?
    \item[] Answer: \answerYes{}
    \item[] Justification: The discussion and appendices state the shortcomings of the framework together with informative hyperparameter sweeps.
    \item[] Guidelines:
    \begin{itemize}
        \item The answer \answerNA{} means that the paper has no limitation while the answer \answerNo{} means that the paper has limitations, but those are not discussed in the paper.
        \item The authors are encouraged to create a separate ``Limitations'' section in their paper.
        \item The paper should point out any strong assumptions and how robust the results are to violations of these assumptions (e.g., independence assumptions, noiseless settings, model well-specification, asymptotic approximations only holding locally). The authors should reflect on how these assumptions might be violated in practice and what the implications would be.
        \item The authors should reflect on the scope of the claims made, e.g., if the approach was only tested on a few datasets or with a few runs. In general, empirical results often depend on implicit assumptions, which should be articulated.
        \item The authors should reflect on the factors that influence the performance of the approach. For example, a facial recognition algorithm may perform poorly when image resolution is low or images are taken in low lighting. Or a speech-to-text system might not be used reliably to provide closed captions for online lectures because it fails to handle technical jargon.
        \item The authors should discuss the computational efficiency of the proposed algorithms and how they scale with dataset size.
        \item If applicable, the authors should discuss possible limitations of their approach to address problems of privacy and fairness.
        \item While the authors might fear that complete honesty about limitations might be used by reviewers as grounds for rejection, a worse outcome might be that reviewers discover limitations that aren't acknowledged in the paper. The authors should use their best judgement and recognise that individual actions in favour of transparency play an important role in developing norms that preserve the integrity of the community. Reviewers will be specifically instructed to not penalise honesty concerning limitations.
    \end{itemize}

\item {\bf Theory assumptions and proofs}
    \item[] Question: For each theoretical result, does the paper provide the full set of assumptions and a complete (and correct) proof?
    \item[] Answer: \answerYes{}
    \item[] Justification: The propositions derive crossing geometry conditional on the empirical exponential and power-law functions. Their assumptions and algebra are given in the text. Mechanistic arguments are labelled as stylised candidates with additional assumptions, rather than proofs of Transformer dynamics.

    \item {\bf Experimental result reproducibility}
    \item[] Question: Does the paper fully disclose all the information needed to reproduce the main experimental results of the paper to the extent that it affects the main claims and/or conclusions of the paper (regardless of whether the code and data are provided or not)?
    \item[] Answer: \answerYes{}
    \item[] Justification: The experimental sections specify architecture, optimisation, dataset construction, accuracy thresholds, and seed pools. Code to replicate results is also provided in a GitHub link.
    \item[] Guidelines:
    \begin{itemize}
        \item The answer \answerNA{} means that the paper does not include experiments.
        \item If the paper includes experiments, a \answerNo{} answer to this question will not be perceived well by the reviewers: Making the paper reproducible is important, regardless of whether the code and data are provided or not.
        \item If the contribution is a dataset and\slash or model, the authors should describe the steps taken to make their results reproducible or verifiable.
        \item Depending on the contribution, reproducibility can be accomplished in various ways. For example, if the contribution is a novel architecture, describing the architecture fully might suffice, or if the contribution is a specific model and empirical evaluation, it may be necessary to either make it possible for others to replicate the model with the same dataset, or provide access to the model. In general. releasing code and data is often one good way to accomplish this, but reproducibility can also be provided via detailed instructions for how to replicate the results, access to a hosted model (e.g., in the case of a large language model), releasing of a model checkpoint, or other means that are appropriate to the research performed.
        \item While NeurIPS does not require releasing code, the conference does require all submissions to provide some reasonable avenue for reproducibility, which may depend on the nature of the contribution. For example
        \begin{enumerate}
            \item If the contribution is primarily a new algorithm, the paper should make it clear how to reproduce that algorithm.
            \item If the contribution is primarily a new model architecture, the paper should describe the architecture clearly and fully.
            \item If the contribution is a new model (e.g., a large language model), then there should either be a way to access this model for reproducing the results or a way to reproduce the model (e.g., with an open-source dataset or instructions for how to construct the dataset).
            \item We recognise that reproducibility may be tricky in some cases, in which case authors are welcome to describe the particular way they provide for reproducibility. In the case of closed-source models, it may be that access to the model is limited in some way (e.g., to registered users), but it should be possible for other researchers to have some path to reproducing or verifying the results.
        \end{enumerate}
    \end{itemize}

\item {\bf Open access to data and code}
    \item[] Question: Does the paper provide open access to the data and code, with sufficient instructions to faithfully reproduce the main experimental results, as described in supplemental material?
    \item[] Answer: \answerYes{}
    \item[] Justification: All code used to produce this paper is public on GitHub and linked inside the paper. The paper also contains enough information to replicate our experiments, with any extra information needed available to read out from our code. All data will be released.
    \item[] Guidelines:
    \begin{itemize}
        \item The answer \answerNA{} means that paper does not include experiments requiring code.
        \item Please see the NeurIPS code and data submission guidelines (\url{https://neurips.cc/public/guides/CodeSubmissionPolicy}) for more details.
        \item While we encourage the release of code and data, we understand that this might not be possible, so \answerNo{} is an acceptable answer. Papers cannot be rejected simply for not including code, unless this is central to the contribution (e.g., for a new open-source benchmark).
        \item The instructions should contain the exact command and environment needed to run to reproduce the results. See the NeurIPS code and data submission guidelines (\url{https://neurips.cc/public/guides/CodeSubmissionPolicy}) for more details.
        \item The authors should provide instructions on data access and preparation, including how to access the raw data, preprocessed data, intermediate data, and generated data, etc.
        \item The authors should provide scripts to reproduce all experimental results for the new proposed method and baselines. If only a subset of experiments are reproducible, they should state which ones are omitted from the script and why.
        \item At submission time, to preserve anonymity, the authors should release anonymised versions (if applicable).
        \item Providing as much information as possible in supplemental material (appended to the paper) is recommended, but including URLs to data and code is permitted.
    \end{itemize}

\item {\bf Experimental setting/details}
    \item[] Question: Does the paper specify all the training and test details (e.g., data splits, hyperparameters, how they were chosen, type of optimizer) necessary to understand the results?
    \item[] Answer: \answerYes{}
    \item[] Justification: The experimental sections specify the central protocol and extensions. The primary timescale and cone cohorts use ten planned seeds per cell; secondary sweeps use four or five, dense crossing studies five, and capacity calibration two.

\item {\bf Experiment statistical significance}
    \item[] Question: Does the paper report error bars suitably and correctly defined or other appropriate information about the statistical significance of the experiments?
    \item[] Answer: \answerYes{}
    \item[] Justification: Rank, calibration, and incremental-information tests are reported with identified-pair denominators. Bootstrap intervals identify their resampling unit and conditioning. Missing timescale crossings and unresolved grokking onsets are not reported as successes; nonsignificant nested tests are not interpreted as proof of sufficiency. Threshold and aggregation sensitivities are reported separately.
    \item[] Guidelines:
    \begin{itemize}
        \item The answer \answerNA{} means that the paper does not include experiments.
        \item The authors should answer \answerYes{} if the results are accompanied by error bars, confidence intervals, or statistical significance tests, at least for the experiments that support the main claims of the paper.
        \item The factors of variability that the error bars are capturing should be clearly stated (for example, train/test split, initialisation, random drawing of some parameter, or overall run with given experimental conditions).
        \item The method for calculating the error bars should be explained (closed form formula, call to a library function, bootstrap, etc.)
        \item The assumptions made should be given (e.g., Normally distributed errors).
        \item It should be clear whether the error bar is the standard deviation or the standard error of the mean.
        \item It is OK to report 1-sigma error bars, but one should state it. The authors should preferably report a 2-sigma error bar than state that they have a 96\% CI, if the hypothesis of Normality of errors is not verified.
        \item For asymmetric distributions, the authors should be careful not to show in tables or figures symmetric error bars that would yield results that are out of range (e.g., negative error rates).
        \item If error bars are reported in tables or plots, the authors should explain in the text how they were calculated and reference the corresponding figures or tables in the text.
    \end{itemize}

\item {\bf Experiments compute resources}
    \item[] Question: For each experiment, does the paper provide sufficient information on the computer resources (type of compute workers, memory, time of execution) needed to reproduce the experiments?
    \item[] Answer: \answerYes{}
    \item[] Justification: GPU types and the one-GPU-per-run configuration are stated, with per-run times where available.

\item {\bf Code of ethics}
    \item[] Question: Does the research conducted in the paper conform, in every respect, with the NeurIPS Code of Ethics \url{https://neurips.cc/public/EthicsGuidelines}?
    \item[] Answer: \answerYes{}
    \item[] Justification: This synthetic-task study has no human subjects or personal data.

\item {\bf Broader impacts}
    \item[] Question: Does the paper discuss both potential positive societal impacts and negative societal impacts of the work performed?
    \item[] Answer: \answerNA{}
    \item[] Justification: This is foundational research on the training dynamics of small Transformers on a synthetic algorithmic task. We see no direct path from these findings to deployment-level positive or negative societal impacts.
    \item[] Guidelines:
    \begin{itemize}
        \item The answer \answerNA{} means that there is no societal impact of the work performed.
        \item If the authors answer \answerNA{} or \answerNo, they should explain why their work has no societal impact or why the paper does not address societal impact.
        \item Examples of negative societal impacts include potential malicious or unintended uses (e.g., disinformation, generating fake profiles, surveillance), fairness considerations (e.g., deployment of technologies that could make decisions that unfairly impact specific groups), privacy considerations, and security considerations.
        \item The conference expects that many papers will be foundational research and not tied to particular applications, let alone deployments. However, if there is a direct path to any negative applications, the authors should point it out. For example, it is legitimate to point out that an improvement in the quality of generative models could be used to generate Deepfakes for disinformation. On the other hand, it is not needed to point out that a generic algorithm for optimising neural networks could enable people to train models that generate Deepfakes faster.
        \item The authors should consider possible harms that could arise when the technology is being used as intended and functioning correctly, harms that could arise when the technology is being used as intended but gives incorrect results, and harms following from (intentional or unintentional) misuse of the technology.
        \item If there are negative societal impacts, the authors could also discuss possible mitigation strategies (e.g., gated release of models, providing defenses in addition to attacks, mechanisms for monitoring misuse, mechanisms to monitor how a system learns from feedback over time, improving the efficiency and accessibility of ML).
    \end{itemize}

\item {\bf Safeguards}
    \item[] Question: Does the paper describe safeguards that have been put in place for responsible release of data or models that have a high risk for misuse (e.g., pre-trained language models, image generators, or scraped datasets)?
    \item[] Answer: \answerNA{}
    \item[] Justification: The paper does not release pre-trained language models, image generators, or scraped datasets; the intended artefact is research code that trains small Transformers on synthetic modular-arithmetic data and analyses the resulting training curves. There is no comparable misuse risk.
    \item[] Guidelines:
    \begin{itemize}
        \item The answer \answerNA{} means that the paper poses no such risks.
        \item Released models that have a high risk for misuse or dual-use should be released with necessary safeguards to allow for controlled use of the model, for example by requiring that users adhere to usage guidelines or restrictions to access the model or implementing safety filters.
        \item Datasets that have been scraped from the Internet could pose safety risks. The authors should describe how they avoided releasing unsafe images.
        \item We recognise that providing effective safeguards is challenging, and many papers do not require this, but we encourage authors to take this into account and make a best faith effort.
    \end{itemize}

\item {\bf Licenses for existing assets}
    \item[] Question: Are the creators or original owners of assets (e.g., code, data, models), used in the paper, properly credited and are the license and terms of use explicitly mentioned and properly respected?
    \item[] Answer: \answerYes{}
    \item[] Justification: The study uses scientific software libraries and synthetic data. Attribution and licensing information is in the linked GitHub repo.

\item {\bf New assets}
    \item[] Question: Are new assets introduced in the paper well documented and is the documentation provided alongside the assets?
    \item[] Answer: \answerNA{}
    \item[] Justification: The paper does not release new assets.
    \item[] Guidelines:
    \begin{itemize}
        \item The answer \answerNA{} means that the paper does not release new assets.
        \item Researchers should communicate the details of the dataset\slash code\slash model as part of their submissions via structured templates. This includes details about training, license, limitations, etc.
        \item The paper should discuss whether and how consent was obtained from people whose asset is used.
        \item At submission time, remember to anonymize your assets (if applicable). You can either create an anonymised URL or include an anonymised zip file.
    \end{itemize}

\item {\bf Crowdsourcing and research with human subjects}
    \item[] Question: For crowdsourcing experiments and research with human subjects, does the paper include the full text of instructions given to participants and screenshots, if applicable, as well as details about compensation (if any)?
    \item[] Answer: \answerNA{}
    \item[] Justification: The paper does not involve crowdsourcing or research with human subjects.

\item {\bf Institutional review board (IRB) approvals or equivalent for research with human subjects}
    \item[] Question: Does the paper describe potential risks incurred by study participants, whether such risks were disclosed to the subjects, and whether Institutional Review Board (IRB) approvals (or an equivalent approval/review based on the requirements of your country or institution) were obtained?
    \item[] Answer: \answerNA{}
    \item[] Justification: The paper does not involve crowdsourcing or research with human subjects.
    \item[] Guidelines:
    \begin{itemize}
        \item The answer \answerNA{} means that the paper does not involve crowdsourcing nor research with human subjects.
        \item Depending on the country in which research is conducted, IRB approval (or equivalent) may be required for any human subjects research. If you obtained IRB approval, you should clearly state this in the paper.
        \item We recognise that the procedures for this may vary significantly between institutions and locations, and we expect authors to adhere to the NeurIPS Code of Ethics and the guidelines for their institution.
        \item For initial submissions, do not include any information that would break anonymity (if applicable), such as the institution conducting the review.
    \end{itemize}

\item {\bf Declaration of LLM usage}
    \item[] Question: Does the paper describe the usage of LLMs if it is an important, original, or non-standard component of the core methods in this research? Note that if the LLM is used only for writing, editing, or formatting purposes and does \emph{not} impact the core methodology, scientific rigor, or originality of the research, declaration is not required.
    \item[] Answer: \answerYes{}
    \item[] Justification: An LLM coding assistant helped implement code, analyse data, work through mathematical derivations, and revise manuscript text. However, the core methodology developed in this research did not involve the original contribution of LLMs.

\end{enumerate}

%% file: references.bib
@article{morris2025memorize,
  title={How much do language models memorize?},
  author={Morris, John X and Sitawarin, Chawin and Guo, Chuan and Kokhlikyan, Narine and Suh, G Edward and Rush, Alexander M and Chaudhuri, Kamalika and Mahloujifar, Saeed},
  journal={arXiv preprint arXiv:2505.24832},
  year={2025}
}

@article{power2022grokking,
  title={Grokking: Generalization beyond overfitting on small algorithmic datasets},
  author={Power, Alethea and Burda, Yuri and Edwards, Harri and Babuschkin, Igor and Misra, Vedant},
  journal={arXiv preprint arXiv:2201.02177},
  year={2022}
}

@article{davies2023unifying,
  title   = {Unifying Grokking and Double Descent},
  author  = {Davies, Xander and Langosco, Lauro and Krueger, David},
  journal = {arXiv preprint arXiv:2303.06173},
  year    = {2023}
}

@article{varma2023circuit,
  title   = {Explaining Grokking through Circuit Efficiency},
  author  = {Varma, Vikrant and Shah, Rohin and Kenton, Zachary and Kram{\'a}r, J{\'a}nos and Kumar, Ramana},
  journal = {arXiv preprint arXiv:2309.02390},
  year    = {2023}
}

@inproceedings{mohamadi2024why,
  title     = {Why Do You Grok? {A} Theoretical Analysis on Grokking Modular Addition},
  author    = {Mohamadi, Mohamad Amin and Li, Zhiyuan and Wu, Lei and Sutherland, Danica J.},
  booktitle = {Proceedings of the 41st International Conference on Machine Learning},
  series    = {Proceedings of Machine Learning Research},
  volume    = {235},
  pages     = {35934--35967},
  year      = {2024},
  publisher = {PMLR}
}

@article{gardner1988space,
  title={The space of interactions in neural network models},
  author={Gardner, Elizabeth},
  journal={Journal of physics A: Mathematical and general},
  volume={21},
  number={1},
  pages={257},
  year={1988},
  publisher={IOP Publishing}
}

@inproceedings{carlini2022quantifying,
title={Quantifying Memorization Across Neural Language Models},
author={Nicholas Carlini and Daphne Ippolito and Matthew Jagielski and Katherine Lee and Florian Tramer and Chiyuan Zhang},
booktitle={The Eleventh International Conference on Learning Representations },
year={2023},
url={https://openreview.net/forum?id=TatRHT_1cK}
}

@article{wang2024grokked,
  title={Grokked transformers are implicit reasoners: A mechanistic journey to the edge of generalization},
  author={Wang, Boshi and Yue, Xiang and Su, Yu and Sun, Huan},
  journal={arXiv preprint arXiv:2405.15071},
  year={2024}
}

@article{liu2022towards,
  title={Towards understanding grokking: An effective theory of representation learning},
  author={Liu, Ziming and Kitouni, Ouail and Nolte, Niklas S and Michaud, Eric and Tegmark, Max and Williams, Mike},
  journal={Advances in Neural Information Processing Systems},
  volume={35},
  pages={34651--34663},
  year={2022}
}

@article{nanda2023progress,
  title={Progress measures for grokking via mechanistic interpretability},
author={Neel Nanda and Lawrence Chan and Tom Lieberum and Jess Smith and Jacob Steinhardt},
booktitle={The Eleventh International Conference on Learning Representations },
year={2023},
url={https://openreview.net/forum?id=9XFSbDPmdW}
}

@article{loshchilov2017decoupled,
  title={Decoupled weight decay regularization},
  author={Loshchilov, Ilya and Hutter, Frank},
  journal={arXiv preprint arXiv:1711.05101},
  year={2017}
}

@article{tirumala2022memorization,
  title={Memorization without overfitting: Analyzing the training dynamics of large language models},
  author={Tirumala, Kushal and Markosyan, Aram and Zettlemoyer, Luke and Aghajanyan, Armen},
  journal={Advances in Neural Information Processing Systems},
  volume={35},
  pages={38274--38290},
  year={2022}
}

@inproceedings{roberts2020much,
  title={How much knowledge can you pack into the parameters of a language model?},
  author={Roberts, Adam and Raffel, Colin and Shazeer, Noam},
  booktitle={Proceedings of the 2020 conference on empirical methods in natural language processing (EMNLP)},
  pages={5418--5426},
  year={2020}
}

@inproceedings{lu2024scaling,
  title={Scaling laws for fact memorization of large language models},
  author={Lu, Xingyu and Li, Xiaonan and Cheng, Qinyuan and Ding, Kai and Huang, Xuan-Jing and Qiu, Xipeng},
  booktitle={Findings of the Association for Computational Linguistics: EMNLP 2024},
  pages={11263--11282},
  year={2024}
}

@article{allen2024physics,
  title={Physics of language models: Part 3.3, knowledge capacity scaling laws},
  author={Allen-Zhu, Zeyuan and Li, Yuanzhi},
  journal={arXiv preprint arXiv:2404.05405},
  year={2024}
}

@inproceedings{zhang2017rethinking,
  title        = {Understanding Deep Learning Requires Rethinking Generalization},
  author       = {Zhang, Chiyuan and Bengio, Samy and Hardt, Moritz and Recht, Benjamin and Vinyals, Oriol},
  booktitle    = {International Conference on Learning Representations (ICLR)},
  year         = {2017},
  url          = {https://arxiv.org/abs/1611.03530}
}

@inproceedings{arpit2017closer,
  title={A closer look at memorization in deep networks},
  author={Arpit, Devansh and Jastrz{\k{e}}bski, Stanis{\l}aw and Ballas, Nicolas and Krueger, David and Bengio, Emmanuel and Kanwal, Maxinder S and Maharaj, Tegan and Fischer, Asja and Courville, Aaron and Bengio, Yoshua and others},
  booktitle={International conference on machine learning},
  pages={233--242},
  year={2017},
  organization={PMLR}
}

@inproceedings{
humayun2024deep,
title={Deep Networks Always Grok and Here is Why},
author={Ahmed Imtiaz Humayun and Randall Balestriero and Richard Baraniuk},
booktitle={Forty-first International Conference on Machine Learning},
year={2024},
url={https://openreview.net/forum?id=zMue490KMr}
}

@article{furuta2024interpreting,
  title={Towards empirical interpretation of internal circuits and properties in grokked transformers on modular polynomials},
  author={Furuta, Hiroki and Minegishi, Gouki and Iwasawa, Yusuke and Matsuo, Yutaka},
  journal={arXiv preprint arXiv:2402.16726},
  year={2024}
}

@inproceedings{he2024learning,
  title={Learning to grok: Emergence of in-context learning and skill composition in modular arithmetic tasks},
  author={He, Tianyu and Doshi, Darshil and Das, Aritra and Gromov, Andrey},
  booktitle={ICML 2024 Workshop on Mechanistic Interpretability}
}

@article{rubin2023grokking,
  title         = {Grokking as a first order phase transition in two layer networks},
  author        = {Rubin, Noa and Seroussi, Inbar and Ringel, Zohar},
  journal       = {arXiv preprint arXiv:2310.03789},
  year = {2023}
}

@inproceedings{lyu2024dichotomy,
  title={Dichotomy of early and late phase implicit biases can provably induce grokking},
  author={Lyu, Kaifeng and Jin, Jikai and Li, Zhiyuan and Du, Simon and Lee, Jason and Hu, Wei},
  booktitle={International Conference on Learning Representations},
  volume={2024},
  pages={33897--33936},
  year={2024}
}

@article{clauw2024information,
  title={Information-theoretic progress measures reveal grokking is an emergent phase transition},
  author={Clauw, Kenzo and Stramaglia, Sebastiano and Marinazzo, Daniele},
  journal={arXiv preprint arXiv:2408.08944},
  year={2024}
}

@article{huang2024unified,
  title={Unified view of grokking, double descent and emergent abilities: A perspective from circuits competition},
  author={Huang, Yufei and Hu, Shengding and Han, Xu and Liu, Zhiyuan and Sun, Maosong},
  journal={arXiv preprint arXiv:2402.15175},
  year={2024}
}

@article{thilak2022slingshot,
  title={The slingshot mechanism: An empirical study of adaptive optimizers and the grokking phenomenon},
  author={Thilak, Vimal and Littwin, Etai and Zhai, Shuangfei and Saremi, Omid and Paiss, Roni and Susskind, Joshua},
  journal={arXiv preprint arXiv:2206.04817},
  year={2022}
}

@article{lee2024grokfast,
  title={Grokfast: Accelerated grokking by amplifying slow gradients},
  author={Lee, Jaerin and Kang, Bong Gyun and Kim, Kihoon and Lee, Kyoung Mu},
  journal={arXiv preprint arXiv:2405.20233},
  year={2024}
}

@inproceedings{nakkiran2020deep,
title={Deep Double Descent: Where Bigger Models and More Data Hurt},
author={Preetum Nakkiran and Gal Kaplun and Yamini Bansal and Tristan Yang and Boaz Barak and Ilya Sutskever},
booktitle={International Conference on Learning Representations},
year={2020},
url={https://openreview.net/forum?id=B1g5sA4twr}
}

@article{liu2022omnigrok,
  title={Omnigrok: Grokking beyond algorithmic data},
  author={Liu, Ziming and Michaud, Eric J and Tegmark, Max},
  journal={arXiv preprint arXiv:2210.01117},
  year={2022}
}

@inproceedings{kumar2024grokking,
  title={Grokking as the transition from lazy to rich training dynamics},
  author={Kumar, Tanishq and Bordelon, Blake and Gershman, Samuel and Pehlevan, Cengiz},
  booktitle={International Conference on Learning Representations},
  volume={2024},
  pages={23010--23035},
  year={2024}
}

@article{merrill2023tale,
title={A Tale of Two Circuits: Grokking as Competition of Sparse and Dense Subnetworks},
author={William Merrill and Nikolaos Tsilivis and Aman Shukla},
booktitle={ICLR 2023 Workshop on Mathematical and Empirical Understanding of Foundation Models},
year={2023},
url={https://openreview.net/forum?id=8GZxtu46Kx}
}

@article{demoss2025complexity,
  title={The complexity dynamics of grokking},
  author={DeMoss, Branton and Sapora, Silvia and Foerster, Jakob and Hawes, Nick and Posner, Ingmar},
  journal={Physica D: Nonlinear Phenomena},
  pages={134859},
  year={2025},
  publisher={Elsevier}
}

@article{manir2026systematic,
  title={A systematic empirical study of grokking: Depth, architecture, activation, and regularization},
  author={Manir, Shalima Binta and Rupa, Anamika Paul},
  journal={arXiv preprint arXiv:2603.25009},
  year={2026}
}

@article{gromov2023grokking,
  author        = {Gromov, Andrey},
  title         = {Grokking modular arithmetic},
  journal       = {arXiv preprint arXiv:2301.02679},
  year          = {2023}
}

@article{vanlaarhoven2017l2,
  author        = {van Laarhoven, Twan},
  title         = {L2 regularization versus batch and weight normalization},
  journal       = {arXiv preprint arXiv:1706.05350},
  year          = {2017}
}

@inproceedings{wan2021spherical,
  author        = {Wan, Ruosi and Zhu, Zhanxing and Zhang, Xiangyu and Sun, Jian},
  title         = {Spherical Motion Dynamics: Learning Dynamics of Normalized Neural Network using {SGD} and Weight Decay},
  booktitle     = {Advances in Neural Information Processing Systems},
  year          = {2021}
}

@inproceedings{kosson2024rotational,
  author        = {Kosson, Atli and Messmer, Bettina and Jaggi, Martin},
  title         = {Rotational Equilibrium: How Weight Decay Balances Learning Across Neural Networks},
  booktitle     = {International Conference on Machine Learning},
  year          = {2024}
}

@misc{pearce2023memorize,
  title        = {Do Machine Learning Models Memorize or Generalize?},
  author       = {Pearce, Adam and Ghandeharioun, Asma and Hussein, Nada and Thain, Nithum and Wattenberg, Martin and Dixon, Lucas},
  year         = {2023},
  month        = aug,
  howpublished = {Google PAIR Explorables},
  url          = {https://pair.withgoogle.com/explorables/grokking/}
}

@article{minegishi2025bridging,
  title   = {Bridging Lottery Ticket and Grokking: Understanding Grokking from Inner Structure of Networks},
  author  = {Minegishi, Gouki and Iwasawa, Yusuke and Matsuo, Yutaka},
  journal = {Transactions on Machine Learning Research},
  year    = {2025},
  url     = {https://openreview.net/forum?id=eQeYyup1tm}
}
